\documentclass[10pt,journal,compsoc]{IEEEtran}
\usepackage{color}

\ifCLASSOPTIONcompsoc
\usepackage[nocompress]{cite}
\else
\usepackage{cite}
\fi

\usepackage[utf8]{inputenc}
\usepackage{longtable}
\usepackage{graphicx}
\usepackage{color}

\usepackage[cmex10]{amsmath}

\usepackage{array}

\usepackage{mdwmath}
\usepackage{mdwtab}
\usepackage{subfig}
\usepackage{url}

\usepackage{multirow}
\usepackage{booktabs}
\usepackage{amssymb}
\usepackage{lscape}

\usepackage{rotating}

\usepackage{tikz}
\usepackage{pifont}

\usepackage{nicefrac}

\newcommand{\image}{I}              		
\newcommand{\images}{\textbf{I}}     	
\newcommand{\nimages}{n}            		

\newcommand{\leftimage}{\image_{left}}              
\newcommand{\rightimage}{\image_{right}}          
\newcommand{\referenceimage}{\image_{ref}}     

\newcommand{\depthmap}{D}			
\newcommand{\estimateddepthmap}{\hat{\depthmap}}  
\newcommand{\depth}{d}				
\newcommand{\estimateddepth}{\hat\depth}				

\newcommand{\disparitymap}{D}		
\newcommand{\leftdisparitymap}{\disparitymap_{left}}		
\newcommand{\rightdisparitymap}{\disparitymap_{right}}		
\newcommand{\estimateddisparitymap}{\hat{\disparitymap}}  
\newcommand{\disparity}{d}			
\newcommand{\estimateddisparity}{\hat{\disparity}}			

\newcommand{\costvolume}{C}              		

\newcommand{\pixel}{x}				

\newcommand{\npixels}{N}

\newcommand{\recofunc}{f}               	
\newcommand{\confidencefunc}{C} 		
\newcommand{\threshfunc}{H}			
  
\newcommand{\params}{\theta}			

\newcommand{\featuredim}{c}			
\newcommand{\featuremap}{\textbf{f}}	

\newcommand{\objectivefunc}{\mathcal{L}}
\newcommand{\loss}{\objectivefunc}			
\newcommand{\weights}{W}				

\newcommand{\ie}{\emph{i.e., }}
\newcommand{\eg}{\emph{e.g., }}
\newcommand{\etal}{\emph{et al.}}
\newcommand{\noi}{\noindent}

\newcommand{\ltwo}{L_2}						
\newcommand{\lone}{L_1}						

\newcommand{\width}{W}						
\newcommand{\height}{H}						

\newcommand{\ndisparities}{n_d}				
\newcommand{\neighborhood}{\mathcal{N}}		

\newcommand{\Distance}{\mathcal{D}}
\newcommand{\distance}{d}

\newcommand{\energy}{E}			
\newcommand{\smoothnessenergy}{\energy_s}	

\newcommand{\argmin}{\arg\min}

\graphicspath{{figures/}{figures/feature_learning/}{figures/plots/}{figures/plots/Badn/}{figures/plots/MSE/}{figures/GT_data/}{figures/ApolloScape/}}
\DeclareGraphicsExtensions{.pdf,.eps, .png}

\begin{document}
\bstctlcite{IEEEexample:BSTcontrol}


\title{A Survey on Deep Learning Techniques\\ for Stereo-based Depth Estimation}

\author{Hamid Laga,  Laurent Valentin Jospin, Farid Boussaid,   \and Mohammed Bennamoun~\IEEEmembership{Senior Member,~IEEE}
	\IEEEcompsocitemizethanks{
		\IEEEcompsocthanksitem Hamid Laga  is with the Information Technology, Mathematics and Statistics Discipline, Murdoch University (Australia), and with the Phenomics and Bioinformatics Research Centre, University of South Australia. Email: H.Laga@murdoch.edu.au
		
		\IEEEcompsocthanksitem Laurent Valentin Jospin is with the University of Western Australia, Perth, WA 6009, Australia. Email: laurent.jospin@research.uwa.edu.au
		
		\IEEEcompsocthanksitem Farid Boussaid is with the University of Western Australia, Perth, WA 6009, Australia. Email: farid.boussaid@uwa.edu.au
		\IEEEcompsocthanksitem Mohammed Bennamoun is with the University of Western Australia, Perth, WA 6009, Australia. Email: mohammed.bennamoun@uwa.edu.au
	}
	\thanks{Manuscript received June, 2020; revised June, 2020.}}

%
%

\markboth{A Survey on Deep Learning Techniques for Stereo-based Depth Estimation}%
{Laga \MakeLowercase{\etalnospace}: A Survey on Deep Learning Techniques for Stereo-based Depth Estimation}

\IEEEcompsoctitleabstractindextext{
	\begin{abstract}
		Estimating depth from RGB images  is a long-standing ill-posed problem, which has been explored for decades by the computer vision,  graphics, and machine learning communities.  Among  the existing techniques, stereo matching remains one of the most widely used in the literature due to its strong connection to the human binocular system. Traditionally, stereo-based depth estimation has been addressed through matching hand-crafted features across multiple images. Despite the extensive amount of research, these traditional techniques still suffer in the presence of highly textured areas, large uniform regions, and occlusions. Motivated by their growing success in solving various  2D and 3D vision problems, deep learning  for stereo-based depth estimation has attracted a growing interest from the community, with more than 150 papers published in this area between 2014 and 2019. This new generation of methods has demonstrated a significant leap in performance, enabling  applications such as autonomous driving and augmented reality.  In this article, we provide a comprehensive survey of this new and continuously growing field of research, summarize the  most commonly used pipelines, and discuss their benefits and limitations. In retrospect of what has been achieved so far, we also conjecture what the future may hold for deep learning-based stereo for depth estimation research. 
		
	\end{abstract}

}

\maketitle


\IEEEdisplaynotcompsoctitleabstractindextext

%
\IEEEpeerreviewmaketitle

\section{Introduction}
\label{sec:introduction}

\IEEEPARstart{D}{epth} estimation from one or multiple RGB images  is a long standing ill-posed problem, with applications in various domains such as  robotics, autonomous driving, object recognition and scene understanding, 3D modeling and animation, augmented reality, industrial control, and medical diagnosis.  This problem has been extensively investigated for many decades. Among all the techniques that have been proposed in the literature, stereo matching is  traditionally the most explored one  due to its strong connection to the human binocular system.  

The \textbf{first} generation of  stereo-based depth estimation methods  relied typically   on matching pixels across  multiple images captured using accurately calibrated cameras. Although these techniques can achieve good results, they are still limited in many aspects. For instance, they are not suitable when dealing with occlusions, featureless regions, or highly textured regions with repetitive patterns. 
Interestingly, we, as humans, are good at solving such ill-posed inverse problems by leveraging prior knowledge. For example, we can easily infer the approximate  sizes of  objects, their relative locations, and even their approximate relative distance to our eye(s).  We can do this because all the previously seen objects and scenes have enabled us to build  prior knowledge and develop mental models of how the 3D world   looks like. The \textbf{second} generation of  methods tries to leverage this prior knowledge by formulating the problem as a learning task. The advent of deep learning techniques in computer vision~\cite{khan2018guide} coupled with the increasing availability of large training datasets, have led to a \textbf{third} generation of methods that are able to recover the lost dimension.  Despite being recent, these methods have demonstrated  exciting and promising results on various tasks related to computer vision and graphics.

In this article,  we provide a comprehensive and structured review of the recent advances in stereo image-based depth estimation using deep learning  techniques. These methods use two or more images captured with spatially-distributed RGB cameras\footnote{Deep learning-based depth estimation from monocular images and videos is an emerging field and requires  a separate survey.}. We have gathered more than $150$ papers, which appeared between January $2014$  and December $2019$ in leading computer vision, computer graphics, and machine learning conferences and journals\footnote{At the time of writing this article.}.  The goal is to help the reader navigate in this emerging field, which has gained a significant momentum in the past few years. 

The major contributions of this article are as follows;
\begin{itemize}
	\item To the best of our knowledge, this is the first article that surveys stereo-based depth estimation using deep learning techniques. We present a comprehensive review of  more than $150$ papers, which appeared in the past six years in leading conferences and journals.
	
	\item We provide a comprehensive taxonomy of the state-of-the-art. We first describe the common pipelines and then discuss the similarities and differences between methods within each pipeline.
	 
	\item  We provide a comprehensive review and an insightful analysis on all the aspects of the problem, including the training data, the network architectures and their effect on the  reconstruction performance, the training strategies, and the generalization ability. 
	
	\item We provide a comparative summary of the properties and performances of some key methods using publicly available datasets and in-house images. The latter have been chosen to test how these methods would perform on completely new scenarios. 
	
\end{itemize} 

\noi The rest of this article is organized as follows; Section~\ref{sec:problemstatement} formulates the problem and lays down the taxonomy. Section~\ref{sec:datasets} surveys the various datasets which have been used to train and test stereo-based depth reconstruction algorithms.  Section~\ref{sec:depth_by_stereo_matching}  focuses on the  works that use deep learning architectures to learn how to match pixels across images.  Section~\ref{sec:end_to_end_stereo} reviews  the end-to-end methods for stereo matching, while Section~\ref{sec:mvs_architectures} discusses how these methods have been extended to  the multi-view stereo case.   Section~\ref{sec:training_end_to_end} focuses on the training procedures including the choice of the loss functions and the degree of supervision.   Section~\ref{sec:discussion_and_comparison}  discusses the performance of  key methods. Finally, Section~\ref{sec:future}  discusses the potential future research directions, while Section~\ref{sec:conclusion}  summarizes the main contributions of this article. 


\section{Scope and taxonomy}
\label{sec:problemstatement}

Let $\images = \{\image_k, k=1, \dots, \nimages \}$ be a set of $\nimages \ge 1$ RGB images of the same 3D scene, captured using cameras whose intrinsic and extrinsic parameters can be   \emph{known} or \emph{unknown}.   The goal is to estimate one or multiple depth maps, which can be from the same viewpoint as the input~\cite{li2015depth,eigen2015predicting,garg2016unsupervised,liu2016learning},   or from a new arbitrary viewpoint~\cite{yang2015weakly,kulkarni2015deep,tatarchenko2016multi,zhou2016view,park2017transformation}.  This article focuses on deep learning methods for  stereo-based depth estimation, \ie $\nimages = 2$ in the case of stereo matching, and $\nimages > 2$ for the case of Multi-View Stereo (MVS). Monocular and video-based depth estimation methods are beyond the scope of this article and require a separate survey.




Learning-based depth reconstruction can be summarized as the process of learning a predictor $\recofunc_{\params}$ that can infer from the set of images $\images$,  a depth map $\estimateddepthmap$ that is as close as possible to the unknown depth map $\depthmap$.  In other words, we seek to find a function $\recofunc_{\params}$ such that $ \objectivefunc(\images) = \distance\left(\recofunc_{\theta}(\images), \depthmap  \right)$ is minimized. Here, $\theta$ is a set of parameters, and  $\distance(\cdot, \cdot)$ is a certain measure of distance between the real depth map $\depthmap$ and the reconstructed depth map $\recofunc_\params(\images)$.  The reconstruction objective  $\objectivefunc$ is also known as the  \emph{loss function}.  

We can distinguish two main categories of methods. Methods in the \textbf{first} class (Section~\ref{sec:depth_by_stereo_matching}) mimic the traditional stereo-matching techniques~\cite{scharstein2002taxonomy} by explicitly learning how to match, or put in correspondence, pixels across the input images. Such correspondences can then be converted into an optical flow or a disparity map, which in turn can be converted into depth at each  pixel in the reference image.   The predictor $\recofunc$ is composed of three modules: a feature extraction module, a feature matching and cost aggregation module, and a disparity/depth estimation module. Each module is trained independently from the others. 

The \textbf{second} class of methods (Section~\ref{sec:end_to_end_stereo}) solves the stereo matching problem using a pipeline that is trainable end-to-end.  Two main classes of methods have been proposed. Early methods formulated the depth estimation as a regression problem. In other words, the depth map is directly regressed from the input without explicitly matching features across the views. While these methods are simple and fast at runtime, they require a large amount of training data, which is hard to obtain. Methods in the second class mimic the traditional stereo matching pipeline by breaking the problem into stages composed of differentiable blocks and thus allowing end-to-end training. While a large body of the literature focused on pairwise stereo methods, several papers have also addressed the multi-view stereo case and these will be reviews in Section~\ref{sec:mvs_architectures}.

In all methods, the estimated depth maps can  be further refined using refinement modules~\cite{eigen2014depth,li2015depth,wang2015designing,eigen2015predicting} and/or  progressive reconstruction strategies where the reconstruction is  refined every time new images become available. 

Finally, the performance of deep learning-based stereo methods depends not only on the network architecture but also on the datasets on which they have been trained (Section~\ref{sec:datasets}) and on the training procedure used to optimise their parameters (Section~\ref{sec:training_end_to_end}). The latter includes the choice of the loss functions and the supervision mode, which can be fully supervised with 3D annotations, weakly supervised, or self-supervised. We will discuss all these aspects in the subsequent sections.


\section{Datasets}
\label{sec:datasets}

\begin{table*}[t]
	\caption{\label{tab:3ddatasets_sterep}\label{tab:3ddatasets_mvs}Datasets for depth/disparity estimation. "GT": ground-truth, "Tr.": training, "Ts.": testing, "fr.": frames, "Vol.": volumetric,  "Eucl": Euclidean, "Ord": ordinal, "Int.": intrinsic, "Ext.": extrinsic.   }
	\resizebox{\linewidth}{!}
	{

    \begin{tabular}{@{}p{8.5em}@{ }c@{ } c@{ }p{11.915em}c@{ }p{7.5em}@{ }p{5.335em}@{ }p{5.5em}c@{ }c@{ }l@{ }p{8.665em}lll @{ }l  @{ }l@{ }l@{}}
   \toprule
   &   \multirow{2}{*}{\textbf{Year}}  &   \multirow{2}{7em}{\textbf{Type}} & \multicolumn{1}{p{11.915em}}{\multirow{2}{*}{\textbf{Purpose}}} & \multicolumn{5}{c}{\textbf{Images}}  & & \multicolumn{5}{c}{\textbf{Depth}}   & &\multicolumn{2}{l}{\textbf{Cam. params.}} \\
    \cline{5-9} \cline{11-15} \cline{17-18}
    
   &&  &  & \multicolumn{1}{c}{\textbf{Resolution}} & \multicolumn{1}{p{7.5em}}{\textbf{\# Scenes}} & \multicolumn{1}{p{5.335em}}{\textbf{\# Views per scene}} & \multicolumn{1}{p{5.585em}}{\textbf{\# Tr. scenes}} & \multicolumn{1}{p{5.335em}}{\textbf{\# Ts. scenes}} & &\multicolumn{1}{l}{\textbf{Resolution}} & \multicolumn{1}{p{7em}}{\textbf{\#GT frames}} & \textbf{Type} & \multicolumn{1}{p{5em}}{\textbf{Depth range}} & \multicolumn{1}{p{5em}}{\textbf{Disparity range}} & & \multicolumn{1}{l}{\textbf{Int.}} & \multicolumn{1}{l}{\textbf{Ext.}} \\

    \toprule
    
        {Make3D~\cite{saxena2009make3d}} & \multicolumn{1}{c}{2009} & \multicolumn{1}{p{7em}}{Real} & \multicolumn{1}{p{11.915em}}{Monocular depth} & \multicolumn{1}{c}{$2272\times 1704$} & \multicolumn{1}{p{7.5em}}{$534$} & \multicolumn{1}{p{5.335em}}{monocular} & \multicolumn{1}{p{5.585em}}{$400$} & \multicolumn{1}{c}{$134$} & & \multicolumn{1}{l}{$78\times 51$} & \multicolumn{1}{l}{$534$} & \multicolumn{1}{l}{Dense} & $-$ & $-$ & & \multicolumn{1}{r}{} & \multicolumn{1}{r}{} \\

     \hline   
     {KITTI2012~\cite{geiger2012we} } & \multicolumn{1}{c}{2012} & \multicolumn{1}{p{7em}}{Real} & \multicolumn{1}{p{11.915em}}{Stereo} & \multicolumn{1}{c}{$1240\times376$} & \multicolumn{1}{p{7.5em}}{$389$} & \multicolumn{1}{p{5.335em}}{$2$} & \multicolumn{1}{p{5.585em}}{$194$} & \multicolumn{1}{c}{$195$} &&  \multicolumn{1}{l}{$1226\times370$} & \multicolumn{1}{l}{$-$} & Sparse & $-$ & $-$ && \multicolumn{1}{l}{Y} & \multicolumn{1}{l}{Y} \\
    
    \hline
      {MPI Sintel~\cite{butler2012naturalistic} } & \multicolumn{1}{c}{2012} & \multicolumn{1}{p{7em}}{Synthetic} & \multicolumn{1}{p{11.915em}}{Optical flow} & \multicolumn{1}{c}{$1024 \times 436$} & \multicolumn{1}{p{7.5em}}{$35$ videos} & \multicolumn{1}{p{5.335em}}{$50$ } & \multicolumn{1}{p{5.585em}}{$23$ videos} & \multicolumn{1}{p{4.335em}}{$12$ videos} & & \multicolumn{1}{l}{$-$} & \multicolumn{1}{l}{$-$} & Dense & $-$ & $-$ & & \multicolumn{1}{r}{} & \multicolumn{1}{r}{} \\

    \hline

   {NYU2~\cite{silberman2012indoor}} & \multicolumn{1}{c}{2012} & \multicolumn{1}{p{7em}}{Real - indoor} & \multicolumn{1}{p{11.915em}}{Monocular depth, object segmentation} & \multicolumn{1}{c}{$640\times 480$ } & \multicolumn{1}{p{7.5em}}{$464$ videos, 100$+$  fr. per video} & \multicolumn{1}{p{5.335em}}{ monocular} & \multicolumn{1}{p{5.585em}}{$-$} & \multicolumn{1}{c}{$-$} & & \multicolumn{1}{l}{$-$} & \multicolumn{1}{l}{$1,449$} & Kinect depth & $-$ & $-$ & & \multicolumn{1}{l}{N} & \multicolumn{1}{l}{N} \\
    \hline

   {RGB-D SLAM~\cite{sturm2012benchmark}} & \multicolumn{1}{c}{2012} & \multicolumn{1}{p{7em}}{Real} & \multicolumn{1}{p{11.915em}}{SLAM} & \multicolumn{1}{c}{$640\times 480$} & \multicolumn{1}{p{7.5em}}{$19$ videos} & \multicolumn{1}{c}{} & \multicolumn{1}{p{5.585em}}{$15$ videos} & \multicolumn{1}{c}{$4$ videos} & &  \multicolumn{1}{l}{$-$} & \multicolumn{1}{l}{$-$} & Dense & $-$ & $-$ & & \multicolumn{1}{l}{Y} & \multicolumn{1}{l}{Y} \\
       \hline

    {SUN3D~\cite{xiao2013sun3d}} & \multicolumn{1}{c}{2013} & \multicolumn{1}{p{7em}}{Real - rooms} & \multicolumn{1}{p{11.915em}}{Monocular video} & \multicolumn{1}{c}{$640\times 480$} & \multicolumn{1}{p{7.5em}}{$415$ videos, 10$-$1000$+$ fr. per video} & \multicolumn{1}{p{5.335em}}{$-$} & \multicolumn{1}{p{5.585em}}{$-$} & \multicolumn{1}{c}{$-$} & & \multicolumn{1}{l}{$-$} & \multicolumn{1}{l}{$-$} & Dense, SfM & $-$ & $-$ & & \multicolumn{1}{r}{} & \multicolumn{1}{l}{Y} \\ 
    
       \hline
    Middleburry~\cite{scharstein2014high} & 2014  &  \multicolumn{1}{p{7em}}{Indoor} &  \multicolumn{1}{p{8em}}{Stereo} & $2948\times1988$ & \multicolumn{1}{p{7.5em}}{$30$} &  \multicolumn{1}{p{5.335em}}{$2$} &  \multicolumn{1}{p{5.585em}}{$15$} &  {$15$}  & & \multicolumn{1}{l}{$2948\times1988$} & \multicolumn{1}{l}{$30$} & Dense & $-$    & $260$ & &  \multicolumn{1}{l}{Y}     &  \multicolumn{1}{l}{Y} \\

    \hline

    {KITTI 2015~\cite{menze2015object}} & \multicolumn{1}{c}{2015} & \multicolumn{1}{p{7em}}{Real} & \multicolumn{1}{p{11.915em}}{Stereo} & \multicolumn{1}{c}{$1242\times375$} & \multicolumn{1}{p{7.5em}}{$400$} & \multicolumn{1}{p{5.335em}}{$4$} & \multicolumn{1}{p{5.585em}}{$200$} & \multicolumn{1}{c}{$200$}  & & \multicolumn{1}{l}{$1242\times375$} & \multicolumn{1}{l}{$-$} & Sparse & $-$ & $-$ && \multicolumn{1}{l}{Y} & \multicolumn{1}{l}{Y} \\
    \hline

   {KITTI-MVS2015~\cite{menze2015object}} & \multicolumn{1}{c}{2015} & \multicolumn{1}{p{7em}}{Real} & \multicolumn{1}{p{11.915em}}{MVS} & \multicolumn{1}{c}{$1242\times375$} & \multicolumn{1}{p{7.5em}}{$400$} & \multicolumn{1}{p{5.335em}}{$20$} & \multicolumn{1}{p{5.585em}}{$200$} & \multicolumn{1}{c}{$200$} & & \multicolumn{1}{l}{$-$} & \multicolumn{1}{l}{$-$} & Sparse  & $-$ & $-$ & &\multicolumn{1}{l}{Y} & \multicolumn{1}{l}{Y} \\
    \hline
    

   FlyingThings3D, Monkaa, Driving~\cite{mayer2016large} & \multicolumn{1}{c}{2016} & \multicolumn{1}{p{7em}}{Synthetic} & \multicolumn{1}{p{11.915em}}{Stereo, Video, Optical flow} & \multicolumn{1}{c}{$960 \times 540$} & \multicolumn{1}{p{7.5em}}{$39$K frames} & \multicolumn{1}{p{5.335em}}{$2$} & \multicolumn{1}{p{5.585em}}{$21,818$} & \multicolumn{1}{c} {$4,248$} &  & \multicolumn{1}{l}{$384\times192$} & \multicolumn{1}{l}{$-$} & Dense & $-$ & $160$px & & \multicolumn{1}{l}{Y} & \multicolumn{1}{l}{Y} \\

   \hline
 {CityScapes~\cite{cordts2016cityscapes}} & \multicolumn{1}{c}{2016} & \multicolumn{1}{p{7em}}{Street scenes} & \multicolumn{1}{p{11.915em}}{Semantic seg., dense  labels} & \multicolumn{1}{c}{$2048\times 1024$} & \multicolumn{1}{p{7.5em}}{$5$K} & \multicolumn{1}{p{5.335em}}{$2$} & \multicolumn{1}{p{5.585em}}{$2975$} & \multicolumn{1}{c}{$1525$} & & \multicolumn{1}{l}{$-$} & \multicolumn{1}{l}{$-$} & NA    & $-$ & $-$ & & \multicolumn{2}{l}{Ego-motion} \\
	\cline{4-18}
   &  &  & \multicolumn{1}{p{11.915em}}{Semantic seg.,coarse labels} & $2048\times1024$  & \multicolumn{1}{p{7.5em}}{$20$K} & \multicolumn{1}{p{5.335em}}{$2$} & \multicolumn{1}{c}{} & \multicolumn{1}{c}{} & & \multicolumn{1}{l}{NA} & \multicolumn{1}{l}{NA} & NA    & NA    & NA    && \multicolumn{2}{l}{Ego-motion} \\
 
    \hline

   {DTU~\cite{aanaes2016large} } & \multicolumn{1}{c}{2016} & \multicolumn{1}{p{7em}}{Real, small objects} & \multicolumn{1}{p{11.915em}}{MVS} & \multicolumn{1}{c}{$1200\times 1600$} & \multicolumn{1}{p{7.5em}}{$80$} & \multicolumn{1}{p{5.335em}}{$49-64$} & \multicolumn{1}{p{5.585em}}{$-$} & \multicolumn{1}{c}{$-$} & &  \multicolumn{1}{l}{$-$} & \multicolumn{1}{l}{$-$} & \multicolumn{1}{p{5.585em}}{Structured light scans} & $-$ & $-$ & & \multicolumn{1}{l}{Y} & \multicolumn{1}{l}{Y} \\
    \hline

   {ETH3D~\cite{schops2017multi}} & \multicolumn{1}{c}{2017} & \multicolumn{1}{p{7em}}{Real, in/outdoor} & \multicolumn{1}{p{11.915em}}{Low-res, Stereo} & \multicolumn{1}{c}{ $940\times490$} & \multicolumn{1}{p{7.5em}}{$47$} & \multicolumn{1}{p{5.335em}}{$2$} & \multicolumn{1}{p{5.585em}}{$27$} & \multicolumn{1}{c}{$20$} &  & \multicolumn{1}{l}{$-$} & \multicolumn{1}{l}{$47$} & \multicolumn{1}{l}{Dense} & $-$ & $-$ &  &\multicolumn{1}{l}{Y} & \multicolumn{1}{l}{Y} \\
    \cline{4-18}

      &   &   & \multicolumn{1}{p{11.915em}}{Low-res, MVS on video} & \multicolumn{1}{c}{ $940\times490$} & \multicolumn{1}{p{7.5em}}{$10$ videos} & \multicolumn{1}{p{5.335em}}{$4$} & \multicolumn{1}{p{5.585em}}{$5$ videos} & \multicolumn{1}{c}{$5$ videos} & & \multicolumn{1}{l}{$-$} & \multicolumn{1}{l}{$-$} & \multicolumn{1}{l}{Dense} & $-$ & $-$ & & \multicolumn{1}{l}{Y} & \multicolumn{1}{l}{Y} \\
    \cline{4-18}
    
    &  & & \multicolumn{1}{p{11.915em}}{High-res, MVS on images from DSLR camera} & \multicolumn{1}{c}{ $940\times490$} & \multicolumn{1}{p{7.5em}}{$25$} & \multicolumn{1}{p{5.335em}}{$14-76$} & \multicolumn{1}{p{5.585em}}{$13$} & \multicolumn{1}{c}{$12$} & & \multicolumn{1}{l}{$-$} & \multicolumn{1}{l}{$25$} & \multicolumn{1}{l}{Dense} & $-$ & $-$ & & \multicolumn{1}{l}{Y} & \multicolumn{1}{l}{Y} \\
    \hline
    
        {SUNCG~\cite{song2017semantic}} & \multicolumn{1}{c}{2017} & \multicolumn{1}{p{7em}}{Synthetic, indoor} & \multicolumn{1}{p{11.915em}}{Scene completion} & \multicolumn{1}{c}{$-$} & \multicolumn{1}{p{7.5em}}{$45$K} & \multicolumn{1}{p{5.335em}}{$-$} & \multicolumn{1}{p{5.585em}}{$-$} & \multicolumn{1}{c}{$-$} &  & \multicolumn{1}{l}{$640\times 480$} & \multicolumn{1}{l}{$-$} &   \multicolumn{1}{p{2cm}}{Depth and Vol. GT} & $-$ & $-$ & &\multicolumn{1}{r}{} & \multicolumn{1}{r}{} \\
    \hline

       {MVS-Synth~\cite{huang2018deepmvs}} & \multicolumn{1}{c}{2018} & \multicolumn{1}{p{7em}}{Synth - urban} & \multicolumn{1}{p{11.915em}}{MVS} & \multicolumn{1}{c}{$1920\times1080$ } & \multicolumn{1}{p{7.5em}}{$120$} & \multicolumn{1}{p{5.335em}}{$100$} & \multicolumn{1}{c}{} & \multicolumn{1}{c}{} & &\multicolumn{1}{l}{$-$}  & \multicolumn{1}{l}{$-$} & Dense & $-$ & $-$ & &\multicolumn{1}{l}{Y} & \multicolumn{1}{l}{Y} \\
    \hline

    {MegaDepth~\cite{Li_2018_CVPR} } & \multicolumn{1}{c}{2018} & \multicolumn{1}{p{7em}}{Real (Internet images)} & \multicolumn{1}{p{11.915em}}{Monocular,  Eucl. and ord. depth} & \multicolumn{1}{c}{$1600\times 1600$} & \multicolumn{1}{p{7.5em}}{$130$K} & \multicolumn{1}{p{5.335em}}{monocular} & \multicolumn{1}{p{5.585em}}{$-$} & \multicolumn{1}{c}{$-$} & & \multicolumn{1}{l}{$-$} &  \multicolumn{1}{p{6em}}{$100$K (Eucl.), $30$K (Ord.)} &   \multicolumn{1}{p{2cm}}{Dense, Eucl., Ord.} & $-$ & $-$ & & \multicolumn{1}{r}{} & \multicolumn{1}{r}{} \\
 
     \hline
  
  Jeon and Lee~\cite{Jeon_2018_ECCV}  & $2018$ &  \multicolumn{1}{p{7em}}{Real} &  \multicolumn{1}{p{7em}}{Depth enhancement} &   $-$ &  \multicolumn{1}{p{7.5em}}{$4$K images} &  \multicolumn{1}{l}{$-$} & \multicolumn{1}{l}{$-$} & $-$  & & \multicolumn{1}{l}{$640\times 480$}  & $4,000$ & Dense & $0.01 - 30$m  & $-$ & &  \multicolumn{1}{l}{Y} &  \multicolumn{1}{l}{Y} \\

    \hline 
    {OmniThings~\cite{Won_2019_ICCV,Won_2020_PAMI}} & \multicolumn{1}{c}{2019} & \multicolumn{1}{p{7em}}{Synthetic, fisheye images} & \multicolumn{1}{p{11.915em}}{Omnidirectional MVS} & $800\times768$ & \multicolumn{1}{p{7.5em}}{$10240$} & \multicolumn{1}{p{5.335em}}{$4$} & \multicolumn{1}{p{5.585em}}{$9216$} & \multicolumn{1}{c}{$1024$} & & \multicolumn{1}{l}{$640\times 320$} & \multicolumn{1}{l}{$-$} & Dense & $-$ & $\le192$px & &  \multicolumn{1}{r}{} & \multicolumn{1}{r}{} \\
    \hline
  
   {OmniHouse~\cite{Won_2019_ICCV,Won_2020_PAMI}} & \multicolumn{1}{c}{2019} & \multicolumn{1}{p{7em}}{Synthetic, fisheye images} & \multicolumn{1}{p{11.915em}}{Omnidirectional MVS} & $800\times768$ & \multicolumn{1}{p{7.5em}}{$2,560$} & \multicolumn{1}{p{5.335em}}{$4$} & \multicolumn{1}{p{5.585em}}{$2048$} & \multicolumn{1}{c}{$512$} & & \multicolumn{1}{l}{$640\times 320$} & \multicolumn{1}{l}{$-$} & Dense & $-$ & $\le192$px &  &\multicolumn{1}{r}{} & \multicolumn{1}{r}{} \\
    \hline

   {HR-VS~\cite{Yang_2019_CVPR}} & \multicolumn{1}{c}{2019} & \multicolumn{1}{p{7em}}{Synthetic, outdoor} & \multicolumn{1}{p{11.915em}}{High res. stereo} & \multicolumn{1}{c}{$2056\times2464$} & \multicolumn{1}{p{7.5em}}{$780$} & \multicolumn{1}{p{5.335em}}{$2$} & \multicolumn{1}{p{5.585em}}{$-$} & \multicolumn{1}{c}{$-$} & &  \multicolumn{1}{l}{$1918\times2424$} & \multicolumn{1}{l}{780} & Dense, Eucl. & $2.52$ to $200$m & $9.66$ to $768$px &  & \multicolumn{1}{l}{} & \multicolumn{1}{l}{} \\
   \cline{3-18}
  
     &  & \multicolumn{1}{p{7em}}{Real, outdoor} & \multicolumn{1}{p{11.915em}}{High res. stereo} & \multicolumn{1}{c}{$1918\times2424$} & \multicolumn{1}{l}{$33$} & \multicolumn{1}{p{5.335em}}{$2$} & \multicolumn{1}{l}{$-$} &  \multicolumn{1}{c}{$-$} & &\multicolumn{1}{l}{$1918\times2424$} & \multicolumn{1}{l}{33} & Dense, Eucl. &       & $5.41$ to $182.3$px & & \multicolumn{1}{l}{} & \multicolumn{1}{l}{} \\
    \hline
  
  {DrivingStereo~\cite{yang2019drivingstereo}} & \multicolumn{1}{c}{2019} & \multicolumn{1}{p{7em}}{Driving} & \multicolumn{1}{p{11.915em}}{High res. stereo} & \multicolumn{1}{c}{$1762\times800$} & \multicolumn{1}{p{7.5em}}{$182,188$} & \multicolumn{1}{p{5.335em}}{$2$} & \multicolumn{1}{p{5.585em}}{$174,437$} & \multicolumn{1}{c}{$7,751$} & & \multicolumn{1}{l}{$1762\times800$} & \multicolumn{1}{l}{$182,188$} & Sparse & up to $80$m &     &    & \multicolumn{1}{l}{} & \multicolumn{1}{l}{} \\
  
  \hline
  ApolloScape~\cite{wang2019apolloscape} & 2019 & \multicolumn{1}{p{7em}}{Auto. driving}   & \multicolumn{1}{p{11.915em}}{High res. stereo} & \multicolumn{1}{c}{$3130\times960$} &   \multicolumn{1}{p{7.5em}}{$5,165$} &  \multicolumn{1}{p{5.335em}}{$2$} & \multicolumn{1}{p{5.585em}}{$4,156$} &  \multicolumn{1}{c}{$1,009$}  & &  
   \multicolumn{1}{l}{$-$} & \multicolumn{1}{l}{$5165$} & LIDAR & $-$ to $-$m &   $-$   &    & \multicolumn{1}{l}{Y} & \multicolumn{1}{l}{$-$}  \\
  \hline
  A2D2~\cite{geyer2020a2d2} & 2020 & \multicolumn{1}{p{7em}}{Auto. driving}   & \multicolumn{1}{p{11.915em}}{High res. stereo} & \multicolumn{1}{c}{$2.3$M pixel} &   \multicolumn{1}{p{7.5em}}{$41,277$} &  \multicolumn{1}{p{5.335em}}{$6$} & \multicolumn{1}{p{5.585em}}{$-$} &  \multicolumn{1}{c}{$-$}  & &  
   \multicolumn{1}{l}{$-$} & \multicolumn{1}{l}{$-$} & LIDAR & up to $100$m &   $-$   &    & \multicolumn{1}{l}{Y} & \multicolumn{1}{l}{Y}  \\
 \bottomrule
 \end{tabular}%
		}
\end{table*}	

Table~\ref{tab:3ddatasets_sterep} summarizes some of the datasets that have been used to train and test deep learning-based depth estimation algorithms. Below, we discuss these datasets based on their sizes, their spatial and depth resolution, the type of depth annotation they provide, and the domain gap (or shift) issue faced by many deep learning-based algorithms.  


\vspace{6pt}
\noi\textit{(1) Dataset size. }  The first datasets, which appeared prior to $2016$, are of small scale due to the difficulty of creating ground-truth 3D annotations. An example is the two KITTI datasets~\cite{geiger2012we,menze2015object}, which contain $200$ stereo pairs with their corresponding disparity ground-truth.   They have been extensively used to train and test patch-based CNNs for stereo matching algorithms (see Section~\ref{sec:depth_by_stereo_matching}), which have a small receptive field. As such a  single stereo pair can result in thousands of training samples.  However, in end-to-end architectures (Sections~\ref{sec:end_to_end_stereo} and~\ref{sec:mvs_architectures}), a stereo pair  corresponds to only one sample.  End-to-end networks have a large number of parameters, and thus require large datasets for efficient training.  While collecting large image datasets is very easy, \eg by using video sequences as in \eg NYU2~\cite{silberman2012indoor}, ETH3D~\cite{schops2017multi}, SUN3D~\cite{xiao2013sun3d}, and ETH3D~\cite{schops2017multi}, annotating them with 3D labels is time consuming.  Recent works, \eg the AppoloScape~\cite{wang2019apolloscape} and A2D2~\cite{geyer2020a2d2}, use LIDAR to  acquire dense 3D annotations.

Data augmentation strategies, \eg by applying  geometric and photometric transformations to the images that are available,  have been extensively used in the literature.  There are, however, a few other strategies that  are specific to depth estimation. This includes artificially  synthesizing and rendering  from  3D CAD models 2D and 2.5D views from various (random) viewpoints, poses, and lighting conditions. One can also overlay  rendered 3D models on the  top of real images. This approach has been used to generate the FlyingThings3D, Monkaa, and Driving datasets of~\cite{mayer2016large}, and the OmniThings and OmniHouse datasets for benchmarking MVS for omnidirectional images~\cite{Won_2019_ICCV,Won_2020_PAMI}. Huang \etal~\cite{huang2018deepmvs} followed a similar idea but used scenes from video games to generate MVS-Synth, a photo-realistic synthetic dataset prepared for learning-based Multi-View Stereo algorithms.  

The main challenge  is that generating large amounts of synthetic data containing varied real-world appearance and motion is not trivial~\cite{mayer2018makes}.  As a result, a number of works overcome the need for ground-truth depth information by training their deep networks without 3D supervision, see Section~\ref{sec:end2enddegree_supervision}.  Others used traditional depth estimation and structure-from-motion (SfM) techniques to generate 3D annotations. For example, Li \etal~\cite{Li_2018_CVPR} used modern structure-from-motion and multiview stereo (MVS) methods together with multiview Internet photo collections to create the large-scale MegaDepth dataset providing improved depth estimation accuracy via bigger training dataset sizes. This dataset has  also been automatically augmented with ordinal depth relations generated using semantic segmentation.

\vspace{6pt}
\noi\textit{(2) Spatial and depth resolutions. } The disparity/depth information can be either in the form of maps of the same or lower resolution than the input images, or in the form of sparse depth values at some locations in the reference image. Most of the existing datasets are of low spatial resolution. In recent years, however, there has been a growing focus on stereo matching with high-resolution images. An example of a high-resolution dataset is the HR-VS and HR-RS of Yang \etal~\cite{Yang_2019_CVPR}, where each RGB pair of resolution $1918 \times 2424$ is annotated with a depth map of the same resolution. However, the dataset  only contains $800$ pairs of stereo images, which is relatively small for end-to-end training. Other datasets such as  the  ApolloScape~\cite{wang2019apolloscape} and   A2D2~\cite{geyer2020a2d2} contain very high resolution images, of the order of $3130\times960$, with more that $100+$ hours of stereo driving videos, in the case of ApolloScape, have been specifically designed to test autonomous driving algorithms.

\vspace{6pt}
\noi\textit{(3) Euclidean vs. ordinal depth. } Instead of manually annotating images with exact, \ie Euclidean, depth values, some papers, \eg MegaDepth~\cite{Li_2018_CVPR}, provide ordinal annotations, \ie pixel $\pixel_1$ is closer, farther, or at the same depth, as pixel $\pixel_2$. Ordinal annotation is  simpler and faster to achieve than Euclidean annotation. In fact, it can be accurately obtained using traditional stereo matching algorithms, since ordinal depth is less sensitive to innacuracies in depth estimation

\vspace{6pt}
\noi\textit{(4) Domain gap. }  While artificially augmenting training datasets allows enriching existing ones, the  domain shift caused by the very different conditions between real and synthetic data can result in a lower accuracy when applied to real-world environments. We will discuss, in Section~\ref{sec:domain_adaptation}, how this domain shift issue has been addressed in the literature.

\section{Depth by stereo matching}
\label{sec:depth_by_stereo_matching}\label{sec:stereo_pipeline}

\begin{table*}
	\caption{\label{tab:taxonomy_stereomatching} Taxonomy and comparison of deep learning-based stereo matching techniques. }
	
	\resizebox{\linewidth}{!}{%
	\begin{tabular}{@{}l@{ }c@{ }c@{ }c@{ }c@{ }c@{ }c@{ }c@{ }c@{ }c@{ }c@{}}
	\toprule
	  \multirow{2}{*}{\textbf{Method}}  & \multirow{2}{*}{\textbf{Year}}   & \multicolumn{2}{c}{\textbf{Feature computation}} & &\multirow{2}{*}{\textbf{Similarity}} & & \multicolumn{2}{c}{\textbf{Training}} &  &\multirow{2}{*}{\textbf{Regularization}}\\
	  
	  \cline{3-4}   \cline{8-9}  
	  & &\textbf{Architectures} & \textbf{Dimension} & & & & \textbf{Degree of supervision} & \textbf{Loss} & &  \\
	  \midrule
	  
	Zagoruyko~\cite{zagoruyko2015learning} & $2015$ & ConvNet & Multiscale & &  FCN & &  Supervised with positive/negative samples & Hinge  and squared
$\ltwo$  & &  NA\\
	\hline
	Han~\cite{han2015matchnet}&$2015$& ConvNet & Fixed scale & &  FCN & & Supervised & Cross-entropy& &NA \\
	\hline
	Zbontar~\cite{zbontar2015computing}&$2015$ & ConvNet &  Fixed scale & & Hand-crafted & & Triplet contrastive learning  & $\lone$ & & MRF\\
	\hline
	Chen~\cite{chen2015deep}&$2015$&  ConvNet& Multiscale & & Correlation $+$ voting& &  Supervised with positive/negative samples &$\lone$ & & MRF\\
	\hline
	Simo~\cite{simo2015discriminative}&$2015$& ConvNet & Fixed scale& &  $\ltwo$  & & Supervised with positive/negative samples & $\ltwo$ & &  NA\\
	\hline
	Zbontar~\cite{zbontar2016stereo}&$2016$& ConvNet &  Fixed scale& &  Hand-crafted, FCN & & Supervised with known disparity & Hinge & & Classic stereo\\
	\hline
	Balantas~\cite{balntas2016pn}&$2016$& ConvNet	& Fixe scale & & $\ltwo$ & & Supervised, triplet contrastive learning  & Soft-Positive-Negative (Soft-PN)& & $-$\\	
	\hline
	Mayer~\cite{mayer2016large}&$2016$& ConvNet& Fixed-scale& & Hand-crafted  & & Supervised & $-$& & Encoder-decoder \\ 
	\hline
	Luo~\cite{luo2016efficient}&$2016$& ConvNet & Fixed scale & &  Correlation & & Supervised & Cross-entropy& & MRF\\
	\hline
	
	Kumar~\cite{kumar2016learning}& 2016 & ConvNet & Fixed scale  &    & ConvNet &  & Supervised, triplet contrastive learning  &  Maximise inter-class distance, & & $-$ \\
	 							 &   &   &      &    &   &  &  & minimize inter-class distance.& & \\
	\hline
	Shaked~\cite{shaked2017improved}&$2017$&  Highway network with &  Fixed scale & & FCN & & Supervised & Hinge$+$cross-entropy& & Classic$+$4Conv$+$\\
									& 	& multilevel skip connections & && &&& & & 5FC \\
	\hline

	Hartmann~\cite{hartmann2017learned}&$2017$& ConvNet & Fixed scale&  &  ConvNet& &Supervised & Croos-entropy& & Encoder \\
	\hline

	Park~\cite{park2017look}&$2017$& ConvNet & Fixed scale & & $1\times1$ Convs,   & & Supervised & $-$ & & NA \\
						& & & & &   ReLU, SPP & & & & & \\	
	\hline

	Ye~\cite{ye2017efficient}&$2017$& ConvNet &  Fixed scale & &  FCN    & & Supervised& $\lone$ & & SGM\\
						      & & Multisize pooling &    & &    ($1\times1$ convs) & & &  & & \\
	\hline

	Tulyakov~\cite{tulyakov2017weakly}&$2017$&  \multicolumn{4}{c}{Generic - independent of the network architecture}  & & Weakly supervised & MIL, Contrastive, Contrastive-DP& & $-$\\

	\bottomrule	
	\end{tabular}
	}
\end{table*}

Stereo-based depth reconstruction methods take $\nimages=2$ RGB images  and produce  a disparity map  $\disparitymap$ that minimizes an energy function of the form:
\begin{equation}
	\energy(\depthmap) = \sum_{\pixel} \costvolume(\pixel, \depth_\pixel) + \sum_\pixel\sum_{y\in \neighborhood_\pixel} \smoothnessenergy(\depth_\pixel, \depth_y).
	\label{eq:stereomatching_energy}
\end{equation}

\noi Here, $\pixel$ and $y$ are image pixels, and $\neighborhood_\pixel$ is the set of pixels that are within the neighborhood of $\pixel$. The first term of Eqn.~\eqref{eq:stereomatching_energy} is the matching cost. When using rectified stereo pairs, $ \costvolume(\pixel, \depth_\pixel) $ measures the cost of matching the pixel $\pixel =(i, j)$ of the left image with the pixel $y=(i, j-\depth_{\pixel})$ of the right image.   In this case, $\depth_{\pixel} = \disparitymap(\pixel) \in [\disparity_{min}, \disparity_{max}]$ is  the disparity at pixel $\pixel$. Depth can then be inferred by triangulation.  When the disparity range is discritized into $\ndisparities$ disparity levels, $\costvolume$ becomes a 3D cost volume of size $\width \times \height \times \ndisparities$.  In the more general multiview stereo case, \ie $\nimages \ge 2$, the cost $\costvolume(\pixel, \depth_\pixel)$ measures the inverse likelihood of $\pixel$ on the reference image having depth $\depth_\pixel$.   The second term of Eqn.~\eqref{eq:stereomatching_energy} is a regularization term used to impose  constraints  such as   smoothness and left-right  consistency.

\begin{figure}
	\includegraphics[trim=0cm 7.0cm 2cm 4.5cm, clip=true, width=.5\textwidth]{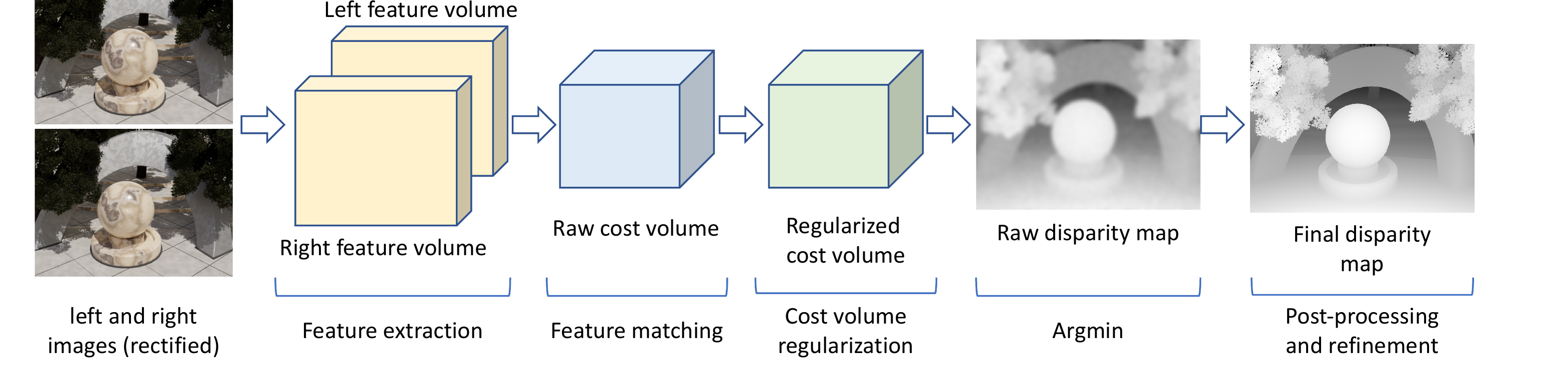}
	\caption{\label{fig:stereo_matching_pipeline} The building blocs of a stereo matching pipeline. }
\end{figure}

Traditionally, this problem has been solved using a pipeline of  four building blocks~\cite{scharstein2002taxonomy}, see Fig.~\ref{fig:stereo_matching_pipeline}:  \textbf{(1)} feature extraction, \textbf{(2)} feature matching across images,  \textbf{(3)} disparity  computation, and \textbf{(4)} disparity refinement and post-processing.  The first two blocks construct the cost volume $\costvolume$.  The third block regularizes the cost volume and then finds, by minimizing Eqn.~\eqref{eq:stereomatching_energy}, an initial estimate of the disparity map. The last block refines and post-processes the initial disparity map.  

This section focuses on how these individual blocks have been implemented using deep learning-based methods. Table~\ref{tab:taxonomy_stereomatching} summarises the state-of-the-art methods. 


\subsection{Learning feature extraction and matching}
\label{sec:learning_feature_extraction}

\begin{figure*}[t]
\centering
\begin{tabular}{@{}cccccc@{}}
	\includegraphics[height=0.2\textheight]{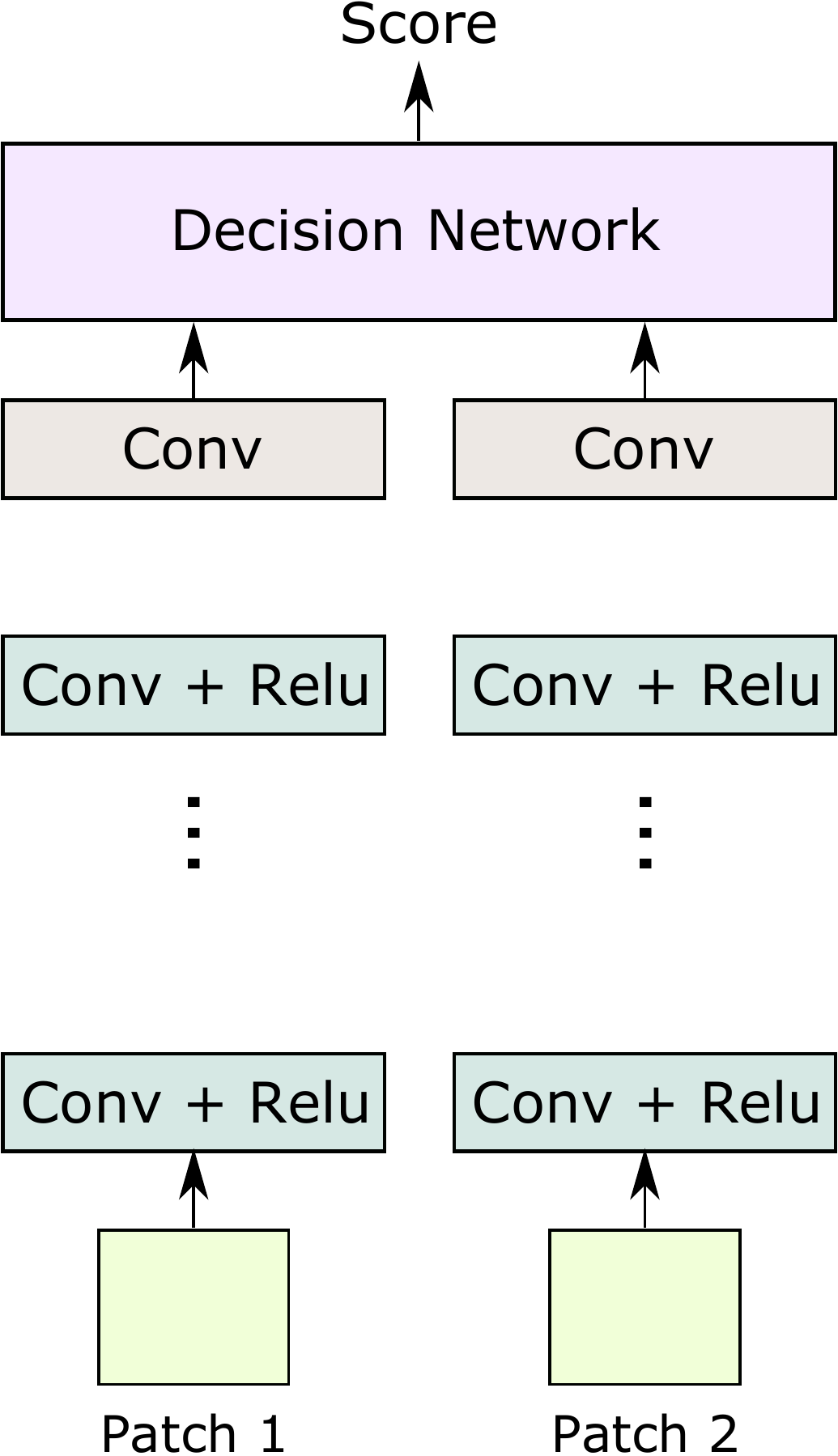}&
	\includegraphics[height=0.2\textheight]{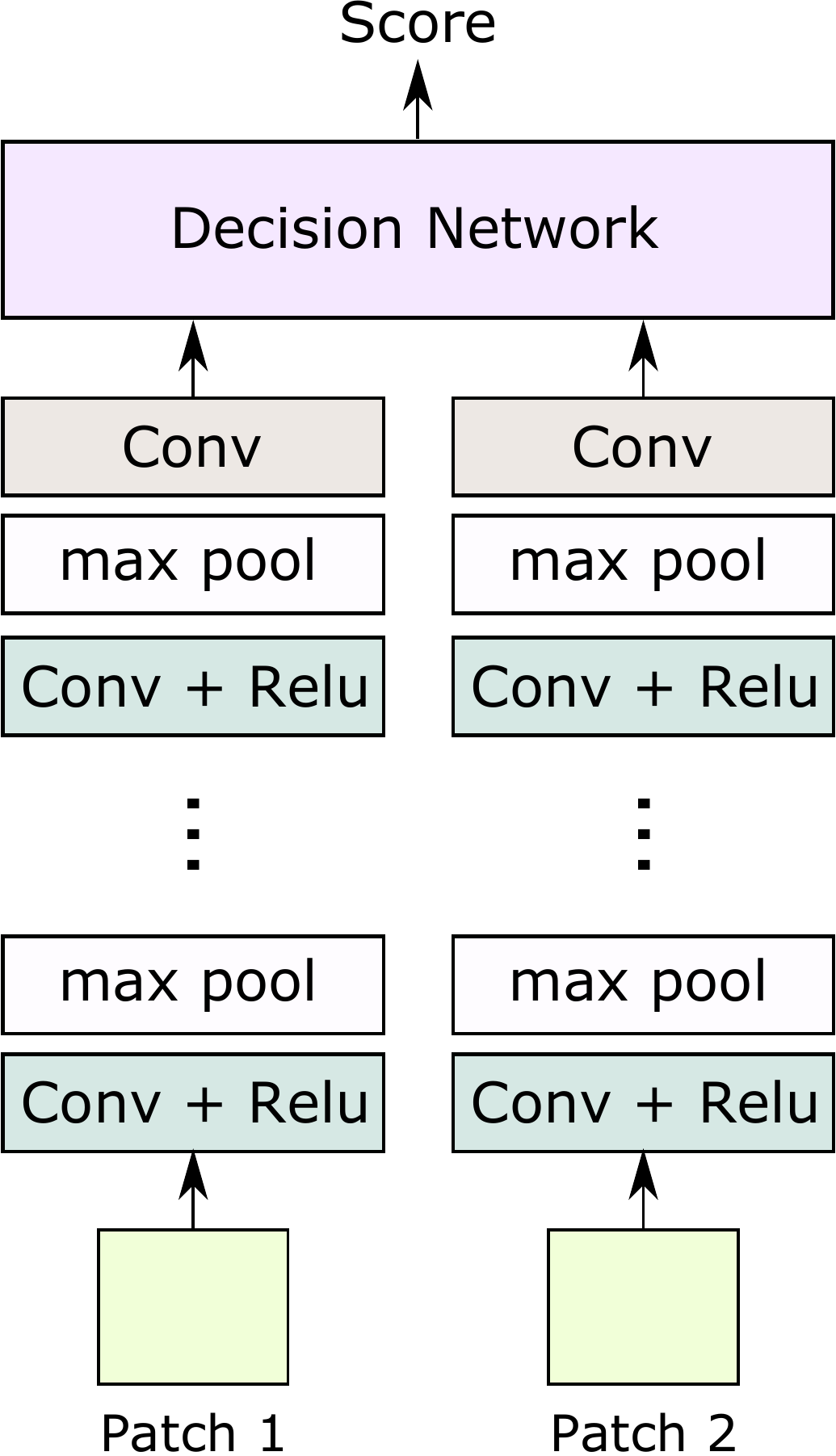}&
	\includegraphics[height=0.2\textheight]{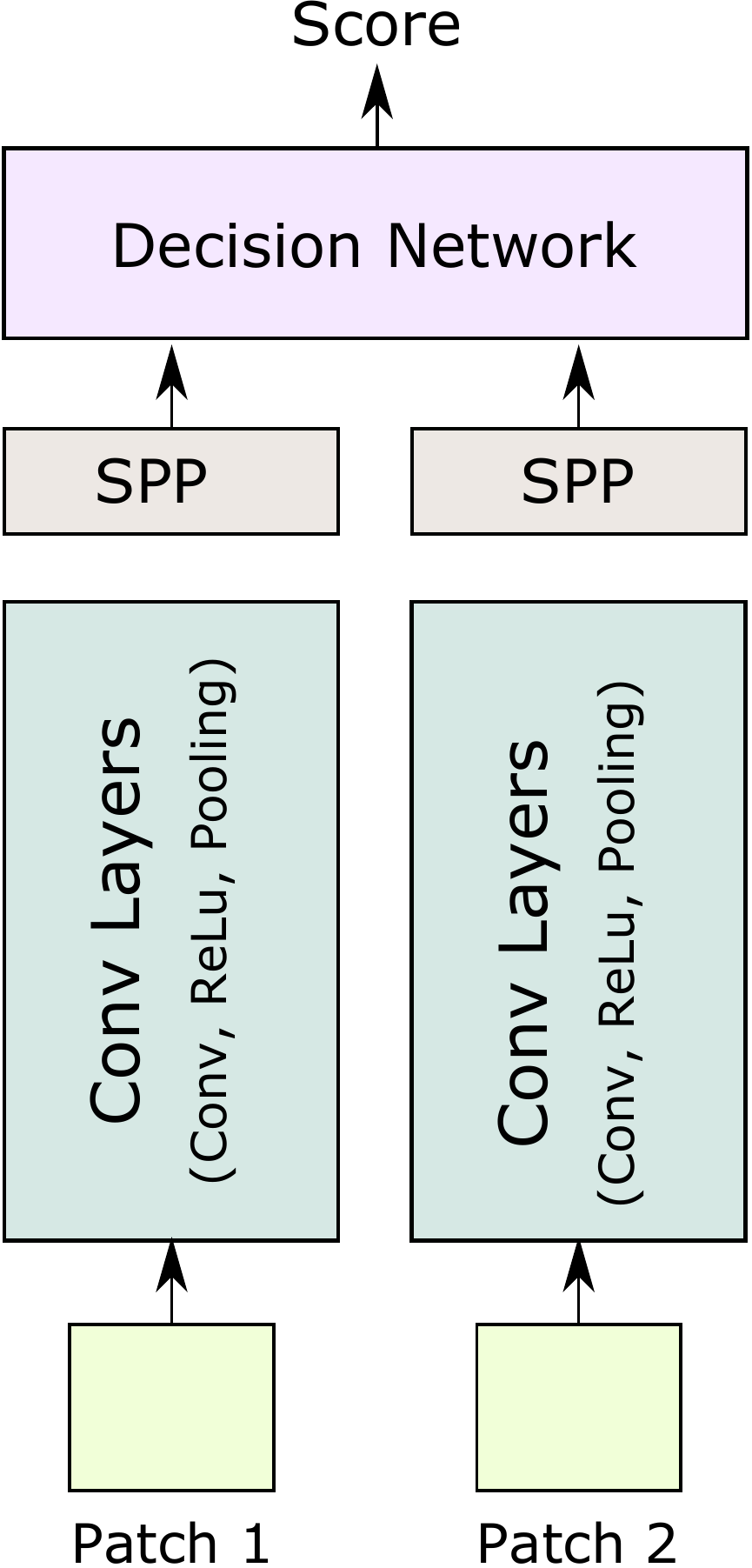}&
	\includegraphics[height=0.2\textheight]{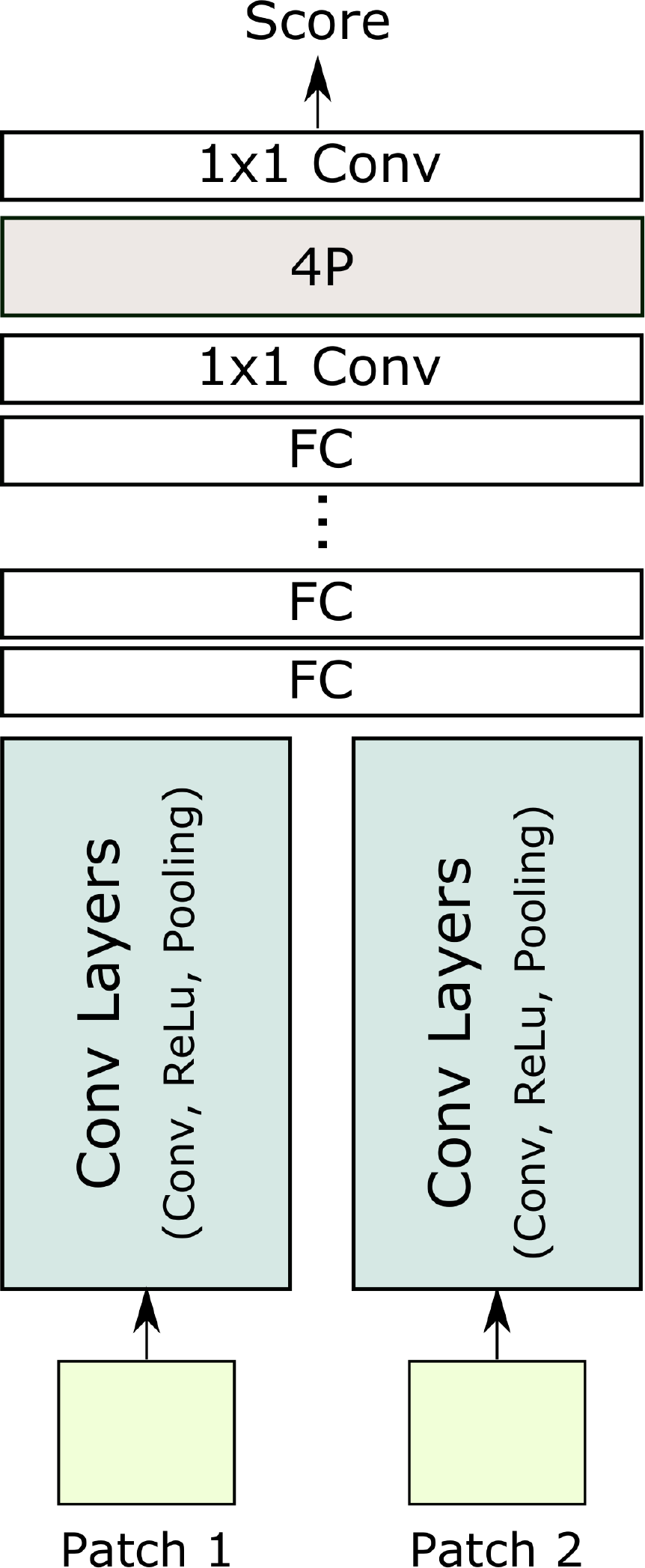}&
	\includegraphics[height=0.2\textheight]{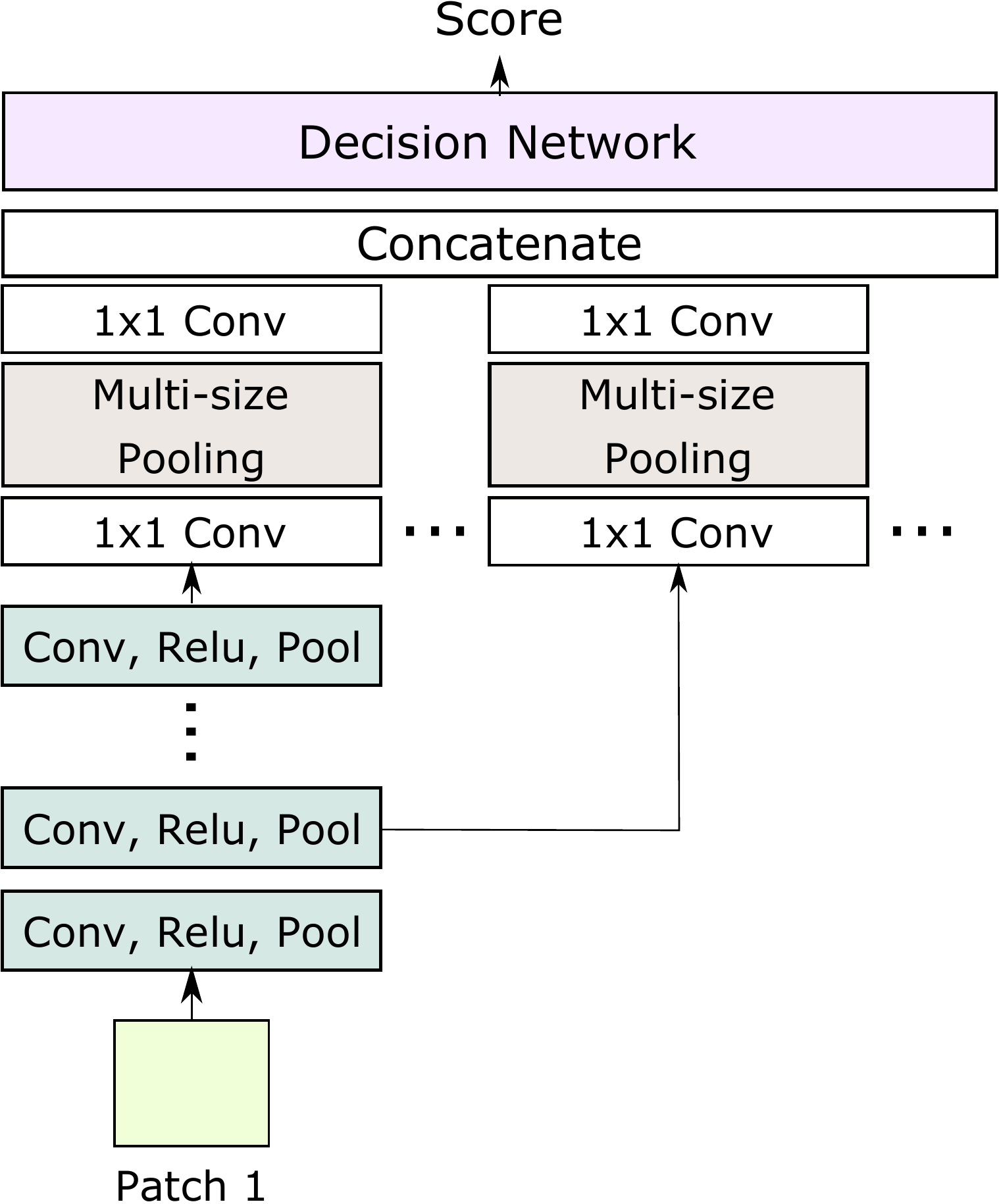}&
	\includegraphics[height=0.2\textheight]{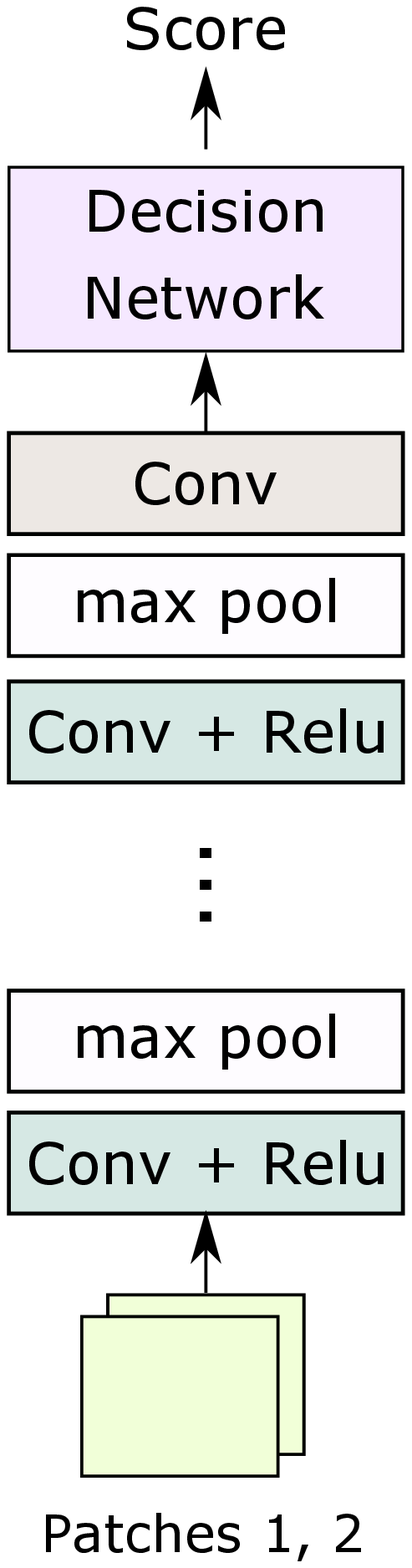} \\
	\footnotesize{(a) MC-CNN~\cite{zbontar2015computing,zbontar2016stereo}.} & \footnotesize{(b)~\cite{zagoruyko2015learning} and~\cite{han2015matchnet}.} & \footnotesize{(c)~\cite{zagoruyko2015learning}.} & \footnotesize{(d) LW-CNN~\cite{park2017look}.} & \footnotesize{(e) FED-D2DRR~\cite{ye2017efficient}.} & \footnotesize{(f)~\cite{zagoruyko2015learning}.} 
\end{tabular}
\caption{\label{fig:feature_learning_architectures}Feature learning architectures.}
\end{figure*}

Early deep learning techniques for stereo matching replace  the hand-crafted features (block A of Fig.~\ref{ig:stereo_matching_pipeline})  with learned features~\cite{zagoruyko2015learning,han2015matchnet,zbontar2015computing,zbontar2016stereo}. They take two patches, one centered at a pixel $\pixel = (i, j)$ on the left image and another one centered at pixel $y = (i, j-\disparity)$ on  the right image (with $\disparity \in \{0, \dots, \ndisparities\}$), compute their corresponding feature vectors using a CNN, and  then match them (block B of Fig.~\ref{ig:stereo_matching_pipeline}), to produce a similarity score $\costvolume(\pixel, \disparity)$, using either standard similarity metrics  such as the $\lone$, the $\ltwo$, and the correlation metric, or metrics learned using a top network. The two components can be trained either separately or jointly.

\subsubsection{The basic network architecture}
\label{sec:baseline_architecture}
The basic network architecture, introduced in~\cite{zagoruyko2015learning,han2015matchnet,zbontar2015computing,zbontar2016stereo} and shown in Fig.~\ref{fig:feature_learning_architectures}-(a), is composed of two CNN encoding branches,  which act as descriptor computation modules.  The first branch takes a patch around a pixel $\pixel = (i, j)$ on the left image and outputs  a feature vector that characterizes that patch. The second branch takes a patch around the pixel $y = (i, j- \disparity)$, where $\disparity \in [\disparity_{min}, \disparity_{max}]$ is a candidate disparity.  Zbontar and LeCun~\cite{zbontar2015computing} and later Zbontar \etal~\cite{zbontar2016stereo} use an encoder composed of four convolutional layers, see Fig.~\ref{fig:feature_learning_architectures}-(a). Each layer, except the last one, is followed by a ReLU unit. Zagoruyko and Komodakis~\cite{zagoruyko2015learning} and Han \etal~\cite{han2015matchnet} use a similar architecture but add:
\begin{itemize}
	\item max-pooling and subsampling after each layer, except the last one, see Fig.~\ref{fig:feature_learning_architectures}-(b). As such, the network is able to account for larger patch sizes and a larger variation in the viewpoint  compared to~\cite{zbontar2015computing,zbontar2016stereo}.
	
	\item a Spatial Pyramid Pooling (SPP) module at the end of each feature extraction branch~\cite{zagoruyko2015learning} so that the network can process patches of arbitrary sizes  while producing features of a fixed size, see Fig.~\ref{fig:feature_learning_architectures}-(c). Its role is to aggregate the features of the last convolutional layer, through spatial pooling, into a feature grid of a fixed size. The module is designed in such a way that the size of the pooling regions varies with the size of the input to ensure that the output feature grid has a fixed size independently of the size of the input patch or image.  Thus, the network is able to process patches/images of arbitrary sizes and compute feature vectors of the same dimension without changing its structure or retraining. 
\end{itemize}

\noi The learned features are then fed to a top module, which returns a similarity score. It can be implemented as a standard similarity metric, \eg the $\ltwo$  distance, the cosine distance, and  the (normalized)  correlation distance (or inner product) as in the MC-CNN-fast (MC-CNN-fst) architecture of~\cite{zbontar2015computing,zbontar2016stereo}. The main advantage of the correlation  over the $\ltwo$ distance is that it can be implemented using a layer of 2D~\cite{dosovitskiy2015flownet}  or 1D~\cite{mayer2016large} convolutional operations, called  \emph{correlation layer}.  A correlation layer does not require training since the filters are in fact the features computed by the second branch of the network.  As such, correlation layers have been extensively used in the literature~\cite{zbontar2015computing,simo2015discriminative,zbontar2016stereo,mayer2016large,luo2016efficient}. 

Instead of using hand-crafted similarity measures, recent works use a decision network composed of fully-connected (FC) layers~\cite{han2015matchnet,zagoruyko2015learning,zbontar2016stereo,shaked2017improved,ye2017efficient}, which can be implemented as $1\times 1$ convolutions, fully convolutional layers~\cite{hartmann2017learned}, or convolutional layers followed by fully-connected layers. The decision network is trained jointly with the feature extraction module to assess the similarity between two image patches. Han \etal~\cite{han2015matchnet} use a top network composed of three fully-connected layers followed by a softmax.   Zagoruyko and Komodakis~\cite{zagoruyko2015learning}  use two linear fully connected layers (each with $512$ hidden units) that are separated by a ReLU activation layer while the MC-CNN-acrt network of Zbontar \etal~\cite{zbontar2016stereo} use up to five fully-connected layers. In all cases, the features computed by the two branches of the feature encoding module are first concatenated and then fed to the top network. Hartmann \etal~\cite{hartmann2017learned}, on the other hand, aggregate the features coming from multiple patches using mean pooling before feeding them to a decision network. The main advantage of aggregation by pooling over concatenation is that the former can handle any arbitrary number of patches without changing the architecture of the network or re-training it. As such, it is suitable for computing multi-patch similarity.


Using a decision network instead of hand-crafted similarity measures  enables learning, from data, the appropriate similarity measure instead of imposing  one at the outset.  It is more accurate than using a correlation layer but is significantly slower. 

\subsubsection{Network architecture variants}

Since its introduction, the baseline architecture has been extended in several ways  in order to: \textbf{(1)} improve training using residual networks (ResNet)~\cite{shaked2017improved}, \textbf{(2)}   enlarge the receptive field of the network without losing in resolution or in computation efficiency~\cite{park2017look,ye2017efficient,Fu_2018_CVPR},  \textbf{(3)} handling multiscale features~\cite{zagoruyko2015learning,chen2015deep},  \textbf{(4)} reducing the number of forward passes~\cite{zagoruyko2015learning,luo2016efficient}, and \textbf{(5)} easing the training procedure by learning similarity without explicitly learning features~\cite{zagoruyko2015learning}.


\paragraph{ConvNet vs. ResNet}
\label{sec:convnet_vs_resnet}

While Zbontar \etal~\cite{zbontar2015computing,zbontar2016stereo} and Han \etal~\cite{han2015matchnet} use standard convolutional layers in the feature extraction block,   Shaked and Wolf~\cite{shaked2017improved} add residual blocks with multilevel weighted residual connections to facilitate the training of very deep networks.   Its particularity is that the network  learns by itself how to adjust the contribution of the added skip connections.    It was demonstrated that this architecture outperforms  the base network of Zbontar \etal~\cite{zbontar2015computing}.   


\paragraph{Enlarging the receptive field of the network}
\label{sec:local_vs_global_context}

The  scale of the learned features is defined by \textbf{(1)} the size of the input patches,  \textbf{(2)} the receptive field of the network, and  \textbf{(3)} the kernel size of the convolutional filters and pooling operations used in each layer.  While increasing the kernel sizes allows the capture of more global interactions between the image pixels, it induces a high computational cost. Also, the conventional pooling,  as used in~\cite{zbontar2015computing,zbontar2016stereo},  reduces resolution and could cause the loss of fine details, which is not suitable  for dense correspondence estimation.

To enlarge the receptive field without losing resolution or increasing the computation time, some techniques, \eg~\cite{Fu_2018_CVPR}, use dilated convolutions, \ie large convolutional filters but with holes and thus they are computationally efficient. Other techniques, \eg~\cite{park2017look,ye2017efficient}, use spatial pyramid pooling (SPP) modules placed at different locations in the network, see Fig.~\ref{fig:feature_learning_architectures}-(c-e).  For instance, Park \etal~\cite{park2017look}, who introduced FW-CNN for stereo matching, append an SPP module at the end of the decision network, see Fig.~\ref{fig:feature_learning_architectures}-(d). As a result, the receptive field can be enlarged. However,  for each pixel in the reference image,  both the fully-connected layers and the pooling operations need to be computed $\ndisparities$ times where $\ndisparities$ is the number of disparity levels. To avoid this, Ye \etal~\cite{ye2017efficient} place the SPP module at the end of each feature computation branch, see   Figs.~\ref{fig:feature_learning_architectures}-(c) and (e). In this way, it is only computed once for each patch. Also, Ye \etal~\cite{ye2017efficient}  employ multiple one-stride poolings, with different window sizes, to different layers and then concatenate their outputs  to generate the feature maps, see Fig.~\ref{fig:feature_learning_architectures}-(e).

\paragraph{Learning multiscale features}
\begin{figure}[t]
\centering{
\begin{tabular}{@{}cc@{}}
	\includegraphics[height=0.2\textheight]{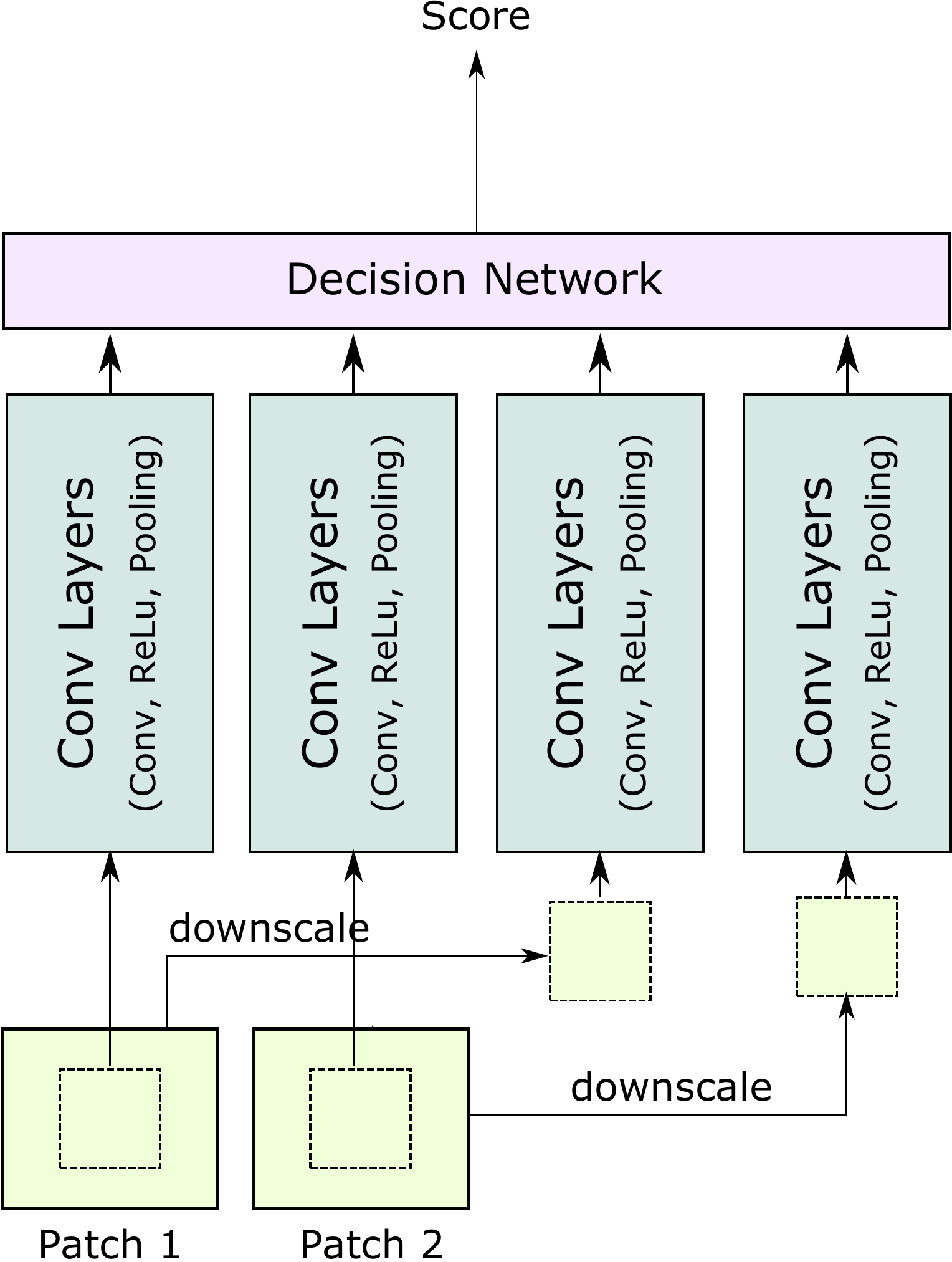}&
	\includegraphics[height=0.2\textheight]{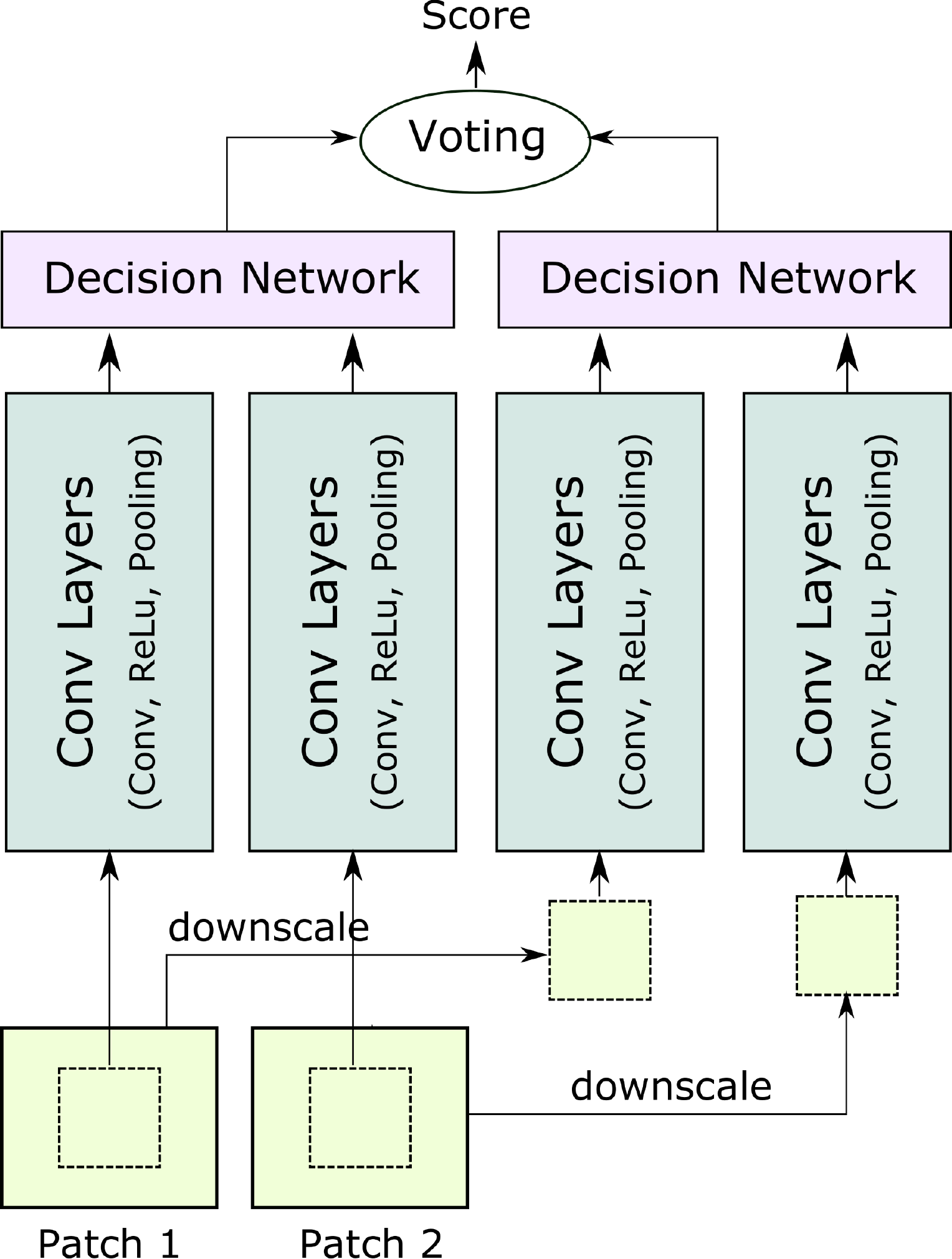} \\
	\footnotesize{(a)  Center-surround~\cite{zagoruyko2015learning}} & \footnotesize{(b) Voting-based~\cite{chen2015deep}.} \\
\end{tabular}
\caption{\label{fig:multi_scale_feature_learning_architectures} Multiscale feature learning architectures.  }
}
\end{figure}

The methods described so far can be extended to learn features at multiple scales by using multi-stream networks, one stream per patch size~\cite{zagoruyko2015learning,chen2015deep}, see Fig.~\ref{fig:multi_scale_feature_learning_architectures}. Zagoruyko and Komodakis~\cite{zagoruyko2015learning} propose a two-stream network, which is essentially a network composed of two siamese networks combined at the output by a top network, see Fig.~\ref{fig:multi_scale_feature_learning_architectures}-(a). The first siamese network, called central high-resolution stream, receives as input two $32 \times 32$ patches centered around the pixel of interest.  The second network, called surround low-resolution stream,  receives as input two $64 \times 64$ patches but down-sampled to $32 \times 32$.  The output of the two streams are then concatenated and fed to a top decision network, which returns a matching score.  Chen \etal~\cite{chen2015deep}  use a similar approach but instead of aggregating the features computed by the two streams prior to feeding them to the top  decision network,  it appends a top network on each stream to produce a matching score. The two scores are then aggregated by voting, see Fig.~\ref{fig:multi_scale_feature_learning_architectures}-(b). 

The main advantage of the multi-stream architecture is that it can compute  features at multiple scales in a single forward pass. It, however,  requires one stream per scale, which  is not practical if more than two scales are needed. 

\paragraph{Reducing the number of forward passes}

Using the approaches described so far, inferring the raw cost volume  from a pair of stereo images is performed using a moving window-like approach, which would require multiple forward passes, $\ndisparities$ forward passes per pixel where  $\ndisparities$ is the number of disparity levels.   However, since correlations are highly parallelizable,  the number of forward passes can be significantly reduced. For instance,  Luo \etal~\cite{luo2016efficient} reduce the number of forward passes to one pass per pixel by using a siamese network, whose first branch takes a patch around a pixel while the second branch takes a larger patch that expands over all possible disparities.    The output is  a single 64D representation for the left branch, and $\ndisparities \times 64$ for the right branch.  A correlation layer then computes a vector of length $\ndisparities$, where its $\depth-$th element is the cost of matching the pixel $\pixel$ on the left image with the pixel $\pixel - \depth$ on the rectified right image.

Zagoruyko and Komodakis~\cite{zagoruyko2015learning}  showed that the  outputs of the two feature extraction sub-networks need to be computed only once per pixel, and do not need to be recomputed for every disparity under consideration. This can be done in a single forward pass, for the entire image,  by propagating full-resolution images instead of small patches. Also, the output of the top network composed of fully-connected layers in the accurate architecture (\ie MC-CNN-Accr) can be computed in a single forward pass by replacing the fully-connected layers with convolutional layers of $1\times 1$ kernels. However, it still requires one forward pass for each disparity under consideration. 

\paragraph{Learning similarity without  feature learning}
Joint training  of feature extraction  and similarity computation networks unifies the feature learning and the metric learning steps. Zagoruyko and Komodakis~\cite{zagoruyko2015learning} propose another architecture that does not have a direct notion of features, see Fig.~\ref{fig:feature_learning_architectures}-(f). In this architecture, the left and right patches are packed together and fed jointly into a two-channel network composed of convolution and ReLU layers followed by a set of fully connected layers. Instead of computing features, the network directly outputs the similarity between the input pair of patches. Zagoruyko and Komodakis~\cite{zagoruyko2015learning} showed that this architecture is easy to train. However, it is expensive at runtime since the whole network needs to be  run $\ndisparities$ times per pixel.

\subsubsection{Training procedures}

The networks described in this section are composed of a feature extraction block and a feature matching block. Since the goal is to learn how to match patches, these two modules are jointly trained either in a supervised  (Section~\ref{sec:supervisied_matching}) or in a weakly supervised manner (Section~\ref{sec:weakly_supervisied_matching}).

\paragraph{Supervised training} 
\label{sec:supervisied_matching}
Existing methods for supervised training use a training set composed of positive and negative examples. Each positive  (respectively negative) example is a pair composed of a reference patch and its matching patch (respectively a non-matching one) from another image.  Training either takes one example at a time, positive or negative, and adapts the similarity~\cite{chen2015deep,simo2015discriminative,han2015matchnet,zagoruyko2015learning},  or takes at each step both a positive and a negative example, and maximizes the difference between the similarities, hence aiming at making the two patches from the positive pair \textit{more similar} than the two patches from the negative pair~\cite{balntas2016pn,kumar2016learning,zbontar2015computing}. This latter scheme is known as \textit{Triplet Contrastive learning}.

Zbontar \etal~\cite{zbontar2015computing,zbontar2016stereo} use the ground-truth disparities of the KITTI2012~\cite{geiger2012we} or Middlebury~\cite{scharstein2014high}  datasets. For each known disparity, the method extracts one negative pair and one positive pair as training examples. As such, the method is able to extract more than $25$ million training samples from KITTI2012~\cite{geiger2012we} and more than $38$ million from the Middlebury dataset~\cite{scharstein2014high}. This method has been also used by Chen \etal~\cite{chen2015deep},   Zagoruyku and  Komodakis~\cite{zagoruyko2015learning}, and Han \etal~\cite{han2015matchnet}. The amount of training data can be further augmented by using data augmentation techniques, \eg flipping patches and rotating them in various directions.

Although  the supervised learning works very well, the complexity of the neural network models requires very large labeled training sets, which are hard or costly to collect for real applications (\eg consider the stereo reconstruction of the Mars landscape). Even when such large sets are available, the ground truth is usually produced from depth sensors  and often contains noise that reduces the effectiveness of the supervised learning~\cite{sukhbaatar2014learning}. This can be mitigated by augmenting the training set with random perturbations~\cite{zbontar2015computing} or synthetic data~\cite{fischer2014descriptor,mayer2016large}. However, synthesis procedures are hand-crafted and do not account for the regularities specific to the stereo system and target scene at hand.

\vspace{6pt}
\noi\textbf{Loss functions. }  Supervised stereo matching networks are trained to minimize a matching loss, which is a function that measures the discrepancy between the ground-truth and the predicted matching scores for each training sample. It can be defined using \textbf{(1) }the $\lone$ distance~\cite{zbontar2016stereo,chen2015deep,shaked2017improved}, \textbf{(2)} the hinge loss~\cite{zbontar2016stereo,shaked2017improved}, or \textbf{(3)} the cross-entropy loss~\cite{luo2016efficient}.  


\paragraph{Weakly supervised learning}
\label{sec:weakly_supervisied_matching}

Weakly supervised techniques exploit one or more stereo constraints to reduce the amount of manual labelling. Tulyakov \etal~\cite{tulyakov2017weakly} consider Multi-Instance Learning (MIL) in conjunction with stereo constraints and coarse information about the scene to train stereo matching networks with datasets for which ground truth is not available. Unlike supervised techniques, which require pairs of matching and non-matching patches, the training set is composed of $N$ triplets. Each triplet is composed of: \textbf{(1)}  $W$ reference patches extracted on a horizontal line of the reference image, \textbf{(2)}  $W$ positive patches extracted from  the corresponding horizontal line  on the right image, and \textbf{(3)} $W$ negative patches, \ie patches that do not match the reference patches, extracted from another horizontal line on the right image.  As such, the training set can automatically be constructed from stereo pairs without manual labelling. 

The method is then trained by exploiting five constraints: the epipolar constraint, the disparity range constraint, the uniqueness constraint, the continuity (smoothness) constraint, and the ordering constraint. They then define three losses that use different subsets of these constraints, namely:
\begin{itemize}
	\item The Multi Instance Learning (MIL) loss, which uses the epipolar and the disparity range constraints.  From these two constraints, we know that every non-occluded reference patch has a matching positive patch in a known index interval, but does not have a matching negative patch. Therefore, for every reference patch, the similarity of the best reference-positive match should be greater than the similarity of the best reference-negative match.
	\item The constractive loss, which adds to the MIL method  the uniqueness constraint. It tells us that the matching positive patch is unique. Thus,  for every patch, the similarity of the best match should be greater than the similarity of the second best match.
	\item The constractive-DP   uses all the constraints  but finds the best match using dynamic programming.
\end{itemize}

\noi The method has been applied to train a deep siamese neural-network that takes two patches as an input and predicts a similarity measure.
Benchmarking on standard datasets shows that the performance is as good or better than the published results on MC-CNN-fst~\cite{zbontar2015computing}, which uses the same network architecture but trained using fully labeled data.

\subsection{Regularization and disparity estimation}

Once the raw  cost volume is estimated, one can estimate the disparity  by  dropping the regularization term of Eqn.~\eqref{eq:stereomatching_energy}, or equivalently block C of Fig.~\ref{fig:stereo_matching_pipeline},  and taking the argmin, the softargmin, or the subpixel MAP approximation (block D of Fig.~\ref{fig:stereo_matching_pipeline}).   However, the raw cost volume computed from image features could be noise-contaminated, \eg due to the existence of non-Lambertian surfaces, object occlusions, or repetitive patterns. Thus, the estimated depth maps can be noisy.   As such, some methods overcome this problem by using traditional MRF-based stereo framework for cost volume regularization~\cite{chen2015deep,zbontar2015computing,luo2016efficient}. In these methods, the initial cost volume $\costvolume$  is fed to a  global~\cite{scharstein2002taxonomy} or a semi-global~\cite{hirschmuller2008stereo} matcher to compute the disparity map.  Semi-global matching  provides a good tradeoff between accuracy and computation requirements. In this method, the smoothness term of Eqn.~\eqref{eq:stereomatching_energy} is defined as:
\begin{equation}
\small{
	\smoothnessenergy(\depth_\pixel, \depth_y) = \alpha_1 \delta(|\depth_{\pixel y} = 1) + \alpha_2 \delta(|\depth_{\pixel y} > 1),
	}
	\label{eq:semiglobal_matching}
\end{equation}

\noi where $\depth_{\pixel y} = \depth_\pixel - \depth_y$,  $\alpha_1$ and $\alpha_2$ are positive weights chosen such that $\alpha_2 > \alpha_1$, and $\delta$ is the Kronecker delta function, which gives $1$ when the condition in the bracket is satisfied, otherwise $0$.  To solve this optimisation problem, the SGM energy is broken down into multiple energies $\energy_s$, each one defined along a path $s$. The energies are  minimised independently and then aggregated. The disparity at $\pixel$ is computed using the winner-takes-all strategy of the aggregated costs of all directions:
\begin{equation}
	\disparity_\pixel = \argmin_{\disparity} \sum_s\energy_s(\pixel, \disparity).
\end{equation}

\noi 
This method requires setting the two parameters $\alpha_1$ and $\alpha_2$ of Eqn.~\eqref{eq:semiglobal_matching}. Instead of manually setting them,  Seki \etal~\cite{seki2017sgm} proposed  SGM-Net, a neural network trained to provide these parameters at each image pixel.  They obtained better penalties than  hand-tuned methods as in~\cite{zbontar2015computing}. 

The SGM method, which uses an aggregated scheme to combine costs from multiple 1D scanline optimizations, suffers from two major issues: \textbf{(1)} streaking artifacts caused by the scanline optimization approach, at the core of this algorithm, may lead to inaccurate results,  and \textbf{(2)}  the high memory footprint that may become prohibitive with high resolution images or devices with constrained resources. As such Schonberger \etal~\cite{schonberger2018learning} reformulate the fusion step as the task of selecting the best amongst all the scanline optimization proposals at each pixel in the image. They  solve this task using   a per-pixel random forest classifier.

Poggi \etal~\cite{poggi2016learning} learn a weighted aggregation where the weight of each 1D scanline optimisation  is defined using a confidence map computed using either traditional techniques~\cite{hu2012quantitative}  or deep neural networks, see Section~\ref{sec:confidencemaps}.

\section{End-to-end depth from stereo}
\label{sec:end_to_end_stereo}
\begin{figure*}[t]
\centering
	\includegraphics[trim=7cm 0cm 0cm 0cm, clip=true, width=\textwidth]{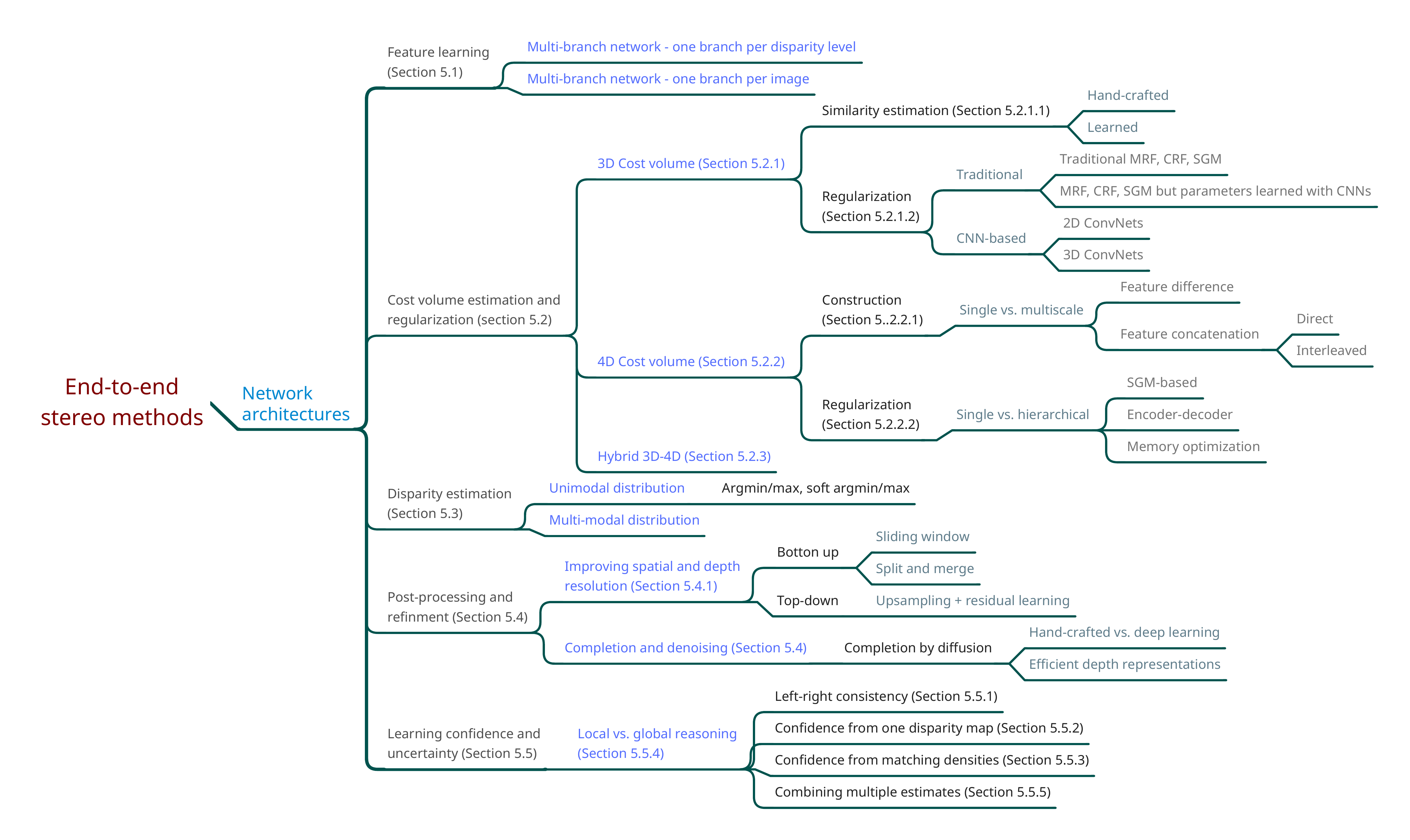}
	\caption{\label{tab:taxonomy_end_to_end} Taxonomy of the network architectures  for stereo-based disparity estimation using end-to-end deep learning. }
\end{figure*}

%
%
%

Recent works solve the stereo matching problem using a pipeline that is trained end-to-end. Two main classes of methods have been proposed.  Early methods, \eg FlowNetSimple~\cite{dosovitskiy2015flownet} and DispNetS~\cite{mayer2016large}, use a single encoder-decoder, which stacks together the left and right images into a 6D volume, and regresses the disparity map. These methods, which do not require an explicit feature matching module, are fast at runtime. They, however, require a large amount of training data, which is hard to obtain. Methods in the second class mimic the traditional stereo matching pipeline by breaking the problem into stages, each stage is composed of differentiable blocks and thus allowing end-to-end training. Below, we review in details these techniques. Fig.~\ref{tab:taxonomy_end_to_end} provides a taxonomy of the state-of-the-art, while Table~\ref{tab:taxonomy_end_to_end_stereomatching} compares $28$ key methods based on this taxonomy.


\subsection{Feature learning}
\label{sec:end_to_end_featurelearning}

Feature learning networks follow the same architectures as the ones described in Figs.~\ref{fig:feature_learning_architectures} and~\ref{fig:multi_scale_feature_learning_architectures}. However, instead of processing individual patches, the entire images are processed in a single forward pass producing feature maps of the same or lower resolution as the input images.  Two strategies have been used to enable  matching features across the images:

\vspace{6pt}
\noi \textit{(1) Multi-branch networks composed of $\nimages$ branches where $\nimages$ is the number of input images}. Each branch produces a feature map that characterizes its input image~\cite{mayer2016large,kar2017learning,kendall2017end,pang2017cascade,liang2018learning,chang2018pyramid,nie2019multi}. These techniques assume that the input images have been rectified so that the search for correspodnences is restricted to be along the horizontal scanlines.  
	
\vspace{6pt}
\noi  \textit{(2) Multi-branch networks composed of $\ndisparities$ branches where $\ndisparities$ is the number of disparity levels.}  The $d$-th  branch, $ 1 \le d \le \ndisparities$, processes a stack of two images, as in Fig.~\ref{fig:feature_learning_architectures}-(f); the first image is the reference image. The second one is the right image but re-projected to the $d$-th depth plane~\cite{flynn2016deepstereo}. Each branch produces a similarity feature map that characterizes the similarity between the reference image and the right image re-projected onto a given depth plane. 	While these techniques do not rectify the images, they assume that the intrinsic and extrinsic camera parameters are known. Also, the number of disparity levels cannot be varied without updating the network architecture and  retraining it.

In both methods, the feature extraction module uses  either  fully convolutional  (ConvNet) networks such as VGG, or  residual networks such as ResNets~\cite{he2016deep}. The latter facilitates the training of very deep networks~\cite{yang2018segstereo}. They can  also capture and incorporate more global context in the unary features by using either dilated convolutions (Section~\ref{sec:local_vs_global_context}) or multi-scale approaches. For instance, the PSM-Net of Chang and Chen~\cite{chang2018pyramid}  append a Spatial Pyramid Pooling (SPP) module in order to extract and aggregate features at multiple scales.  Nie \etal~\cite{nie2019multi} extended PSM-Net using a multi-level context aggregation pattern termed \emph{Multi-Level Context Ultra-Aggregation  (MLCUA)}. It encapsulates all convolutional features into a more discriminative representation by intra and inter-level features combination. It combines the features at the shallowest, smallest scale with features at deeper, larger scales using just shallow skip connections. This results in an improved performance, compared to PSM-Net~\cite{chang2018pyramid}, without significantly increasing the number of parameters in the network.

 \begin{sidewaystable*}
	\caption{\label{tab:taxonomy_end_to_end_stereomatching} Taxonomy and comparison of $28$ end-to-end deep learning-based disparity estimation techniques. "FCN": Fully-Connected Network, "SPN": Spatial Propagation Network. "LRCR": Left-Right Comparative Recurrent model, "MCUA": Multi-Level Context Ultra-Aggregation for Stereo Matching.  "DLA": Deep layer aggregation~\cite{yu2018deeplayer}, "VPP": Volumetric Pyramid Pooling.  The performance is measured on KITTI2015 test dataset.}
	
	\resizebox{\linewidth}{!}{%
	\begin{tabular}{@{}l@{ }l @{ }c@{ }c @{ }c   @{ }c   @{ }c@{ }c@{ }c @{ }c @{ }c @{ }c @{ }c@{ }c@{ }c@{ }c@{ }c@{ } c@{ }c@{ }c@{}}
	\toprule
	\multirow{2}{*}{\textbf{Method}}  &  \multirow{2}{*}{\textbf{Year}} &&  \multicolumn{2}{c}{\textbf{Feature computation}} & & \multicolumn{3}{c}{\textbf{Cost volume}} & & \multirow{2}{*}{\textbf{Disparity} }& & \multicolumn{2}{c}{\textbf{Refinement/post processing}}  & & \multirow{2}{*}{\textbf{Supervision}}   &  \multicolumn{2}{c}{\textbf{Performance}}  \\
	\cline{4-5} \cline{7-9} \cline{13-14} \cline{17-18}
	
		& & &\textbf{Architecture} & \textbf{Dimension} & & \textbf{Type} & \textbf{Construction} & \textbf{Regularization} & & & & \textbf{Spatial/depth resolution} & \textbf{Completion/denoising} & & & \textbf{D1-all/est} & \textbf{D1-all/fg}  \\
	\toprule

	FlowNetCorr~\cite{dosovitskiy2015flownet}& 2015  & & ConvNet& Single scale&  & 3D&Correlation & 2D ConvNet  & & & & Up-convolutions & Ad-hoc, variational & & Supervised & $-$ & $-$\\
	\midrule

	DispNetC~\cite{mayer2016large} & 2016  &  & ConvNet & Single scale & & 3D & Correlation& 2D ConvNet& & $-$& &$-$ & $-$  & & Supervised & $4.34$ & $4.32$ \\ 
	\midrule
	
	Zhong \etal~\cite{zhong2017self}& 2017  & &  ConvNet & Single scale& & 4D & Interleaved & 3D Conv, & & Soft argmin&  &  \multicolumn{2}{c}{Self-improvement at runtime} & & Self-supervised & $3.57$ & $7.12$ \\
							  &          &  & with skip conn.  & & &  &  feature concat.& encoder-decoder& & & & & \\ 
	\midrule
	Kendall \etal~\cite{kendall2017end}& 2017  & &  ConvNet & Single scale & & 4D & Feature concat.& 3D Conv, encoder-&  & Soft argmax & &  $-$& $-$ & & Supervised & $2.87$ & $6.16$\\ 
 								&   & & with skip conn.  & & &  & & decoder, hierarchical& & & & & \\
	\midrule
	Pang \etal~\cite{pang2017cascade}& 2017  & & ConvNet & Single scale  & &3D & Correlation & 2D ConvNet& & & & Upsampling$+$ residual learning & & & Supervised & $2.67$ & $3.59$\\
	
	\midrule
	Knobelreiter \etal~\cite{knobelreiter2017end}& 2017  & & ConvNet&  Single scale & & 3D& Correlation &  Hybrid CNN-CRF & & &   &   \multicolumn{2}{c}{No post-processing} & & Supervised\\
	
	\midrule
	Chang  \etal~\cite{chang2018pyramid}& 2018  &  & SPP& Multiscale & & 4D & Feature concat.& 3D Conv, Stacked encoder-decoders& & Soft argmin& &  Progressive refinement & & & Supervised & $2.32$ & $4.62$  \\
	
	\midrule  
	Khamis \etal~\cite{khamis2018stereonet}& 2018  & & ResNet & Single scale & & 3D & $\ltwo$ & 3D ConvNet& & Soft argmin & & \multicolumn{2}{c}{Hierarchical, Upsamling$+$residual learning} & & Supervised & $4.83$ & $7.45$\\ 

	\midrule
	Liang \etal~\cite{liang2018learning}& 2018  & & ConvNet& Multiscale & & 3D& Correlation& 2D ConvNet & & Encoder-decoder & &  \multicolumn{2}{c}{Iterative upsampling$+$residual learning} & & Supervised & $2.67$ & $3.59$\\

	\midrule 
	\multirow{2}{*}{Yang \etal~\cite{yang2018segstereo}}& \multirow{2}{*}{2018}  & & Shallow  & \multirow{2}{*}{Single scale}& & \multirow{2}{*}{3D}& \multicolumn{2}{c}{\multirow{2}{*}{Correlation, Left features, segmentation mask}} &  & Regression with & & $-$& $-$ & & Self-supervised & $2.25$& $4.07$ \\
								&    &  & ResNet   & & &  &  & & & Encoder-decoder & & & & \\
	
	\midrule 
	Zhang \etal~\cite{zhang2018activestereonet}& 2018  & & ConvNet& Single scale& & 3D &Hand-crafted & NA & & Soft argmin & &  \multicolumn{2}{c}{Upsampling$+$residual learning} & & Self-supervised & $-$ & $-$\\

	\midrule
	Jie \etal~\cite{jie2018left}& 2018  & & Constant highway net& Single scale & & 3D & FCN& $-$ &  &RNN-based LRCR & & $-$ &  $-$& & Supervised & $3.03$ & $5.42$\\

	\midrule
	Ilg \etal~\cite{Ilg_2018_ECCV}& 2018  & & ConvNet & Single scale& & 3D &Correlation & \multicolumn{3}{c}{Encoder-decoder, joint disparity and occlusion}& &  \multicolumn{2}{c}{Cascade of encoder-decoders, residual learning}&  & Supervised & $2.19$ & $-$\\

	\midrule
	Song \etal~\cite{song2018stereo}& 2018  & & SHallow ConvNet& Single scale & & 3D & Correlation & Edge-guided, Context Pyramid, & & Residual pyramid& & $-$ & $-$& & Supervised  & $2.59$ & $4.18$\\
							 & 		& 	& 			& 			&		&	& Encoder & & & & \\
	
	\midrule
	Yu \etal~\cite{yu2018deep}& 2018  & & ResNet& Single scale & & 3D& Feature concatenation & 3D Conv  $+$ SGM   & & Soft argmin & & $-$& $-$ & & Supervised  & $2.79$ & $5.46$\\
	 					 &     & &  &  & &  &   Encoder-decoder  &  & & & & & \\
						 
	\midrule
	Tulyakov \etal~\cite{tulyakov2018practical}& 2018  & & $-$ & Single scale& & 4D&   Compressed &  3D Conv & &Multimodal -  & & $-$& $-$ & & Supervised & $2.58$ & $4.05$ \\
									 &    & & & & &  & matching features&  & & Sub-pixel MAP& & & \\	

	\midrule
	EMCUA \etal~\cite{nie2019multi}& 2019  & & SPP& Multiscale & & 4D& Feature concat. & 3D Conv, MCUA scheme& & Arg softmin & & $-$ & $-$ & & Supervised & $2.09$ & $4.27$\\ 
	
	\midrule 
	Yang \etal~\cite{Yang_2019_CVPR}& 2019  & & SPP& Multiscale & & Pyramidal 4D & Feature difference & Conv3D blocks $+$ Volume Pyramid Pooling& & Arg softmax & & \multicolumn{2}{c}{Spatial \& depth res. by cost volume upsampling} & & Supervised & $2.14$ & $3.85$\\  

	\midrule
	Wu \etal~\cite{Wu_2019_ICCV}& 2019  & & ResNet50, SPP& Multiscale & & Pyramidal 4D&  Feature concat.& Encoder-decoder & & 3D Conv, soft argmin & & $-$ & $-$ &  & Supervised &$2.11$ & $3.89$ \\
								    &  &   & & & & & & $+$ Feature fusion   & & & & & & & Disparity \& boundary loss \\

	\midrule
	Yin \etal~\cite{yin2019hierarchical}& 2019  & & DLA net& Multiscale & & 3D &Correlation & Density decoder &  &Outputs discrete  & & $-$& $-$ & & Supervised & $\textbf{2.02}$ & $3.63$\\ 
									  & &  &   & &   &  &  &  & & matching distribution & & & & &   \\
	\midrule
	Chabra \etal~\cite{chabra2019stereodrnet}& 2019  & & ConvNet $+$ & Multiscale & & 3D & $\lone$ &  Dilated 3D ConvNet& & Soft argmax & & \multicolumn{2}{c}{Upsampling$+$residual learning}  & & Supervised & $2.26$ & $4.95$\\
									 & & & Vortex pooling~\cite{xie2018vortex} &    \\

	\midrule
	Duggal \etal~\cite{Duggal_2019_ICCV}& 2019  & & ResNet, SPP & Multiscale & & 3D, sparse& Correlation, Adaptive& Encoder-decoder& & Soft argmax& &Encoder & $-$ & & Supervised & $2.35$ & $\textbf{3.43}$\\  
								    &		&  & & & &		 &  pruning with PatchMatch \\

	\midrule
	Tonioni \etal~\cite{tonioni2019real}& 2019  & & ConvNet& Multiscale& &3D & Correlation & & & Encoder & & \multicolumn{2}{c}{Recrusively upsampling $+$ residual learning}  & & Online self-adaptive & $4.66$ & $-$ \\	
	
	\midrule
	Yang \etal~\cite{Yang_2019_CVPR}& 2019  & & ConvNet, SPP& Multiscale & & Pyramid, 4D &  Concatenation & Decoder,  Residual blocs, & & Conv3D block& & $-$& $-$& & Supervised & $2.14$ & $3.85$\\ 
									    & & 			& 		   &  & 		      &  & &  VPP & & & &\\
	
	\midrule
	Zhang \etal~\cite{zhang2019ga}& 2019  & & Stacked hourglass & Single scale& & 4D& Concatenation& Semi-global aggregation layers,  & & Soft argmax& & $-$ & $-$  & & Supervised & $2.03$ & $3.91$\\
							&    & & & & &  & & Local-guided aggregation layers & & & & & \\

	\midrule 
	Guo \etal~\cite{guo2019group}& 2019  & & SPP& Multiscale  & & Hybrid 3D-4D& Group-wise correlation & Stacked hourglass nets& & Soft argmin  & & $-$& $-$  & &  Supervised & $2.11$ & $3.93$ \\
	
	\midrule
	Chen \etal~\cite{Chen_2019_ICCV}& 2019  & & & & & & & & &Single-modal weighted avg  & & & & &  Supervised & $2.14$ & $4.33$ \\

	\midrule
	Wang \etal~\cite{wang2019anytime}& 2019  & & ConvNet& Multiresolution maps & & 3D &  $\lone$&  Progressive refinement& & Soft argmin & &  Usampling, &  Spatial propagation & & Supervised & $-$ & $-$\\
							 & & &  &     & &   &  & (3D Conv)& & & &  residual learning& network \\
	\bottomrule
	
	\end{tabular}
	}
\end{sidewaystable*}

\subsection{Cost volume construction}
\label{sec:end_to_end_cost_volume}

Once the features have been computed, the next step is to compute the matching scores, which will be fed, in the form of a  cost volume,  to a top network for regularization and disparity estimation. The cost volume can be  three dimensional (3D) where the third dimension is  the disparity level (Section~\ref{sec:3D_cost_volumes}), four dimensional (4D) where the third dimension is  the feature dimension and the fourth one is the  disparity level (Section~\ref{sec:4D-cost_volumes}),  or hybrid to benefit from the properties of the 3D and 4D cost volumes (Section~\ref{sec:hybrid_cost_volume}). In general, the cost volume is constructed at a lower resolution,  \eg at $\nicefrac{1}{8}$-th, than the input~\cite{khamis2018stereonet,zhang2018activestereonet}. It is then either subsequently upscaled and refined, or used as is to estimate a low resolution disparity map, which is then upscaled and refined using a refinement module. 

\subsubsection{3D cost volumes}
\label{sec:3D_cost_volumes}

\paragraph{Construction}

A 3D cost volume can be simply built by taking the $\lone$, $\ltwo$, or  correlation distance between the features of the left image and those of the right image that are within a pre-defined disparity range, see~\cite{mayer2016large,zhang2018activestereonet,jie2018left,khamis2018stereonet,yin2019hierarchical,chabra2019stereodrnet,Duggal_2019_ICCV},  and the FlowNetCorr of~\cite{dosovitskiy2015flownet}.  The advantage of correlation-based dissimilarities is that they can be implemented using a convolutional layer that does not require training (its filters are the features computed by the second branch of the network). Flow estimation networks such as FlowNetCorr~\cite{dosovitskiy2015flownet} use 2D correlations. Disparity estimation networks,  such as~\cite{mayer2016large,yang2018segstereo}, iResNet~\cite{liang2018learning},  DispNet3~\cite{Ilg_2018_ECCV}, EdgeStereo~\cite{song2018stereo}, HD$^3$~\cite{yin2019hierarchical}, and~\cite{tonioni2019real,Duggal_2019_ICCV}, use 1D correlations. 	

\paragraph{Regularization of 3D cost volumes}

\begin{figure}[t]
\centering
	\includegraphics[width=.45\textwidth]{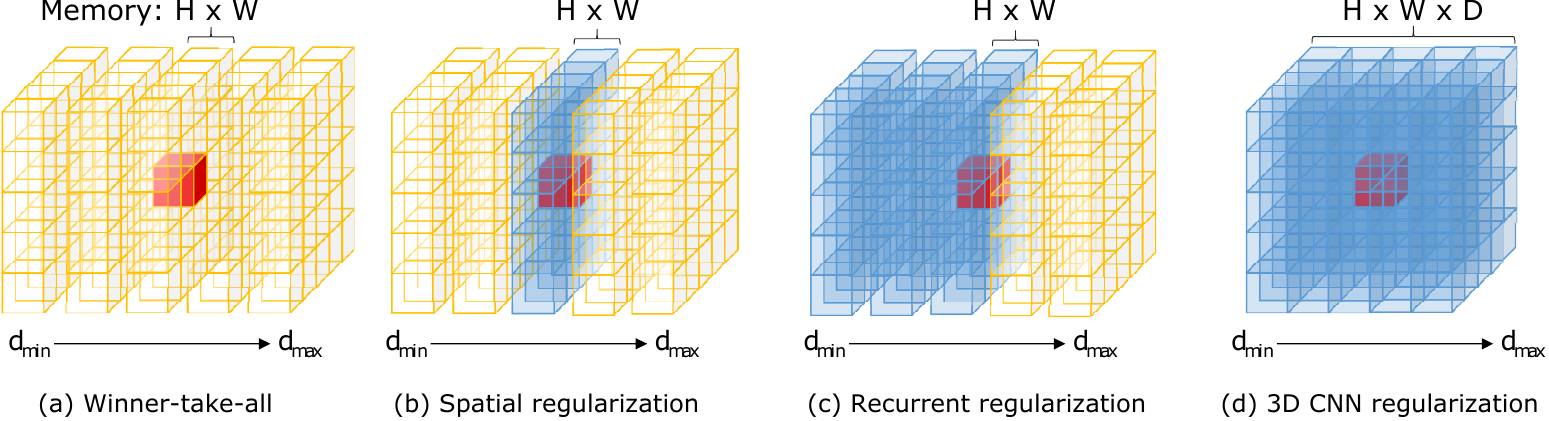}
	\caption{\label{fig:cost_volume_regularization}Cost volume regularization schemes~\cite{Yao_2019_CVPR}: (a) does not consider context, (b) captures context along the spatial dimensions using 2D convolutions, (c) captures context along the spatial and disparity dimensions by recurrent regularization using 2D convolutions, and (d) captures context in all dimensions by using 3D convolutions.   }
\end{figure}

Once a cost volume is computed, an initial disparity map can be estimated using the argmin, the softargmin, or the subpixel MAP approximation over the depth dimension of the cost volume, see for example~\cite{zhang2018activestereonet} and Fig.~\ref{fig:cost_volume_regularization}-(a). This is equivalent to dropping the regularization term of Eqn.~\eqref{eq:stereomatching_energy}.
 In general, however, the raw cost volume is noise-contaminated (\eg due to the existence of non-Lambertian surfaces, object occlusions, and repetitive patterns).   The goal of the regularization module is to leverage context along the spatial and/or disparity dimensions to refine the cost volume before estimating the initial disparity map. 

\vspace{6pt}
\noi\textit{(1) Regularization using traditional methods. } Early papers  use traditional techniques, \eg Markov Random Fields (MRF), Conditional Random Fields (CRF), and Semi-Global Matching (SGM), to regularize the cost volume by  explicitly incorporating spatial constraints, \eg smoothness, of the depth maps. Recent papers showed that deep learning networks can be used to fine-tune the parameters of these methods. For example,  Kn{\"o}belreiter \etal~\cite{knobelreiter2017end} proposed  a hybrid CNN-CRF. The CNN  computes the matching term of Eqn.~\eqref{eq:stereomatching_energy}, which becomes the unary term of a CRF module. The pairwise term of the CRF is parameterized by edge weights computed using another CNN. The  end-to-end trained  CNN-CRF  pipeline could achieve a competitive performance using much fewer parameters (thus a better utilization of the training data) than the earlier methods. 

Zheng \etal~\cite{zheng2015conditional}  provide a way to model CRFs as recurrent neural networks (RNN) for segmentation tasks so that the entire pipeline can be trained end-to-end. Unlike  segmentation, in depth estimation, the number of depth samples, whose counterparts are the semantic labels in segmentation tasks, is expected to vary for different scenarios. As such, Xue \etal~\cite{Xue_2019_ICCV} re-designed the RNN-formed CRF module so that the model parameters are independent of the number of depth samples. Paschalidou \etal~\cite{Paschalidou_2018_CVPR}  formulate the inference in a MRF as a differentiable function, hence allowing end-to-end training using back propagation.  Note that Zheng \etal~\cite{zheng2015conditional}  and  Paschalidou \etal~\cite{Paschalidou_2018_CVPR} focus on multi-view stereo (Section~\ref{sec:mvs_architectures}). Their approaches, however, are generic and can be  used to regularize 3D cost volumes obtained using pairwise stereo networks.

\vspace{6pt}
\noi\textit{(2) Regularization using 2D convolutions (2DConvNet), Figs.~\ref{fig:cost_volume_regularization}-(b) and (c). } Another  approach is to process the 3D cost volume using a series of 2D convolutional layers producing another 3D cost volume~\cite{mayer2016large,dosovitskiy2015flownet,pang2017cascade,liang2018learning}. 2D convolutions are computationally efficient. However, they only capture and aggregate context along the spatial dimensions, see Fig.~\ref{fig:cost_volume_regularization}-(b),  and ignore context along the disparity dimension. Yao \etal~\cite{Yao_2019_CVPR}  sequentially regularize the 2D cost maps along the depth direction via a Gated Recurrent Unit (GRU), see Fig.~\ref{fig:cost_volume_regularization}-(c). This reduces drastically the memory consumption, \eg from  $15.4$GB in~\cite{yao2018mvsnet} to around $5$GB,  making high-resolution reconstruction feasible, while  capturing context along both the spatial  and the disparity dimensions. 

\vspace{6pt}
\noi\textit{(3) Regularization using 3D convolutions (3DConvNet), Fig.~\ref{fig:cost_volume_regularization}-(d). } Khamis \etal~\cite{khamis2018stereonet}  use the $\ltwo$ distance to compute an initial 3D cost volume and 3D convolutions to regularize it across  both the spatial  and disparity dimensions, see Fig.~\ref{fig:cost_volume_regularization}-(d).  Due to its memory requirements, the approach first estimates a low-resolution disparity map, which is then progressively improved using residual learning.   Zhang \etal~\cite{zhang2018activestereonet} follow the same approach but the refinement block starts with separate convolution layers running on the upsampled disparity and input image respectively, and merge the features later to produce the residual.  Chabra \etal~\cite{chabra2019stereodrnet} observe that the cost volume regularization step is the one that uses most of the computational resources.  They then propose a regularization module that uses 3D dilated  convolutions  in the width, height, and disparity dimesions, to reduce the computation time while capturing a wider context.

\subsubsection{4D cost volumes}
\label{sec:4D-cost_volumes}

\paragraph{Construction}
4D cost volumes to preserve the dimension of the features~\cite{zhong2017self,kendall2017end,chang2018pyramid,nie2019multi,Yang_2019_CVPR,Wu_2019_ICCV}. The rational behind 4D cost volumes is to let the top network learn the appropriate similarity measure for comparing the features instead of using hand-crafted ones  as in Section~\ref{sec:3D_cost_volumes}. 

4D cost volumes can be constructed  by feature  differences across a pre-defined disparity range~\cite{Yang_2019_CVPR}, which results in cost volume of size  $\height \times \width \times 2\ndisparities \times \featuredim$, or by concatenating the features  computed by the different branches of the network~\cite{kendall2017end,zhong2017self,chang2018pyramid,nie2019multi,Wu_2019_ICCV}. Using this method, Kendall \etal~\cite{kendall2017end}  build  a 4D volume of size $\height \times \width \times (\ndisparities +1) \times \featuredim$ ($\featuredim$ here is the dimension of the features).    Zhong \etal~\cite{zhong2017self} follow the same approach but  concatenate the features in an interleaved manner. That is, if $\featuremap_L$ is the feature map of the left image and $\featuremap_R$ the feature map of the right image, then  the final feature volume is assembled in such a way that its $2i-$th slice holds the left feature map while the $(2i+1)-$th slice holds the right feature map but at disparity $\disparity = i$.  This results in a 4D cost volume that is twice larger than the cost volume of Kendall \etal~\cite{kendall2017end}.  To capture multi-scale context in the cost volume, Chang and Chen~\cite{chang2018pyramid} generate for each input image  a pyramid of features, upsamples them to the same dimension, and then builds a single 4D cost volume by concatenation.  Wu \etal~\cite{Wu_2019_ICCV} build from the multiscale features (four scales) multiscale 4D cost volumes.

4D cost volumes carry richer information compared to 3D cost volumes. Note, however, that volumes obtained by concatenation  contain no information about the feature similarities, so more parameters are required in the subsequent modules to learn the similarity  function.

%

\paragraph{Regularization of 4D cost volumes}

4D cost volumes are regularized with 3D convolutions, which exploit the correlation in height, width and disparity dimensions, to produce a 3D cost volume.  Kendall \etal~\cite{kendall2017end} use  a U-net encoder-decoder with 3D convolutions and skip connections. Zhong \etal~\cite{zhong2017self}  use a similar approach but add  residual connections from the contracting to the expanding parts of the regularization network. To   take into account a large context without a significant additional computational burden, Kendall \etal~\cite{kendall2017end} regularize the cost volume hierarchically, with four levels of subsampling, allowing to explicitly leverage context with a wide field of view.   Muliscale 4D cost volumes~\cite{Wu_2019_ICCV}  are aggregated into a single 3D cost volume using a 3D multi-cost aggregation module, which operates in a pairwise manner  starting with the smallest volume. Each volume is  processed with an encoder-decoder, upsampled to the next resolution in the pyramid, and  then fused using a 3D feature fusion module.

Also, semi-global matching (SGM) techniques  have been used to regularize the 4D cost volume where their parameters are estimated using convolutional networks. In particular, Yu \etal~\cite{yu2018deep} process the initial 4D cost volume with an encoder-decoder composed of 3D convolutions and upconvolutions, and produces another 3D cost volume. The subsequent aggregation step is performed using an end-to-end two-stream network:  the first stream generates three cost aggregation proposals $\costvolume_i$, one along each of the tree dimensions, \ie the height, width, and disparity.  The second stream is a guidance stream used to select the best proposals.  It uses 2D convolutions to produce three guidance (confidence) maps $W_i$.  The final 3D cost volume is  produced as a weighted sum of the three proposals, \ie $\max_i(\costvolume_i * W_i)$. 


3D convolutions are expensive in terms of memory requirements and computation time. As such, subsequent works that followed the seminal work of Kendall \etal~\cite{kendall2017end} focused on \textbf{(1)} reducing the number of 3D convolutional layers~\cite{zhang2019ga}, \textbf{(2)} progressively refining the cost volume and the disparity map~\cite{chang2018pyramid,wang2019anytime}, and \textbf{(3)} compressing the 4D cost volume~\cite{tulyakov2018practical}. Below, we discuss these approaches.

\vspace{6pt}
\noi\textit{(1) Reducing the number of 3D convolutional layers. } 
Zhang \etal~\cite{zhang2019ga} introduced GANet, which replaces a large number of the 3D convolutional layers in the regularization block with \textbf{(1)} two 3D convolutional layers, \textbf{(2)} a semi-global aggregation layer  (SGA), and \textbf{(3)} a  local guided aggregation layer (LGA). SGA is a differentiable approximation of the semi-global matching (SGM). Unlike SGM, in SGA the user-defined parameters are learnable. Moreover, they are added as penalty coefficients/weights of the matching cost terms.  Thus, they are  adaptive and more flexible at different locations for different situations. The LGA layer, on the other hand, is appended at the end and  aims to refine the thin structures and object edges. The SGA and LGA layers, which are used to replace the  costly 3D convolutions,  capture local and  whole-image cost dependencies. They significantly improve the accuracy of the disparity estimation in challenging regions such as occlusions, large textureless/reflective regions, and thin structures. 

\vspace{6pt}
\noi\textit{(2) Progressive approaches. } Some techniques avoid  directly regularizing  high resolution 4D cost volumes using the expensive 3D convolutions. Instead, they operate in a progressive manner. For instance, Chang and Chen~\cite{chang2018pyramid} introduced PSM-Net, which first estimates a low resolution 4D cost volume, and  then regularizes it using stacked hourglass 3D encoder-decoder blocks. Each block returns a 3D cost volume, which is then upsampled and used to regress a high resolution disparity map using additional 3D convolutional layers followed by  a softmax operator. As such, the stacked hourglass blocks can be seen as refinement modules. 

Wang \etal~\cite{wang2019anytime} use a three-stage disparity estimation network, called AnyNet, which builds cost volumes in a coarse-to-fine manner.  The first stage takes as input low resolution feature maps, builds a low resolution 4D cost volume and then uses 3D convolutions to estimate a low resolution disparity map by searching on a small disparity range.  The prediction in the previous level is then upsampled and used to warp the input feature at the higher scale, with the same disparity estimation network used to estimate disparity residuals. The advantage  is two-fold; \textbf{first}, at higher resolutions, the network only learns to predict residuals, which reduces the computation cost. \textbf{Second}, the approach is progressive and one can select to return the intermediate disparities, trading  accuracy for speed. 


\vspace{6pt}
\noi\textit{(3) 4D cost volume compression. } Tulyakov \etal~\cite{tulyakov2018practical} reduce the memory usage, without  having to sacrify accuracy,  by compressing the  features into compact matching signatures. As such, the memory footprint is significantly reduced. More importantly, it allows the network to handle an arbitrary number of multiview images and to vary the number of inputs at runtime without having to re-train the network. 


\subsubsection{Hybrid 3D-4D cost volumes} 
\label{sec:hybrid_cost_volume}
The correlation layer provides an efficient way to measure feature similarities, but it loses much information because it produces only a single-channel  map for each disparity level.  On the other hand,  4D cost volumes obtained by feature  concatenation carry more information but are resource-demanding. They also  require more parameters in the subsequent aggregation network to learn the similarity  function.  To benefit from both, Guo \etal~\cite{guo2019group} propose a hybrid approach, which constructs two  cost volumes; one by  feature concatenation but compressed into $12$ channels using two convolutions. The second one is built by  dividing the high-dimension feature maps into $N_g$  groups along the feature channel,  computing correlations within each group at all disparity levels, and finally concatenating the correlation maps forming another 4D volume.  The two volumes are then combined together and passed to a 3D regularization module composed of four 3D convolution layers followed by  three stacked 3D hourglass networks.  This approach results  in a significant reduction of parameters compared to 4D cost volumes built by only feature concatenation, without losing too much information like full correlations.


\subsection{Disparity computation}
\label{sec:final_depth_estimation}

The simplest way to estimate  the disparity map  from the regularized cost volume  $\costvolume$ is by using the pixel-wise  argmin, \ie  $\depth_{\pixel}  = \arg\min_{\depth} \costvolume(\pixel, \depth)$ (or equivalently $\arg\max$ if the volume $\costvolume$ encodes the likelihood). However, the agrmin/argmax operator is unable to produce sub-pixel accuracy and cannot be trained with back-propagation due to its non-differentiability.  Another approach is  the differentiable soft argmin/max over disparity~\cite{flynn2016deepstereo,kendall2017end,zhang2018activestereonet,khamis2018stereonet}:
\begin{equation}
	\disparity^* =\frac{1}{ \sum_{j=0}^{\ndisparities}e^{-\costvolume(x, j)}}  \sum_{d=0}^{\ndisparities} \disparity\times  e^{-\costvolume(x, \disparity)}.    
\end{equation}

\noi The soft argmin operator approximates the sub-pixel MAP solution when the distribution is unimodal and symmetric~\cite{tulyakov2018practical}. When this assumption is not fulfilled, the softargmin blends the modes and may produce a solution that is far from all the modes and may result in over smoothing. Chen \etal~\cite{Chen_2019_ICCV} observe that  this is particularly the case  at  boundary pixels  where the estimated disparities  follow multimodal distributions.  To address these issues, Chen \etal~\cite{Chen_2019_ICCV} only apply a weighted average operation on a window centered around the modal with the maximum probability, instead of using a full-band weighted average  on the entire  disparity range.

Tulyakov \etal~\cite{tulyakov2018practical} introduced the sub-pixel MAP approximation, which computes a weighted mean around the disparity with the maximum posterior probability as:
\begin{equation}
	\disparity^* = \sum_{\disparity:  |\hat\disparity - \disparity | \le \delta}\disparity \cdot \sigma( \costvolume(x, \disparity) ),
\end{equation}

\noi where $\delta$ is a meta parameter set to $4$ in~\cite{tulyakov2018practical},  $\sigma( \costvolume(x, \disparity) )$ is the probability of the pixel $x$ having a disparity  $\disparity$, and  $\displaystyle \hat\disparity = \arg\max_{\disparity} \costvolume(x, \disparity)$.  The sub-pixel MAP is only used for inference.  Tulyakov \etal~\cite{tulyakov2018practical} also showed that, unlike the softargmin/max, this  approach allows changing the disparity range at runtime without   re-training the network.

\subsection{Variants}
\label{sec:end2end_variants}

The pipeline described so far infers disparity maps that can be of low-resolution  (along the width, height, and disparity dimensions), incomplete, noisy, missing fine details, and suffering from over-smoothing especially at object boundaries.  As such, many variants have been introduced to  \textbf{(1)} improve their resolution (Section~\ref{sec:hiresdiisparity}),  \textbf{(2)} improve the processing time, especially at runtime (Section~\ref{sec:realtime}),  and \textbf{(3)} perform disparity completion and denoising (Section~\ref{sec:disp_completion}).

\subsubsection{Learning to infer high resolution disparity maps}
\label{sec:hiresdiisparity}

Directly regressing high-resolution depth maps that contain fine details, \eg by adding further upconvolutional layers  to upscale the cost volume, would require a large number of parameters and thus are computationally expensive and difficult to train.  As such, state-of-the-art methods struggle to process high resolution imagery because of memory constraints or speed limitations. This has been addressed  by using either bottom-up or top-down techniques.

\textbf{Bottom-up techniques}  operate in a sliding window-like approach. They take small patches and estimate the refined disparity either for the entire patch or for the pixel at the center of the patch.  Lee \etal~\cite{Lee_2018_CVPR} follow a split-and-merge approach. The input image is split  into regions, and a depth is estimated for each region. The estimates are then merged using a fusion network, which operates in the Fourier domain so that depth maps with different cropping ratios can be handled. While both sliding window and split-and-merge approaches reduce memory requirements, they require multiple forward passes, and thus are not suitable for realtime applications. Also, these methods do not capture the global context, which can limit their performance.


\textbf{Top-down techniques}, on the other hand, operate on the disparity map estimates in a hierarchical manner.  They  first estimate a low-resolution disparity map and then upsample them  to the desired resolution, \eg using bilinear upsampling,  and further process them  using residual learning to recover small details and thin structures~\cite{khamis2018stereonet,zhang2018activestereonet,chabra2019stereodrnet}. This process can also be run progressively by cascading many of such refinement blocks,  each block refines the estimate of the previous block~\cite{pang2017cascade,khamis2018stereonet}. Unlike upsampling cost volumes, refining disparity maps is computationally efficient since it only requires 2D convolutions. Existing methods mainly differ in the type of additional information that is appended to the upsampled disparity map for refinement. For instance:
\begin{itemize}
	\item Khamis \etal~\cite{khamis2018stereonet} concatenate the upsampled disparity map with the original reference image.
	\item Liang \etal~\cite{liang2018learning} append to the initial disparity  map the cost volume and the reconstruction error, defined as the difference between the left image and the right image but warped to the left image using the estimated disparity map. 
	\item Chabra \etal~\cite{chabra2019stereodrnet}  take the left image and the reconstruction error on one side, and the left disparity and the geometric error map, defined as the difference between the estimated left disparity and right disparity but warped onto the left view. These are independently filtered using one layer of convolutions followed by batch normalization. The results of the two streams are concatenated and then further processed using a series of convolutional layers to produce the refined disparity map.
\end{itemize}


\noi These methods improve the spatial resolution but not the disparity resolution.  To refine both the spatial and depth resolution, while operating on high resolution images, Yang \etal~\cite{Yang_2019_CVPR} propose to search for correspondences incrementally over a coarse-to-fine hierarchy. The approach constructs a pyramid of four 4D cost volumes, each with increasing spatial and depth resolutions.  Each  volume is  filtered by six 3D convolution blocks, and further processed with a Volumetric Pyramid Pooling block, an extension of Spatial Pyramid Pooling to feature volumes, to generate features that capture sufficient global context for high resolution inputs.  The output is then either  \textbf{(1)} processed with another conv3D block to generate a 3D  cost volume from which disparity can be directly regressed. This allows to report  on-demand disparities computed from the current scale, or \textbf{(2)} tri-linearly-upsampled to a higher spatial and disparity resolution so that it can be fused with the next 4D  volume in the pyramid.   To minimise memory requirements, the approach uses striding along the disparity dimensions in the last and second last volumes of the pyramid.  The network is trained end-to-end using a multi-scale loss. This hierarchical design  also allows for anytime on-demand reports of disparity by capping intermediate coarse results, allowing  accurate predictions for near-range structures with low latency (30ms).  

This approach shares some similarities with the approach of Kendall \etal~\cite{kendall2017end}, which  constructs hierarchical 4D feature volumes and processes them from coarse to fine using 3D convolutions.  Kendall \etal's approach~\cite{kendall2017end}, however,  has been used to leverage context with a wide field of view while  Yang \etal~\cite{Yang_2019_CVPR} apply  coarse-to-fine principles for high-resolution inputs and anytime, on-demand processing.

\subsubsection{Learning for completion and denoising}
\label{sec:disp_completion}

Raw disparities can  be noisy and incomplete, especially near object boundaries where depth smearing between objects remains a challenge. Several techniques have been developed for denoising and completion.  Some of them are ad-hoc, \ie post-process the noisy and uncomplete initial estimates to generate clean and complete depth maps. Other methods addressed the issue of the lack of training data for completion and denoising. Others proposed novel depth representations that are more suitable for this task, especially for solving  the depth smearing between objects.

\textbf{Ad-hoc methods} process the initially estimated disparities a  using variational approaches~\cite{dosovitskiy2015flownet,brox2011large}, Fully-Connected CRFs (DenseCRF)~\cite{huang2018deepmvs,krahenbuhl2011efficient},  hierarchical CRFs~\cite{li2015depth}, and diffusion processes~\cite{chen2015deep} guided by confidence maps~\cite{sun2014real}.  They encourage pixels that are spatially close and with similar colors to have closer disparity predictions.   They have been also explored by Liu \etal~\cite{liu2016learning}. However, unlike Li \etal~\cite{li2015depth},  Liu \etal~\cite{liu2016learning}  used a CNN to minimize the CRF energy.  Convolutional Spatial Propagation Networks (CSPN)~\cite{liu2017learningaffinity,cheng2018learning}, which implement an anisotropic diffusion process,  are particularly suitable for depth completion since they predict the diffusion tensor using a deep CNN. This is then applied to the initial  map to obtain the refined one.  

One of the main challenges of deep learning-based depth completion and denoising is the lack of \textbf{labelled training data}, \ie pairs of noisy, incomplete depth maps and their corresponding clean depth maps. To address this issue, Jeon and Lee~\cite{Jeon_2018_ECCV} propose a pairwise depth image dataset generation method using dense 3D surface reconstruction
with a filtering method to remove low quality pairs. They also present a multi-scale Laplacian pyramid based neural network and structure preserving loss functions to progressively reduce the noise and holes from coarse to fine scales. The approach first predicts the clean complete depth image at the coarsest scale, which has a quarter of the original resolution. The predicted  depth map  is then progressively upsampled through the pyramid to predict the half and original-sized image.   At the coarse level, the approach captures global context while at finer scales it captures local information.  In addition, the features extracted during the downsampling are passed to the upsampling pyramid with skip connections to prevent the loss of the original details in the input depth image during the upsampling.

Instead of operating on the network architecture, the loss function, or the training datasets, Imran \etal~\cite{Imran_2019_CVPR }  propose a \textbf{new representation for depth called Depth Coefficients (DC)} to address the problem of depth smearing between objects.  The representation enables convolutions to more easily avoid inter-object depth mixing.  The representation uses a multi-channel image of the same size as the target depth map, with each channel representing a fixed depth. The depth values increase in even steps of size $b$. (The approach uses $80$ bins.) The choice of the number of bins  trades-off memory vs. precision.  The vector composed of all these values at a given pixel defines the depth coefficients for that pixel. For each pixel, these coefficients are constrained to be non-negative and sum to $1$.   This representation of depth provides a much simpler way for CNNs to avoid  depth mixing.  First, CNNs can learn to avoid mixing depths in different channels as needed.   Second, since convolutions apply to all channels simultaneously, depth dependencies, like occlusion effects, can be modelled and learned by neural networks. The main limitation, however, is  that the depth range needs to be set in advance and cannot be changed at runtime without re-training the network.  Imran \etal~\cite{Imran_2019_CVPR }  also show that the standard Mean Squared Error (MSE) loss function can promote depth mixing, and thus propose to use cross-entropy loss for estimating the depth coefficients. 


\subsubsection{Learning for realtime processing}
\label{sec:realtime}

The goal  is to design  efficient stereo algorithms that not only produce reliable and accurate estimations, but also run in realtime.  For instance, in the PSMNet~\cite{chang2018pyramid}, the  cost volume construction and aggregation takes more than $250$ms (on nNvidia Titan-Xp GPU). This renders  realtime applications infeasible. To speed the process, Khamis \etal~\cite{khamis2018stereonet} first estimate a low resolution disparity map and then hierarchically refine it. Yin \etal~\cite{yin2019hierarchical} employ a fixed, coarse-to-fine procedure to iteratively find the match.  Chabra \etal~\cite{chabra2019stereodrnet} use 3D dilated convolutions  in the width, height, and disparity channels when filtering the cost volume. Duggal \etal~\cite{Duggal_2019_ICCV}  combine deep learning with PatchMatch~\cite{barnes2009patchmatch} to adaptively prune out the potentially large search space and significantly speed up inference. PatchMatch-based pruner module  is able to predict a confidence range for each pixel, and construct a sparse cost volume that requires significantly less operations. This also allows the model to focus only on regions with high likelihood and save computation and memory. To enable end-to-end training,  Duggal \etal~\cite{Duggal_2019_ICCV}  unroll  PatchMatch as an RNN where each unrolling step is equivalent to an iteration of the algorithm.  This approach  achieved a performance that is comparable to the state-of-the-art, \eg~\cite{chang2018pyramid,yang2018segstereo},  while reducing the computation time from $600$ms to $60$ms per image in the KITTI2015 dataset.

\subsection{Learning confidence maps}
\label{sec:confidencemaps}

The ability to detect, and subsequently remedy to,  failure cases is important for applications such as autonomous driving and medical imaging. Thus, a lot of research has been dedicated to estimating confidence or uncertainty maps, which are then used to sparsify the estimated disparities by removing potential errors and then replacing them from the reliable neighboring pixels.  Disparity maps can also be incorporated in a disparity refinement pipeline to guide the refinement process~\cite{seki2016patch,gidaris2017detect,jie2018left}.  Seki \etal~\cite{seki2016patch}, for example,  incorporate the confidence map  into a Semi-Global Matching (SGM) module for dense disparity estimation. Gidaris \etal~\cite{gidaris2017detect} use confidence maps to detect the incorrect estimates, replace them with disparities from neighbouring regions, and then refine the disparity using a refinement network. Jie \etal~\cite{jie2018left}, on the other hand,  estimate two confidence maps, one for each of the input images,  concatenate them with their associated cost volumes, and use them  as input  to a 3D convolutional LSTM  to selectively focus in  the subsequent step on the left-right mismatched regions.


Conventional confidence estimation methods  are mostly based on assumptions and heuristics on the matching cost volume analysis, see~\cite{hu2012quantitative} for a review and evaluation of the early methods.  Recent techniques are based on supervised learning~\cite{haeusler2013ensemble,spyropoulos2014learning,park2015leveraging,poggi2016learningfrom,wannenwetsch2017probflow,batsos2018cbmv}.  They estimate confidence maps directly from the disparity space either in an ad-hoc manner, or in an integrated fashion so that they can be trained end-to-end along with the disparity/depth estimation. Poggi \etal~\cite{poggi2017quantitative} provide  a quantitative evaluation.  Below, we discuss some of these techniques.



\subsubsection{Confidence from left-right consistency check}  Left-right consistency  is one of the most commonly-used criteria for measuring confidence in disparity estimates.  The idea is to  estimate two disparity maps, one from the left image ($\leftdisparitymap$), and another  from the right image ($\rightdisparitymap$). An error map can then be computed by taking a pixel-wise difference between $\leftdisparitymap$ and $\rightdisparitymap$, but warped back onto the left image, and converting them into probabilities~\cite{liang2018learning}.  This measure is  suitable for detecting occlusions, \ie regions that are visible in one view but not in the other.

Left-right consistency can  also be learned using  deep or shallow  networks composed of fully convolutional layers~\cite{seki2016patch,jie2018left}.  Seki \etal~\cite{seki2016patch} propose a patch-based confidence prediction (PBCP) network, which requires two disparity maps, one estimated from the left image and the other one from the right image.  PBCP uses a two-channel network. The first channel enforces left-right consistency while the second one enforces local consistency. The network is trained in a classifier manner. It outputs a label per pixel indicating whether the estimated disparity  is correct.

Instead of treating  left-right consistency check as an isolated post-processing step, Jie \etal~\cite{jie2018left} perform it  jointly with disparity estimation, using a  Left-Right Comparative Recurrent (LRCR) model. It   consists of two parallel  convolutional LSTM networks~\cite{xingjian2015convolutional}, which produce   two error maps; one for the left disparity and another for the right disparity.  The two error maps are then concatenated with their associated cost volumes and used as input  to a 3D convolutional LSTM  to selectively focus in  the next step on the left-right mismatched regions.


\subsubsection{Confidence from a single raw disparity map} 

Left-right consistency checks estimate two disparity maps and thus are expensive at runtime.  Shaked and Wolf~\cite{shaked2017improved} train, via the binary cross entropy loss, a network,  composed of two fully-connected layers,  to predict the correctness of an estimated disparity from only  the reference image.  Poggi and Mattoccia~\cite{poggi2016learningfrom} pose the confidence estimation as a regression problem and solve it using a CNN trained on small patches. For each pixel, the approach extracts a square patch around the pixel  and forwards it to a CNN trained to distinguish between patterns corresponding to correct and erroneous disparity assignments. It is a single channel network, designed for $9\times9$ image patches.  Zhang \etal~\cite{zhang2018activestereonet} use a similar confidence map estimation network, called  \emph{invalidation network}. The key idea is to train the network to predict confidence using a pixel-wise error between the left disparity and the right disparity. At runtime, the network only requires the left disparity.  Finally, Poggi and Mattoccia~\cite{poggi2017learning} show that one can improve the confidence maps estimated using previous algorithms by enforcing local consistency in the confidence estimates.


\subsubsection{Confidence map from matching densities}   Traditional deep networks represent  activations and outputs as deterministic point estimates. Gast and Roth~\cite{gast2018lightweight} explore the possibility of replacing the deterministic outputs by probabilistic output layers. To go one step further, they   replace all intermediate activations by distributions. As such, the network can be used to estimate the matching probability densities, hereinafter referred to as \emph{matching densities}, which can then be converted into uncertainties (or confidence) at runtime. The main challenge of estimating matching densities is the computation time.  To make it tractable, Gast and Roth~\cite{gast2018lightweight} assume parametric distributions.  Yin \etal~\cite{yin2019hierarchical} relax this assumption and propose a pyramidal architecture to make the computation cost sustainable and allow for the estimation of confidence at run time.

\subsubsection{Local vs. global reasoning}  Some techniques, \eg Seki \etal~\cite{seki2016patch}'s, reason locally by enforcing  local consistency. Tosi \etal~\cite{tosi2018beyond} introduced LGC-Net to move beyond local reasoning. The input reference image and its disparity map  are forwarded to a local network, \eg C-CNN~\cite{poggi2016learningfrom},  and a global  network, \eg an encoder/decoder architecture with  a large receptive field.  The output of the two networks and the initial disparity, concatenated with the reference image, are further processed with    three independent convolutional towers whose outputs are concatenated and processed with three $1\times 1$ convolutional layers to finally infer the confidence map. 

\subsubsection{Combining multiple estimators}

Some papers  combine the estimates of multiple algorithms to achieve a better  accuracy. Haeusler \etal~\cite{haeusler2013ensemble} fed a random forest with a pool of $23$   confidence maps, estimated using conventional techniques, yielding a much better accuracy compared  to any confidence map  in the pool.  Batsos \etal~\cite{batsos2018cbmv} followed a similar idea but combine the strengths and mitigate the weaknesses of four basic stereo matchers in order to generate a robust matching volume for the subsequent optimization and regularization steps.   Poggi and Mattoccia~\cite{poggi2016learning} train an ensemble regression trees classifier.  These methods are independent of the disparity estimation module,  and rely on the availability of the cost volume.

\section{Learning multiview stereo}
\label{sec:mvs_architectures}

\begin{figure}[t]
\centering{
	\begin{tabular}{|@{ }c@{ }|@{ }c@{ }|}
		\hline
		\includegraphics[width=0.23\textwidth]{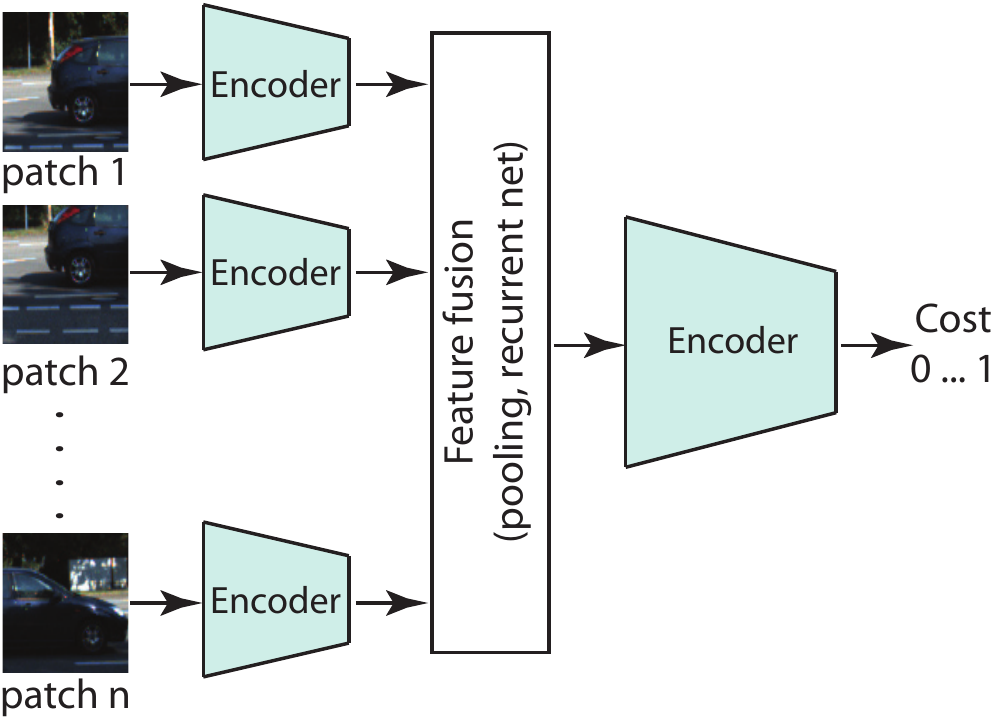} & \includegraphics[width=0.23\textwidth]{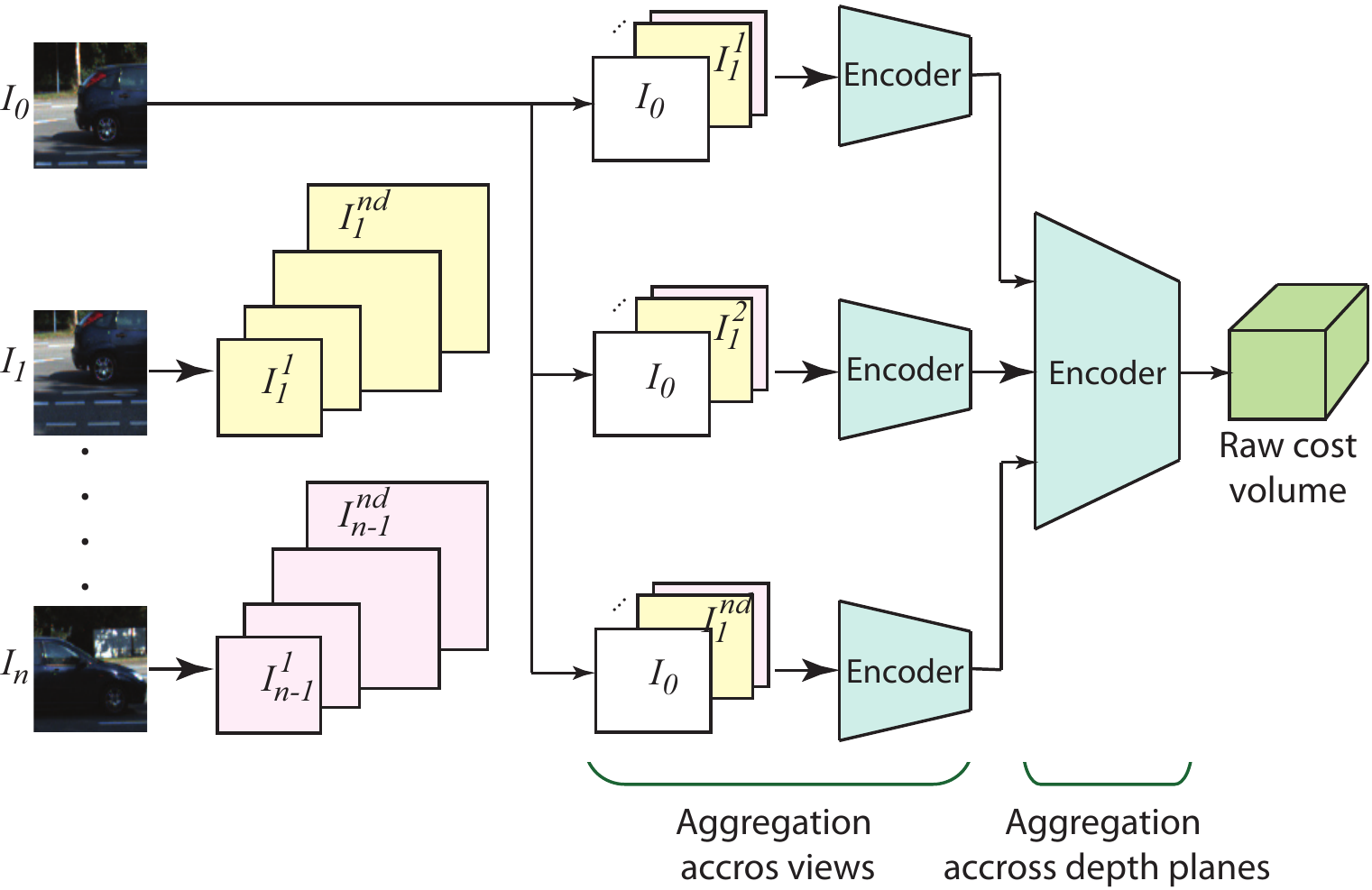} \\
		\small{(a) Hartmann \etal~\cite{hartmann2017learned}.} & \small{(b) Flynn \etal~\cite{flynn2016deepstereo}.} \\
		\hline
		\multicolumn{2}{|c|}{\includegraphics[width=0.46\textwidth]{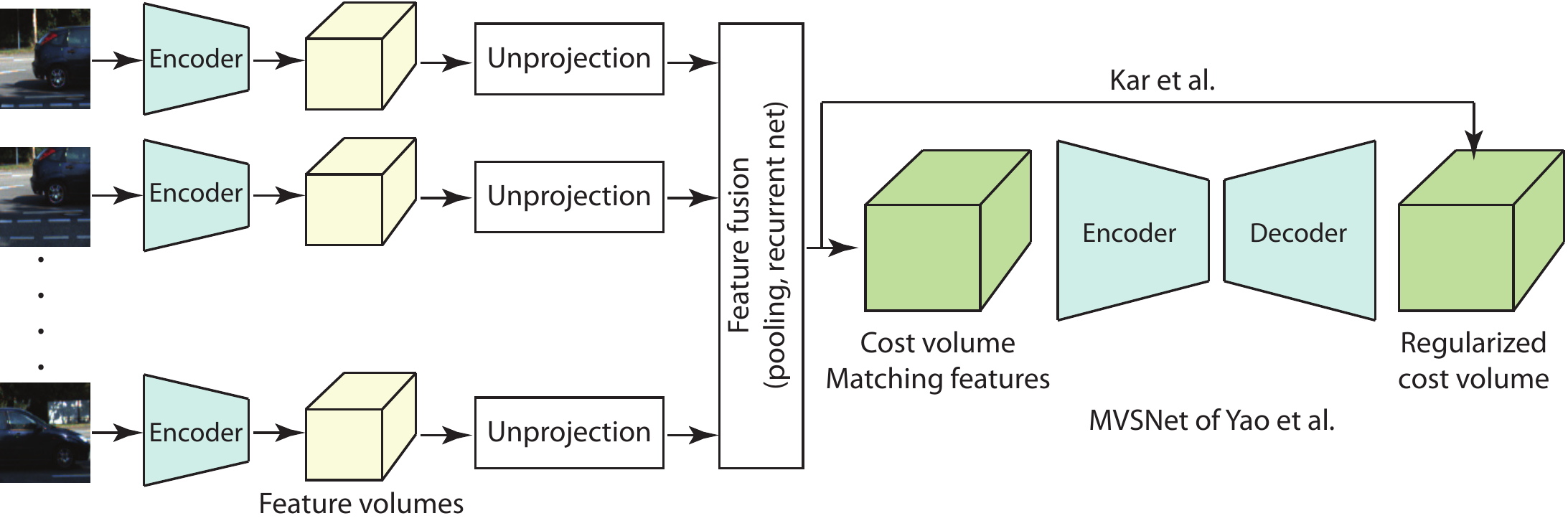}}\\
		\multicolumn{2}{|c|}{\small{(c) Kar \etal~\cite{kar2017learning} and Yao \etal~\cite{yao2018mvsnet}.}}\\
		\hline
		\multicolumn{2}{|c|}{}\\
		 \multicolumn{2}{|c|}{\includegraphics[width=0.46\textwidth]{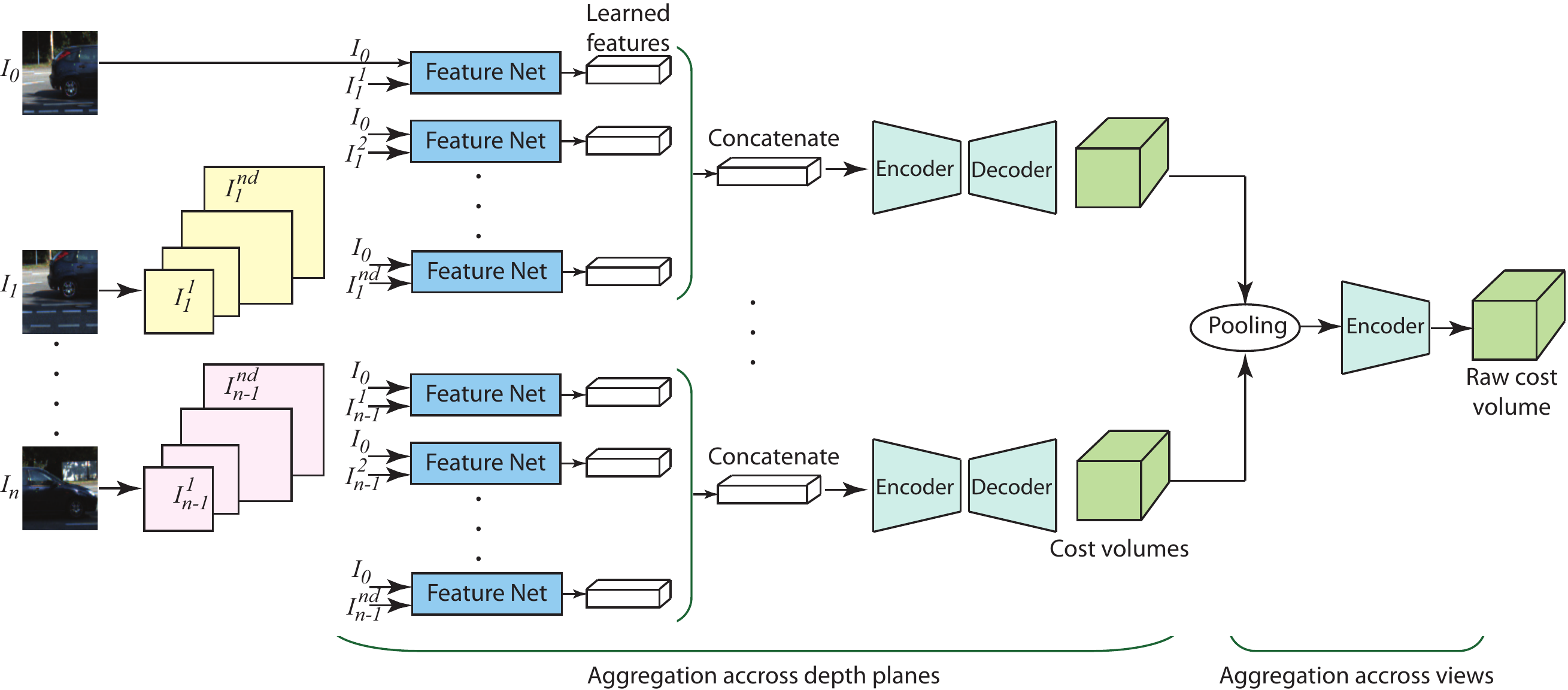}}\\
		 \multicolumn{2}{|c|}{\small{(d) Huang \etal~\cite{huang2018deepmvs}.} }\\
		\hline		
	\end{tabular}
	\caption{\label{fig:mvs} Taxonomy of multivew stereo methods. (a), (b), and (c) perform early fusion, while (d) performs  early fusion by aggregating features across depth plans, and late fusion by aggregating cost volumes across views. }

}
\end{figure}

Multiview Stereo (MVS) methods follow the same pipeline as of depth-from-stereo. Early works focused on  computing the similarity between multiple patches. For instance, Hartmann \etal~\cite{hartmann2017learned} (Fig.~\ref{fig:mvs}-(a)) replace the pairwise correlation layer used in stereo matching by an average pooling layer to aggregate the learned features of  $\nimages \ge 2$ input patches, and then feed the output to a top network, which returns a matching score. With this method, computing the best match for a pixel on the reference image requires $\ndisparities^{\nimages -1}$ forward passes. ($\ndisparities$ is the number of depth levels and $\nimages$ is the number of images.) This is computationally very expensive especially when dealing with  high resolution images. 

Techniques that compute depth maps in a single forward pass  differ in the way the information from the multiple views is fed to the network and  aggregated. We classify them into whether they are volumetric (Section~\ref{sec:mvs_volumetricbased}) or Plane-Sweep Volume (PSV)-based (Section~\ref{sec:mvs_depthbased}). The latter does not rely on intermediate volumetric representations of the 3D geometry. The only exception is the approach of  Hou \etal~\cite{Hou_2019_ICCV}, which performs temporal fusion of the latent representations of the input images. The approach, however, requires temporally-ordered images.  Table~\ref{tab:taxonomy_end_to_end_mvs} provides a taxonomy and compares $13$ state-of-the-art MVS techniques. 

\subsection{Volumetric representations}
\label{sec:mvs_volumetricbased}

One of the main issues for MVS reconstruction is how to  match,  in an efficient way, features across  multiple images. Pairwise stereo methods  rectify the images so that the search for correspondences is restricted to the horizontal epipolar lines. This is  not possible with MVS due to the large view angle differences between the images. This has bee addressed using volumetric representations of the scene geometry~\cite{ji2017surfacenet,kar2017learning}. Depth maps are then generated  by projection from the desired viewpoint.  For  a given input image, with known camera parameters, a ray from the viewpoint is cast through each image pixel. The voxels intersected by that ray are assigned the color~\cite{ji2017surfacenet}  or the learned feature~\cite{kar2017learning} of that pixel. Existing methods differ in the way information from multiple views are fused:

\vspace{6pt}
\noi \textit{(1) Fusing feature grids. } Kar \etal~\cite{kar2017learning} (Fig.~\ref{fig:mvs}-(c)) fuse, recursively, the back-projected 3D feature grids using  a recurrent neural network (RNN). The produced 3D grid is  regularized using an encoder-decoder. To avoid dependency on the order of the images, Kar \etal~\cite{kar2017learning} randomly permute the input images during training while constraining the output to be the same. 
	
\vspace{6pt}
\noi\textit{(2) Fusing pairwise cost volumes. } Choi \etal~\cite{choi2018learning}  fuse  the cost volumes, computed from each pair of images, using a weighted sum where the weight of each volume is the confidence map computed from that cost volume. 
	
\vspace{6pt}
\noi\textit{(3) Fusing the reconstructed surfaces. } Ji \etal~\cite{ji2017surfacenet}   process each pair of volumetric grids using a 3D CNN, which classifies whether a voxel is a surface point or not. To avoid the exhaustive combination of every possible image pairs, Ji \etal~\cite{ji2017surfacenet}  learn their relative importance, using a network composed of  fully-connected layers,  automatically select a few view pairs based on their relative importance to reconstruct multiple volumetric grids, and take their weighted sum to produce the final 3D reconstruction.

 To handle high resolution volumetric grids,  Ji \etal~\cite{ji2017surfacenet} split the whole space into small Colored Voxel Cubes (CVCs) and regress the surface  cube-by-cube. While this reduces the memory requirements, it requires multiple forward passes and thus increases the computation time.   Paschalidou \etal~\cite{Paschalidou_2018_CVPR}  avoid the explicit use of the volumetric representation. Instead, each voxel of the grid is projected onto each of the input views,  before computing the pairwise correlation between the corresponding learned features on each pair of views, and then averaging them over all pairs of views. Repeating this process for each depth value will result in the depth distribution on each pixel. This depth distribution is regularized using an MRF formulated as a differentiable function to enable end-to-end training.

In terms of performance, the volumetric approach of Ji \etal~\cite{ji2017surfacenet}   requires $4$ hours to obtain a full reconstruction of a typical scene in DTU dataset~\cite{aanaes2016large}. The approach of Paschalidou \etal~\cite{Paschalidou_2018_CVPR} takes approximately $25$mins, on an Intel i7 computer with an Nvidia GTX Titan X GPU, for the same task. Finally, methods   that perform fusion post-reconstruction have higher reconstruction errors compared to those that perform early fusion.

\subsection{Plane-Sweep Volume representations}
\label{sec:mvs_depthbased}


These methods directly estimate depth maps from the input without using  intermediate volumetric representations of the 3D geometry. As such, they are computationally more efficient. The main challenge to address is how to efficiently match features across multiple views in a single forward pass.   This is done by using the  Plane-Sweep Volumes (PSV)~\cite{flynn2016deepstereo,huang2018deepmvs,leroy2018shape,yao2018mvsnet,Luo_2019_ICCV,Xue_2019_ICCV}, \ie they back project the input  images~\cite{flynn2016deepstereo,huang2018deepmvs,leroy2018shape}  or their learned features~\cite{yao2018mvsnet,Luo_2019_ICCV,Xue_2019_ICCV} into planes at different depth values, forming PSVs from which the depth map is estimated.  Existing methods differ in the way the PSVs are processed with the feature extraction and feature matching blocks.



Flynn \etal's network~\cite{flynn2016deepstereo} (Fig.~\ref{fig:mvs}-(b)) is composed of $\ndisparities$ branches, one for each depth plane. The $\depth-$th  branch of the network takes as input the reference image and the planes of the PSVs of the other images  which are  located at  depth $\depth$. These are packed together and fed to a two-stage network. The first stage computes  matching features between the reference image and all the PSV planes located at depth $\depth$. The second stage models   interactions across depth planes using convolutional layers.   The final block of the network is a per-pixel softmax over depth, which returns the most probable depth value per pixel.  The approach  requires that the number of views and the camera parameters of each view to be known. 

Huang \etal~\cite{huang2018deepmvs}'s approach (Fig.~\ref{fig:mvs}-(d))  starts with a pairwise matching step where a cost volume is computed between the reference image and each of the input images. For a given pair $(\image_1, \image_i), i=2, \dots, \nimages$,  $ \image_i$ is first back-projected into a PSV. A siamese network then computes a matching cost volume between $\image_1$ and each of the PSV planes. These volumes are  aggregated into a single cost volume using an encoder-decoder network. This is referred to as intra-volume aggregation.  Finally a max-pooling layer is used to aggregate the multi intra-volumes into a single inter-volume, which is then used to predict the depth map. Unlike Flynn \etal~\cite{flynn2016deepstereo}, Huang \etal~\cite{huang2018deepmvs}'s approach does not require a fixed number of input views since  aggregation is performed using pooling.  In fact, the number of views can vary between training and at runtime. 

Unlike~\cite{flynn2016deepstereo,huang2018deepmvs}, which back-project the input images, the MVSNet of Yao \etal~\cite{yao2018mvsnet} use the camera parameters to back-project  the learned features into  a 3D frustum of a reference camera sliced into parallel frontal planes, one for   each depth value. The approach then generates the matching cost volume upon a pixel-wise variance-based metric, and finally a generic 3D U-Net is used to regularize the matching cost volume to estimate the depth maps. Luo \etal~\cite{Luo_2019_ICCV} extend MVSNet~\cite{yao2018mvsnet} to P-MVSNet in two ways. \textbf{First}, a raw cost volume is processed with a learnable patch-wise aggregation function before feeding it to the regularization network. This improves the matching robustness and accuracy for noisy data. \textbf{Second}, instead of using a generic 3D-UNet network for regularization, P-MVSNet uses a hybrid isotropic-anisotropic 3D-UNet. The plane-sweep volumes are essentially anisotropic in depth and spatial directions, but they are often approximated by isotropic cost volumes, which could be detrimental. In fact,  one can infer the corresponding depth map along the depth direction of the matching cost volume, but cannot get the same information along other directions.  Luo \etal~\cite{Luo_2019_ICCV} exploit  this fact, through the proposed  hybrid 3D U-Net with isotropic and anisotropic 3D convolutions,  to guide the regularization of matching confidence volume.

The main advantage of using PSVs is that they eliminate the need to supply rectified images.  In other words, the camera parameters are implicitly encoded.  However, in order to compute the PSVs, the intrinsic and extrinsic camera parameters need to be either provided in advance or estimated using, for example, Structure-from-Motion techniques as in~\cite{huang2018deepmvs}.  Also, these methods require setting in advance the disparity range and its discretisation.  Moreover, they often result in a complex network architecture.  Wang \etal~\cite{wang2018mvdepthnet}  propose a light-weight architecture. It stacks  together the reference image and the cost volume, computed using the absolute difference between the reference image and each other image but at different depth planes,  and feeds them to an encoder-decoder network, with skip connections, to estimate the inverse depth at three different resolutions.  Wang \etal~\cite{wang2018mvdepthnet}  use a view selection rule, which selects the frames that have enough angle or translation difference and then use the selected frames to compute the cost volume.

Finally, note that feature back-projection has been also used   by Won \etal~\cite{Won_2019_ICCV}  for omnidirectional depth estimation from a wide-baseline multi-view stereo setup. The approach uses spherical maps and spherical cost volumes.



\begin{table*}
	\caption{\label{tab:taxonomy_end_to_end_mvs} Taxonomy and comparison of $13$  deep learning-based MVS  techniques.
	}
	
	\resizebox{\linewidth}{!}{%
	\begin{tabular}{@{}l@{ }l @{ }c@{ }c @{ }c   @{ }c   @{ }c@{ }c@{ }c@{ }c@{ }c @{ }c@{ }c@{ }c@{ }c@{}}
	\toprule
	\multirow{2}{*}{\textbf{Method}}  &  \multirow{2}{*}{\textbf{Year}} &  \multirow{2}{*}{\textbf{Representation}} & \multirow{2}{*}{\textbf{Fusion}} &  \multirow{2}{*}{\textbf{Training}} && \multicolumn{4}{c}{\textbf{Peformance on (DTU, SUN3D, ETH3D)} }  & & \multicolumn{4}{c}{\textbf{Complexity}}  \\
	\cline{7-10} \cline{12-15} 
	
	 &  & & & & & \textbf{\#images} &  \textbf{Error ($mm$)}&  \textbf{\% $< 1mm$} &  \textbf{\% $<2mm$} &  & \textbf{\# Params} & \textbf{Memory} & \textbf{Complexity} & \textbf{Time (s)} \\
								
	\midrule

	Kar \etal~\cite{kar2017learning}& 2017 & Volumetric & Recurrent fusion & Supervised& & Variable &   $-$ & $-$ & $-$ &  & $-$ & $-$ & $-$ & $-$\\ 
	     & 		& 	& of 3D feature grids & \\
	\hline
	Hartmann \etal~\cite{hartmann2017learned}&  2017 & \multicolumn{2}{c}{Replace correlation by pooling} & Supervised& &  $5$&   $(1.356, -, -)$ &  $-$ & $-$ & & $-$& $-$& $-$& $-$ \\ 
									   & & & & & & (can vary) & \\
	\hline
	Ji \etal~\cite{ji2017surfacenet}  &  2017& Volumetric& Reconstructed surfaces & Supervised& & $5$  &  $(0.745, -, -)$   & $ 69.95$ &  $ 74.4$ & & $-$& $-$& $-$& $4$ hrs\\ 
	\hline

	Choi \etal~\cite{choi2018learning}& 2018 & Volumetric & Pairwise cost volumes &Supervised & & $5$ &  $(0.6511, -, -)$ &    $-$& $-$ &  & $-$ & $-$ &$-$ & $-$\\ 
	\hline
	Huang \etal~\cite{huang2018deepmvs}&  2018& PSV& Encoder-decoder for intra-volume,  & Supervised&  & Variable  &  $(-, 0.419,  0.412)$ & $-$& $-$ &  & $-$ & $-$ &$-$ & $-$ \\ 
	& 		& 	& Max pooling for inter-volume & \\
	\hline
	Leroy \etal~\cite{leroy2018shape}& 2018 & PSV & Depth fusion& Supervised&  &Variable& $(0.599, -, -)$ &   $-$ & $-$ &  & $72$K & $-$ & $-$ & $-$ \\
	\hline
Paschalidou \etal~\cite{Paschalidou_2018_CVPR}& 2018 & Depth-based& Avg. pooling over & Supervised& &  Variable &  $(-, -, -)$   & $-$& $-$& & $-$ & $7$GB &  $-$ & $25$ mins\\
										& & & pairwsie correlations & \\
	\hline
	Yao \etal~\cite{yao2018mvsnet}&  2018&  PSV& Feature pooling by variance & Supervised& & $5$ &   $(0.462, 0.397,0.470)$  & $75.69$ & $80.25$ & & $363$K & $5.28$GB &  $O(\height\times \width\times\ndisparities)$ & $0.9$s   \\  
	\hline
	Wang \etal~\cite{wang2018mvdepthnet}&2018  & PSV and abs.  & Concatenation of  pairwise&Supervised &  &  Variable&    $(-, 0.114, 0.257)$&  $-$ &  $-$ &  &$33.9$M for  & $-$  & $-$& $0.04$ \\ 
							      &        &	difference		& cost  volumes and ref. image &  & & &  &  &  & &$\ndisparities = 64$ &  &  & \\
	\hline

	Hou \etal~\cite{Hou_2019_ICCV}& 2019 & $-$&  Temporal fusion of &Supervised & & Variable  &   $(-, \textbf{0.101}, 0.229)$  & $-$ &$-$ & & $-$ & $-$ & $-$ & $-$  \\
								     & &  & the latent rep. &  & & (video sequence)\\
	\hline
Luo \etal~\cite{Luo_2019_ICCV}& 2019 & PSV & Feature pooling by variance& Supervised& & Variable &   $(0.406, - , - )$ & $-$ & $-$ &  &  $-$& $-$ & $-$ & $-$ \\  
	\hline
Xue \etal\etal~\cite{Xue_2019_ICCV}& 2019 &  PSV& Cost volume& Supervised & & $5$ &   $(\textbf{0.398}, -, -)$  &  $80.02$ & $83.84$ & & $571$K  & $5.43$GB & $O(\height\times\width\times \ndisparities)$ & $1.8$s\\  
		& & &  pooling by variance& & & (can vary) & \\
	\hline
 Won \etal~\cite{Won_2019_ICCV}& 2019 & Spherical PSV& Concatenation &Supervised & &$-$ & $-$ & $-$ & $-$ & & $-$ & $-$ & $-$ & $-$\\
	
	\bottomrule
	\end{tabular}
	}
\end{table*}

%
\section{Training end-to-end stereo methods}
\label{sec:training_end_to_end}

The training process aims to find the network parameters  $\weights$  that minimize a loss function $ \loss(\weights; \estimateddisparitymap, \Theta)$ where  $\estimateddisparitymap$ is the estimated disparity, and   $\Theta$ are the supervisory cues.  The loss function is  defined as the sum of a data term $\loss_1(\estimateddisparitymap, \Theta, \weights) $, which measures the discrepancy between  the ground-truth and the estimated disparity, and a regularization or smoothness term $ \loss_2(\estimateddisparitymap, \weights)$, which imposes  local or global constraints on the solution. The type of supervisory cues   defines the degree of supervision  (Section~\ref{sec:end2enddegree_supervision}), which can be  supervised with 3D groundtruth (Section~\ref{sec:3D_supervised}), self-supervised using auxiliary cues (Section~\ref{sec:self_supervised}), or weakly supervised (Section~\ref{sec:weakly_supervised}). Some methods  use additional cues, in the form of constraints on the solution, to boost the accuracy and performance (Section~\ref{sec:additional_cues}). One of the main challenges of deep learning-based techniques is their ability to generalize to new domains.  Section~\ref{sec:domain_adaptation} reviews methods that addressed this issue. Finally, Section~\ref{sec:learning_net_arch} reviews methods that learn network architectures.


\subsection{Supervision methods} 
\label{sec:end2enddegree_supervision}

\subsubsection{3D supervision  methods}
\label{sec:3D_supervised}
Supervised methods are trained to minimise a loss function that measures the error between the ground truth disparity and the estimated disparity.  It is of the form:
\begin{equation}
	\loss = \frac{1}{\npixels} \sum \confidencefunc(x) \threshfunc(\confidencefunc(\pixel) - \epsilon ) \Distance\left( \Phi( \disparity_\pixel), \Phi(\estimateddisparity_\pixel) \right), 
	\label{eq:dataterm_supervised}
\end{equation}

\noi where:  $\disparity_\pixel$ a $\estimateddisparity_\pixel$ are, respectively, the groundtruth and the estimated disparity at pixel $\pixel$.   $\Distance$ is a measure of distance, which can be the $\ltwo$, the  $\lone$~\cite{kendall2017end,ummenhofer2017demon,pang2017cascade,cheng2018learning},  the smooth $\lone$~\cite{chang2018pyramid}, or  the smooth $\lone$ but approximated using the two-parameter robust function $\rho(\cdot)$~\cite{khamis2018stereonet,barron2017more}.   $ \confidencefunc(\pixel) \in [0, 1] $ is the confidence of the estimated disparity at $\pixel$. Setting  $ \confidencefunc(\pixel)  = 1$ and the threshold  $\epsilon = 0, \forall \pixel$ is equivalent to ignoring the confidence map. 	$\threshfunc(x)$ is the heavyside function, which is equal to $1$ if $x \ge 0$, and $0$ otherwise. $ \Phi( \cdot)$ is either the identify  or the log function.   The latter  avoids overfitting the network to large disparities. 

Some papers restrict the sum in Eqn.~\eqref{eq:dataterm_supervised} to be  over only the valid pixels or regions of interest, \eg foreground or visible pixels~\cite{zhou2016learning},   to avoid outliers. Other, \eg Yao \etal~\cite{yao2018mvsnet}, divide the loss into two parts, one over the initial disparity and the other one over the refined disparity.  The overall loss is then defined as the weighted sum of the two losses.

\subsubsection{Self-supervised methods}
\label{sec:self_supervised}
Self-supervised methods, originally used in optical flow estimation~\cite{ahmadi2016unsupervised,jason2016back},  have been proposed as a possible solution in the absence of sufficient ground-truth training data.  These methods mainly rely on image reconstruction losses, taking advantage of the projective geometry, and the spatial and temporal coherence when multiple images of the same scene are available. The rationale is that if the estimated disparity map is as close as possible to the ground truth, then the discrepancy between the reference image and any of the other images but unprojected using the estimated depth map onto the reference image, is also minimized.  The general loss function is of the form:
    	\begin{equation}
    		\loss= \frac{1}{\npixels} \sum_{\pixel}  \Distance\left( \Phi\left(\referenceimage\right)(\pixel) -  \Phi\left(\tilde{\image}_{ref}\right)(\pixel) \right),
		\label{eq:unsupervised_loss}
    	\end{equation}

\noi where  $ \tilde{\image}_{ref}$, which is $\rightimage$  but unwarped onto $\referenceimage$ using the estimated disparity, and $\Distance$ is a measure of distance. The mapping function $ \Phi$  can be:
\begin{itemize}
	\item The identity~\cite{bai2016exploiting,zhou2017unsupervised,zhong2017self,yang2018segstereo}. In this case, the loss of Eqn.~\eqref{eq:unsupervised_loss} is called a photometric or image reconstruction loss. 
	
	\item A mapping to the  feature space~\cite{yang2018segstereo}, \ie $\Phi\left(\referenceimage\right) = \featuremap$ where $\featuremap$ is the learned feature map. 
	
	\item The gradient of the image, \ie $\Phi\left(\referenceimage\right) =\nabla \image_{ref}$, which  is less sensitive to variations in lighting and acquisition conditions than the photometric loss.
	
\end{itemize}

\noi The distance  $\Distance$ can be the $\lone$ or $\ltwo$ distance. Some papers~\cite{zhong2017self}  also use more complex metrics such as the structural  dissimilarity~\cite{wang2004image} between patches  in $\referenceimage$ and  in $ \tilde{\image}_{ref}$.

While stereo-based supervision methods do not require ground-truth 3D labels, they rely on the availability of  calibrated stereo pairs during training.

\subsubsection{Weakly supervised methods}
\label{sec:weakly_supervised}

Supervised methods for disparity estimation can achieve promising results if trained on large quantities of ground truth depth data. However, manually obtaining ground-truth depth data is  extremely difficult and expensive, and is prone to noise and inaccuracies. Weakly supervised methods rely on auxiliary signals to reduce the amount of manual labelling. In particular, Tonioni \etal \cite{tonioni2017unsupervised} used as a supervisory signal the depth estimated using traditional stereo matching techniques to fine-tune depth estimation networks. Since such depth data can be sparse, noisy, and prone to errors, they propose a confidence-guided loss that penalizes ground-truth depth values that are deemed not reliable.  It is defined using Eqn.~\eqref{eq:dataterm_supervised} by setting $\Distance(\cdot)$ to be the $\lone$ distance, and   $\epsilon > 0$. Kuznietsov \etal~\cite{kuznietsov2017semi} use  sparse ground-truth depth for supervised learning, while enforcing the deep network to produce photo-consistent dense depth maps in a stereo setup using a direct image alignment/reprojection loss.   These two methods rely on an ad-hoc disparity estimator. To avoid that, Zhou \etal~\cite{zhou2017unsupervisedlearing} propose an iterative approach, which starts with a randomly initialized network.  At each iteration,  it computes matches from the left  to the right images, and matches from the right  to the left images. It then selects the   high confidence matches and adds them as labelled data for further training in the subsequent iterations. The confidence is computed using the left-right consistency of Eqn.~\eqref{eq:leftrightconsistency}. 

\subsection{Incorporating additional cues} 
\label{sec:additional_cues}
Several works  incorporate additional cues and constraints to improve the quality of the disparity estimates. Examples include  smoothness~\cite{zhong2017self}, left-right consistency~\cite{zhong2017self},  maximum depth~\cite{zhong2017self}, and scale-invariant gradient loss~\cite{ummenhofer2017demon}. Such cues can also be in the form of auxiliary information such as semantic cues used to guide the disparity estimation network. Below, we discuss a number  of these works.

\vspace{6pt}
\noi\textit{(1) Smoothness. } In general, one can assume that neighboring pixels have similar disparity values. Such smoothness constraint can be enforced by minimizing:

\begin{itemize}
	\item  The absolute difference between the disparity predicted at $\pixel$ and those predicted at each pixel $y$ within a certain predefined neighborhood $\mathcal{N}_\pixel$ around   $\pixel$: 
\begin{equation}
	\loss= \frac{1}{\npixels} \sum_{\pixel} \sum_{y \in \mathcal{N}_\pixel} |\disparity_\pixel - \disparity_y |.  
\end{equation}
	Here, $\npixels$ is the total number of pixels. 
	\item The magnitude of the first-order gradient $\nabla$ of the estimated disparity map~\cite{yang2018segstereo}:
		\begin{equation}
			\small{ \loss = \frac{1}{N} \sum_x \left\{ (\nabla_u \disparity_x ) +  (\nabla_v \disparity_x)\right\}, \pixel = (u, v).}
			\label{eq:loss_gradient_diff}
		\end{equation}	
		
	\item The  magnitude of the second-order gradient of the estimated disparity~\cite{zhou2017unsupervised,vijayanarasimhan2017sfm}: 		
		\begin{equation}
			\loss= \frac{1}{N} \sum_x \left\{ (\nabla_u^2 \disparity_x )^2 +  (\nabla_v^2 \disparity_x)^2\right\}.
			\label{eq:smoothness_2ndgradient}
		\end{equation}	
	
	\item The second-order gradient of the estimated disparity map  weighted by the image's second-order gradients~\cite{zhong2017self}:
		
\end{itemize}	
		\begin{equation}
			\small{\loss = \frac{1}{N} \sum_{\pixel} \left\{ | \nabla_u^2 \disparity_x | e^{-| \nabla_u^2 \image(x)  |} +  | \nabla_v^2 \disparity_x | e^{-| \nabla_v^2 \image(x)  |} \right\}.}
			\label{eq:smoothness_2ndgradient2}
		\end{equation}	

\vspace{6pt}
\noi\textit{(2) Consistency. } Zhong \etal~\cite{zhong2017self}  introduced the loop-consistency loss, which is constructed as follows. Consider the left image $\leftimage$ and the  synthesized image $\tilde{\image}_{left}$ obtained by warping the right image to the left image coordinate using the disparity map defined on the right image. A second synthesized left image  $\tilde{\tilde{\image}}_{left}$  can also be generated by warping the left image to the right image coordinates, by using the disparities at the left image, and then warping it back to the left image using the disparity at the right image. The three versions of the left image provide two constraints: $\leftimage = \tilde{\image}_{left}$ and $\leftimage = \tilde{\tilde{\image}}_{left} $, which can be used to regularize the disparity maps. Godard \etal~\cite{godard2017unsupervised}  introduce the left-right consistency term, which is a linear approximation of the loop consistency. The loss attempts to make the left-view disparity map  equal to the projected right-view disparity map. It is defined as: 
		\begin{equation}
				\loss	 = \frac{1}{\npixels} \sum_{\pixel} | \disparity_{\pixel}  - \tilde{\disparity}_{\pixel} |,
				\label{eq:leftrightconsistency}
		\end{equation}

\noi where $\tilde{\disparity}$ is the disparity at the right image but reprojected onto the coordinates of the left image. 
		
\vspace{6pt}
\noi\textit{(3) Maximum-depth heuristic. }  There may be multiple warping functions that achieve a similar warping loss, especially for textureless areas. To provide strong regularization in these areas, Zhong \etal~\cite{zhong2017self}  use the Maximum-Depth Heuristic (MDH)~\cite{perriollat2011monocular}  defined as the sum of all depths/disparities:
		\begin{equation}
			\loss = \frac{1}{\npixels} \sum_{\pixel} | \disparity_{\pixel} |.
			\label{eq:mdh}
		\end{equation}
		
\vspace{6pt}
\noi\textit{(4) Scale-invariant gradient loss~\cite{ummenhofer2017demon}.}  It is defined as:  
		\begin{equation}
			\loss  = \sum_{h \in A} \sum_{\pixel} \| g_h[\disparitymap](\pixel)   - g_h[\estimateddisparitymap](\pixel) \|_2,
		\end{equation}
\noi where $A = \{1,2, 4, 8, 16 \}$,  $\pixel = (i, j)$, $f_{i,j} \equiv f(i, j)$, and
		\begin{equation}
			\small{g_h[f](i, j)  = \left( \frac{f_{i+h, j} - f_{i, j} }{| f_{i+h, j} - f_{i, j} |},  \frac{f_{i, j+h} - f_{i, j} }{| f_{i,  j+h} - f_{i, j} |} \right)^\top.}
		\end{equation}

\noi This loss penalizes relative depth errors between neighbouring pixels. This loss stimulates the network to compare depth values within a local neighbourhood for each pixel. It emphasizes depth discontinuities, stimulates sharp edges, and increases smoothness within homogeneous regions.

\vspace{6pt}
\noi \textit{(5) Incorporating semantic cues. }   Some papers incorporate additional cues such as normal~\cite{Qi_2018_CVPR}, segmentation~\cite{yang2018segstereo},  and edge~\cite{song2018stereo} maps, to guide the disparity estimation.  These can be either provided at the outset, \eg estimated with a separate method as in~\cite{song2018stereo}, or  estimated jointly with the disparity map. Qi \etal~\cite{Qi_2018_CVPR} propose a mechanism that uses the depth map to refine the quality of the normal estimates, and the normal map to refine the quality of the depth estimates. This is done using a  two-stream network: a depth-to-normal network for normal map refinement using the initial depth estimates, and a normal-to-depth network for depth refinement using the estimated normal map. 


Yang \etal~\cite{yang2018segstereo} and Song \etal~\cite{song2018stereo}  incorporate semantics by stacking semantic maps (segmentation masks in the case of~\cite{yang2018segstereo} and edge features in the case of~\cite{song2018stereo}) with the 3D cost volume. Yang \etal~\cite{yang2018segstereo}   train jointly a disparity estimation network and a segmentation network by using a loss function defined as a weighted sum of the reconstruction error, a smoothness term, and a segmentation error.   Song \etal~\cite{song2018stereo} further incorporate edge cues in the edge-aware smoothness loss to penalize  drastic depth changes in flat regions. Also, to allow for depth discontinuities at object boundaries,  the edge-aware smoothness loss is defined based on the gradient map  obtained from the edge detection sub-network, which is more semantically meaningful than the variation in raw pixel intensities.


Wu \etal~\cite{Wu_2019_ICCV} introduced an approach that fuses multiscale 4D cost volumes with semantic features obtained using a segmentation sub-network. The approach uses the features of the left and the right images  as input to a semantic segmentation network similar to PSPNet~\cite{zhao2017pyramid}.  Semantic features for each image are then obtained from the output of the classification layer of the segmentation network. A 4D semantic cost volume is  obtained by concatenating each unary semantic feature with their corresponding unary from the opposite stereo image across each disparity level. Both the spatial pyramid cost volumes and the semantic cost volume are fed into a 3D multi-cost aggregation module, which aggregates  them, using an encoder-decoder followed by a 3D feature fusion module,  into a single 3D cost volume in a pairwise manner starting with the smallest volume. 

					 
In summary, appending  semantic features to the  cost volume  improves the reconstruction of fine details, especially near object boundaries.

\subsection{Domain adaptation and transfer learning}
\label{sec:domain_adaptation}

\begin{figure}[t]
\centering{
	\includegraphics[trim=1cm 23.5cm 0.5cm .2cm, clip=true, width=\linewidth]{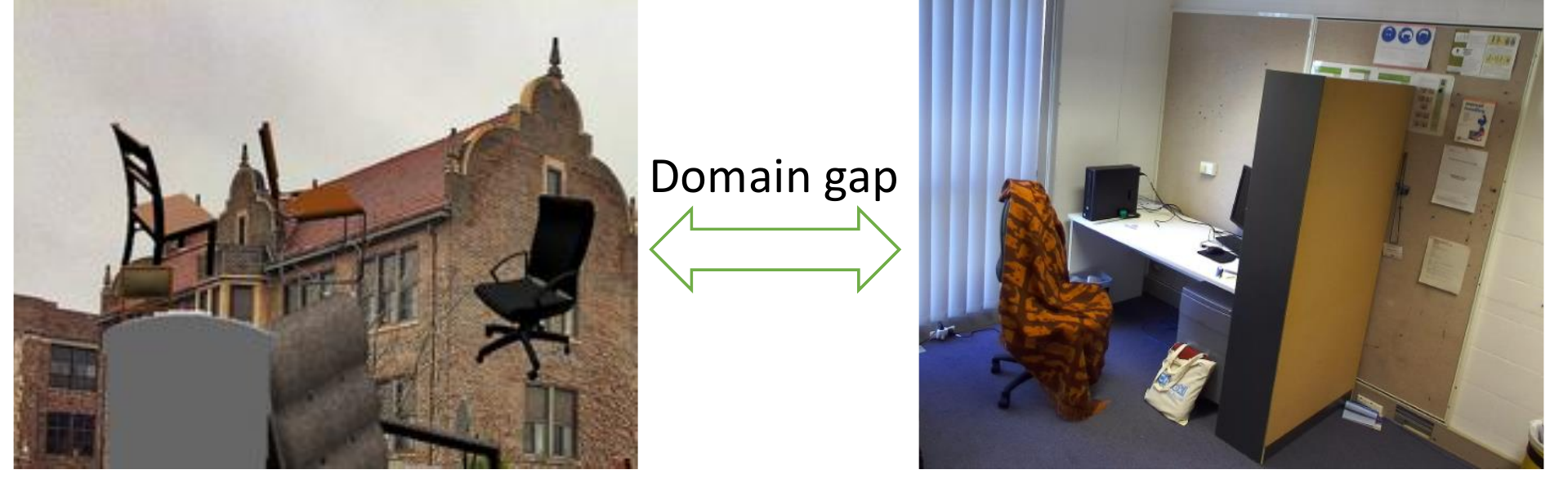}
	\caption{\label{fig:domain_gap} Illustration of the domain gap between synthetic (left)  and real (right)  images. The left image is from the FlyingThings synthetic dataset~\cite{mayer2016large}.
}
}
\end{figure}

Deep architectures for depth estimation are severely affected by the domain shift issue, which hinders their effectiveness when performing inference on images significantly diverse from those used during the training stage. This can be observed, for instance, when moving between indoor and outdoor environments, from synthetic to real data, see Fig.~\ref{fig:domain_gap}, or between different outdoor/indoor environments,  and when changing the camera model/parameters.  As such, deep learning networks trained on one domain, \eg by using synthetic data, suffer when applied to another domain, \eg real data, resulting in blurry object boundaries and errors in ill-posed regions such as object occlusions, repeated patterns, and textureless regions. These are referred to as \emph{generalization glitches}~\cite{pang2018zoom}.

Several strategies have been  proposed to adress this domain bias issue.  They can be classified into two  categories: adaptation by fine-tuning (Section~\ref{sec:adaptation_finetuning}) and adaptation by data transformation (Section~\ref{sec:adaptation_datatransformation}). In both cases, the adaptation can be offline or online.  



\subsubsection{Adaptation by fine-tuning}
\label{sec:adaptation_finetuning}


Methods in this category perform domain adaptation by first training a network on  images from a certain domain, \eg synthetic images as in~\cite{mayer2016large}, and then fine-tuning it on  images from a target domain. A major difficulty is to collect accurate ground-truth depth for stereo or multiview images  from the target domain.  Relying on  active sensors (\eg LiDAR) to obtain such supervised labeled data is not feasible in practical applications.  As such, recent works, \eg~\cite{tonioni2017unsupervised,tonioni2019unsupervised,pang2018zoom} rely on off-the-shelf stereo algorithms to obtain ground-truth disparity/depth labels in an unsupervised manner, together with state-of-the-art confidence measures  to ascertain the correctness of the measurements of the off-the-shelf stereo algorithms.  The latter is used in~\cite{tonioni2017unsupervised,tonioni2019unsupervised}  to discriminate between reliable and unreliable disparity measurements, to select the former and fine tune a pre-trained  model, \eg DispNet~\cite{mayer2016large}, using such smaller and sparse set of points as if they were ground-truth labels.



Pang \etal~\cite{pang2018zoom} also use a similar  approach as in~\cite{tonioni2017unsupervised,tonioni2019unsupervised} to address the generalization glitches. The approach, however, exploits the scale diversity, \ie up-sampling the stereo pairs enables the model to perform stereo matching in a localized manner with subpixel accuracy, by performing iterative optimisation of predictions obtained at multiple resolutions of the input. 

Note that  self-supervised and weakly supervised techniques for disparity estimation, \eg~\cite{zhou2017weakly,godard2017unsupervised,zhang2018deep,poggi2018towards}  can also be used for offline domain adaptation. In particular, if stereo pairs of the target domain are available, these techniques can be fine-tuned, in an unsupervised manner, using reprojection losses, see Sections~\ref{sec:self_supervised} and~\ref{sec:weakly_supervised}.

Although effective, these \textbf{offline} adaptation techniques reduce the usability of the methods since users are required to train the models every time they are exposed to a new domain. As a result, several recent papers developed \textbf{online} adaptation techniques.  For example,  Tonioni \etal~\cite{tonioni2019real} address the domain shift issue by casting adaptation as a continuous learning process whereby a stereo network can evolve online based on the images gathered by the camera during its real deployment.  This is achieved in an unsupervised manner by computing error signals on the current frames, updating the whole network by a single back-propagation iteration, and moving to the next pair of input frames. To keep a high enough frame rate, Tonioni \etal~\cite{tonioni2019real}  propose a lightweight, fast, and modular architecture, called MADNet, which allows training sub-portions of the whole network independently from each other. This allows  adapting disparity estimation  networks to unseen environments without supervision at approximately $25$ fps, while achieving an accuracy comparable to DispNetC~\cite{mayer2016large}.  Similarly, Zhong \etal~\cite{zhong2018open} use video sequences to train a deep network online from a random initialization. They employ an LSTM in their model to leverage the temporal information during the prediction.  

Zhong \etal~\cite{zhong2018open} and Tonioni \etal~\cite{tonioni2019real} consider  online adaptation separately from the initial training. Tonioni \etal~\cite{tonioni2019learning}, on the other hand, incorporate the adaptation procedure to the learning objective to obtain a set of initial parameters that are suitable for online adaptation, \ie they can be adapted quickly to unseen environments. This is implemented using    the model agnostic meta-learning framework of~\cite{finn2017model},  an explicit \emph{learn-to-adapt} framework that enables  stereo methods to adapt quickly and continuously  to new target domains in an unsupervised manner.

\subsubsection{Adaptation by data transformation}
\label{sec:adaptation_datatransformation}

Methods in this category transform the data of one domain to look similar in style to the data of the other domain. For example, Atapour-Abarghoue \etal~\cite{Atapour-Abarghouei_2018_CVPR} proposed a two-staged approach.  The first stage  trains a depth estimation model using synthetic data. The second stage is  trained to transfer the style of synthetic  images to real-world images. 	By doing so, the style of real images is first transformed to match the style of synthetic data and then fed into the depth estimation network, which has been trained on synthetic data.  Zheng \etal~\cite{Zheng_2018_ECCV} perform  the opposite by transforming the synthetic images to become more realistic and using them to train the depth estimation network.   Zhao \etal~\cite{Zhao_2019_CVPR} consider both synthetic-to-real~\cite{Zheng_2018_ECCV} and real-to-synthetic~\cite{Kundu_2018_CVPR,Atapour-Abarghouei_2018_CVPR} translations. 
The two translators are trained in an adversarial manner using an adversarial loss and a cycle-consistency loss. That is, a synthetic image when converted to a real image and converted back to  the synthetic domain should look similar to the original one.


Although these methods have been used for monocular depth estimation, they are applicable to (multi-view) stereo matching methods. 

\subsection{Learning the network architecture}
\label{sec:learning_net_arch}

Much research work in depth estimation is being spent on manually optimizing  network architectures, but what about if the optimal network architecture, along with its parameters, could  be also learnt  from data? Saika \etal~\cite{Saikia_2019_ICCV} show how to use and extend existing AutoML techniques~\cite{hutter2019automated} to efficiently optimize large-scale U-Net-like encoder-decoder architectures for stereo-based depth estimation. Traditional AutoML techniques have extreme computational demand limiting their usage to small-scale classification tasks. Saika \etal~\cite{Saikia_2019_ICCV}   applies Differentiable Architecture Search (DARTs)~\cite{liu2019darts}  to encoder-decoder architectures.  Its main idea is to have a large network that includes all architectural choices and to select the best parts of this network by optimization. This can be relaxed to a continuous optimization problem, which, together with the regular network training, leads to a bilevel optimization problem.  Experiments conducted on DispNet of~\cite{Ilg_2018_ECCV}, an improved version of~\cite{mayer2016large}, show that   the  automatically optimized DispNet (AutoDispNet) yields better performance compared to the  baseline DispNet of~\cite{Ilg_2018_ECCV}, with about the same number of parameters. The paper also shows that  the benefits of automated optimization  carry over to  large stacked networks. 


\section{Discussion and comparison}
\label{sec:discussion_and_comparison}
Tables~\ref{tab:taxonomy_end_to_end_stereomatching}  and~\ref{tab:taxonomy_end_to_end_mvs}, respectively,  compare the performance of the methods surveyed in this article on standard datasets such as KITTI2015 for pairwise stereo methods,  and DTU, SUN3D and ETH3D for multiview stereo methods.  Most of these methods have been trained on subsets of these publicly available datasets.   A good disparity estimation method, once properly trained, should achieve good performance not only on publicly available benchmarks but  on arbitrary novel images. They should not require re-training or fine-tuning every time the domain of usage changes.   In this section, we will look at how some of these methods perform on novel unseen images. We will first describe in Section~\ref{sec:evaluation_protocol} the evaluation protocol,  the images that will be used, and the evaluation metrics. We then discuss the performance of these methods in Sections~\ref{sec:performance_stereo_time} and~\ref{sec:performance_stereo_error}.

\subsection{Evaluation protocol}
\label{sec:evaluation_protocol}

\begin{figure}[t]
\centering
\begin{tabular}{@{}c@{}c@{}c@{}c@{}}
	\includegraphics[trim=43cm 14cm 43cm 7.2cm, clip=true, width=.125\textwidth]{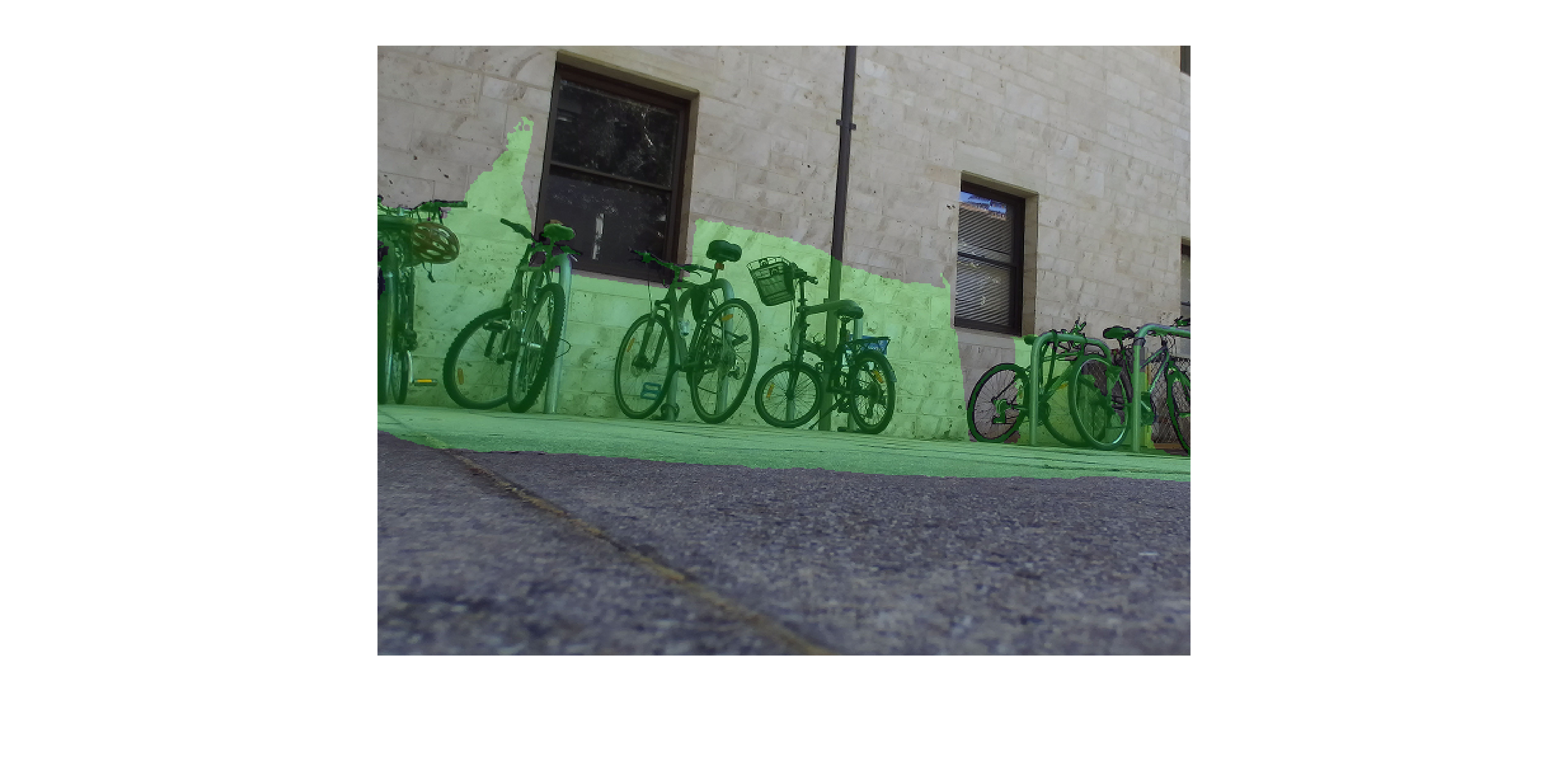}	&
	 \includegraphics[trim=43cm 14cm 43cm 7.2cm, clip=true, width=.125\textwidth]{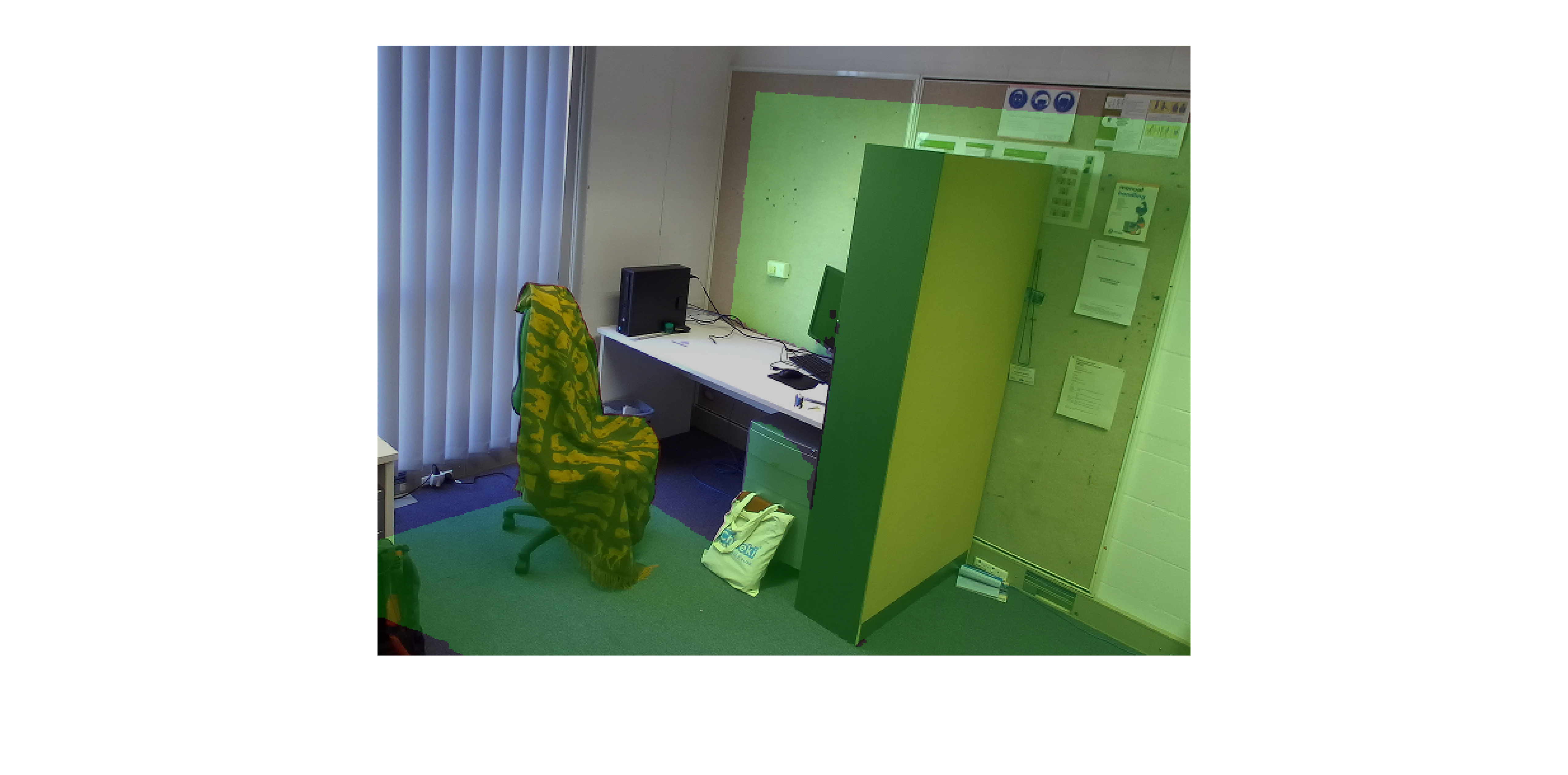} &	
	\includegraphics[trim=43cm 14cm 43cm 7.2cm, clip=true,  width=.125\textwidth]{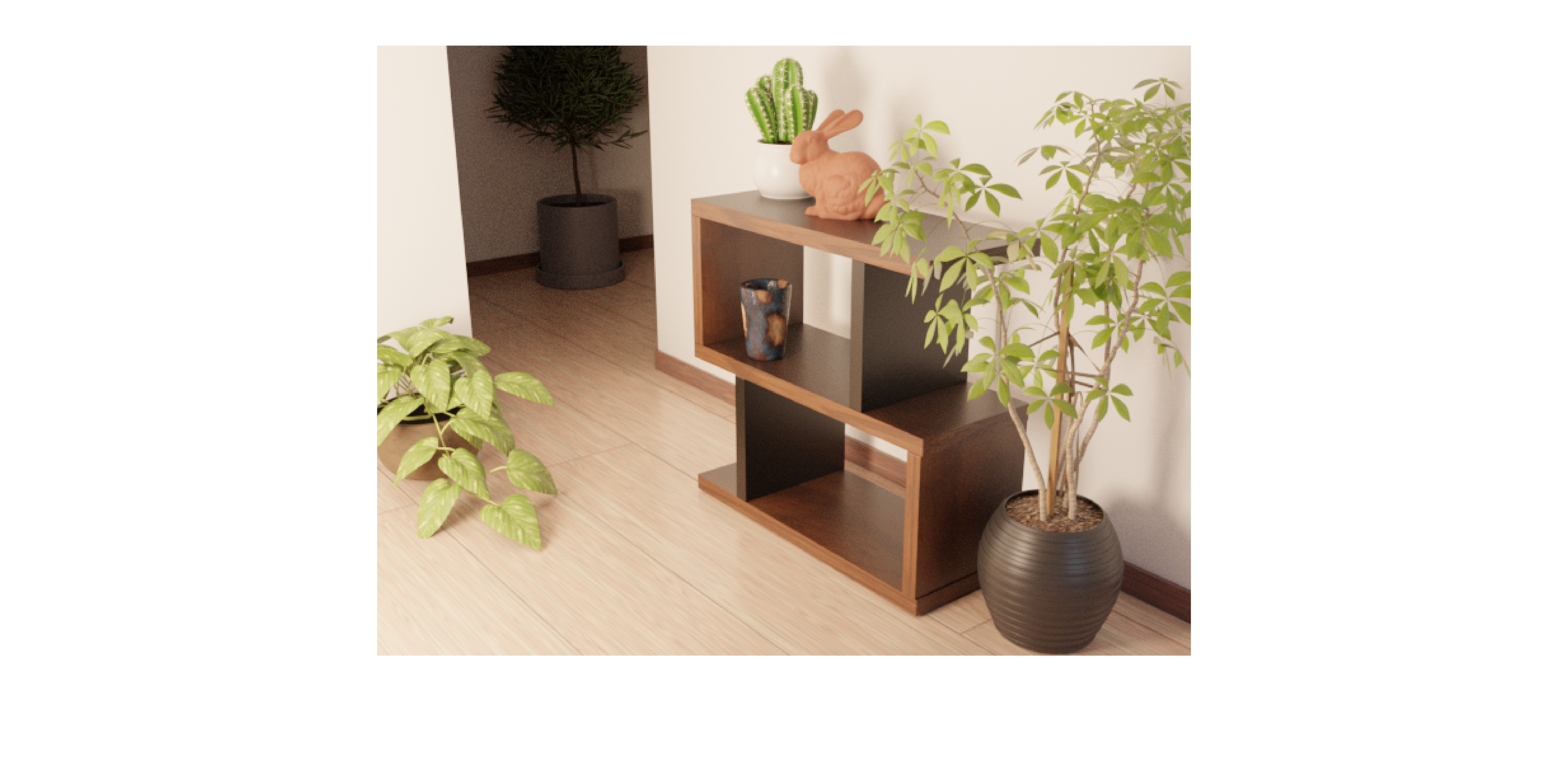} & 
	\includegraphics[trim=43cm 14cm 43cm 7.2cm, clip=true, width=.125\textwidth]{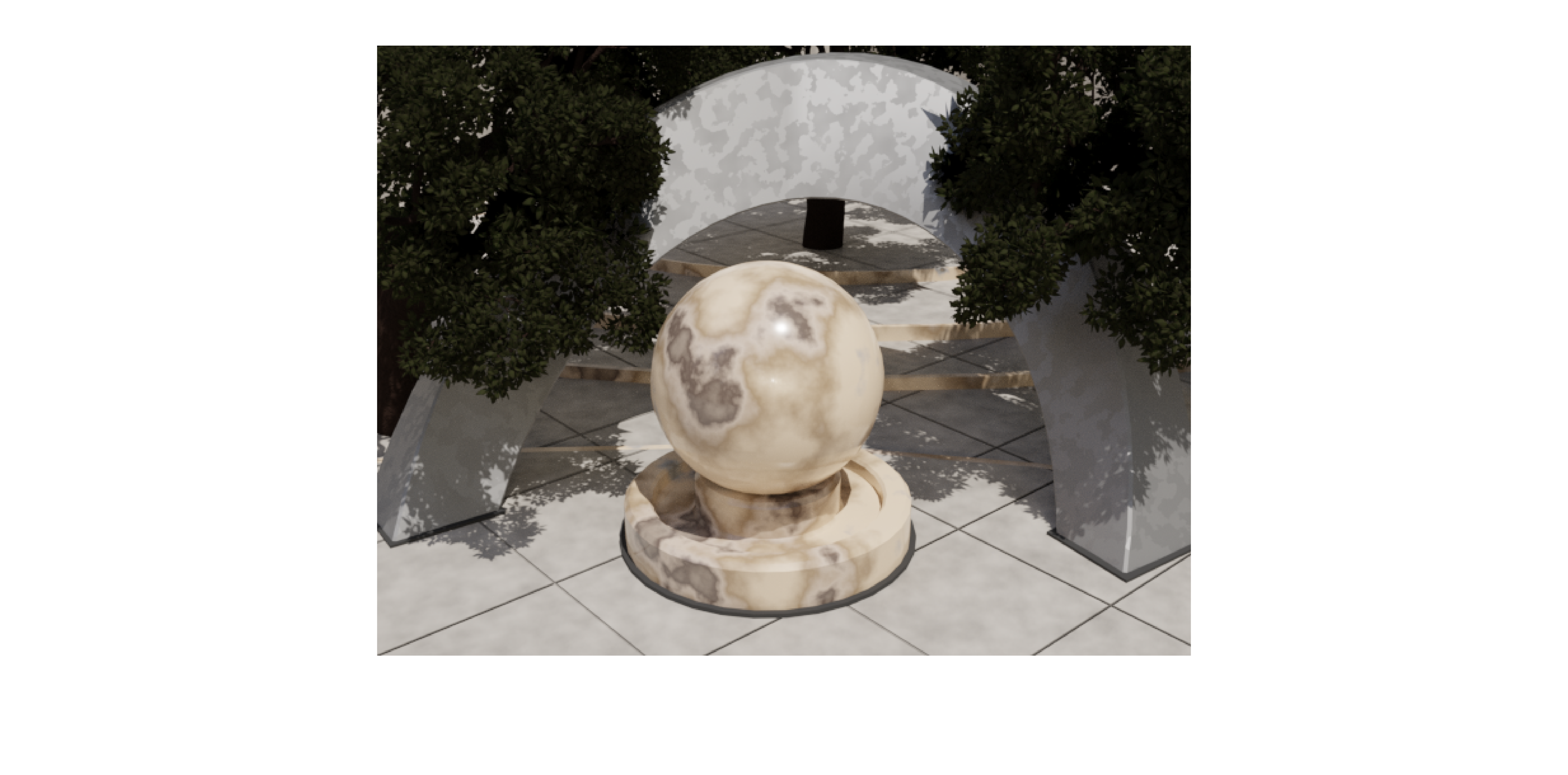}\\
	\multicolumn{4}{c}{ \footnotesize {(a) Left image. }}\\
	
	\includegraphics[trim=3.7cm 3.2cm 3.7cm 1cm, clip=true, width=.125\textwidth]{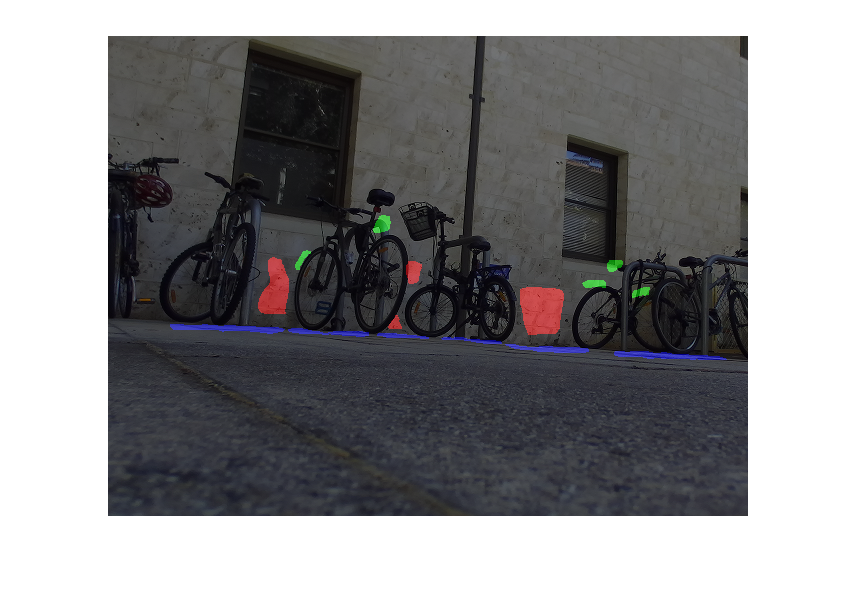}	&
	 \includegraphics[trim=3.7cm 3.2cm 3.7cm 1cm,clip=true, width=.125\textwidth]{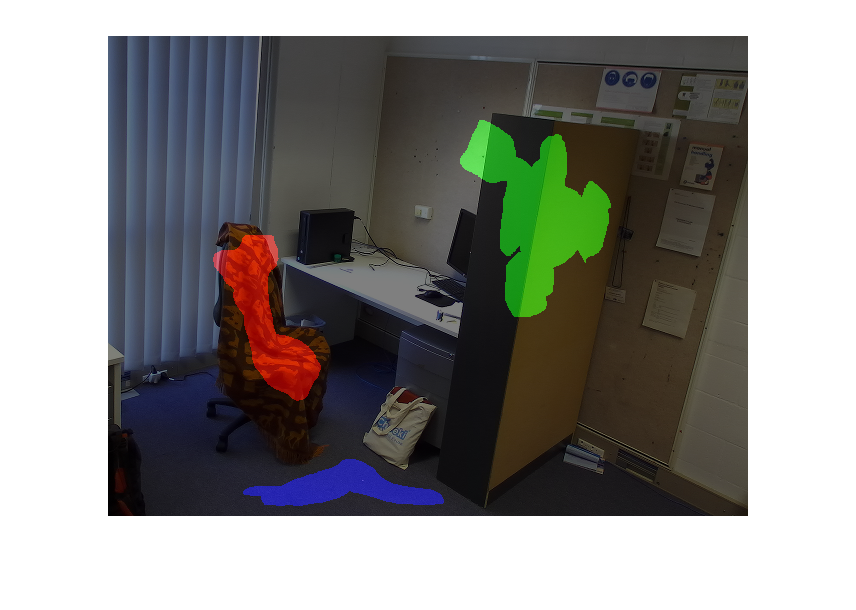} &	
	\includegraphics[trim=3.7cm 3.2cm 3.7cm 1cm,clip=true,  width=.125\textwidth]{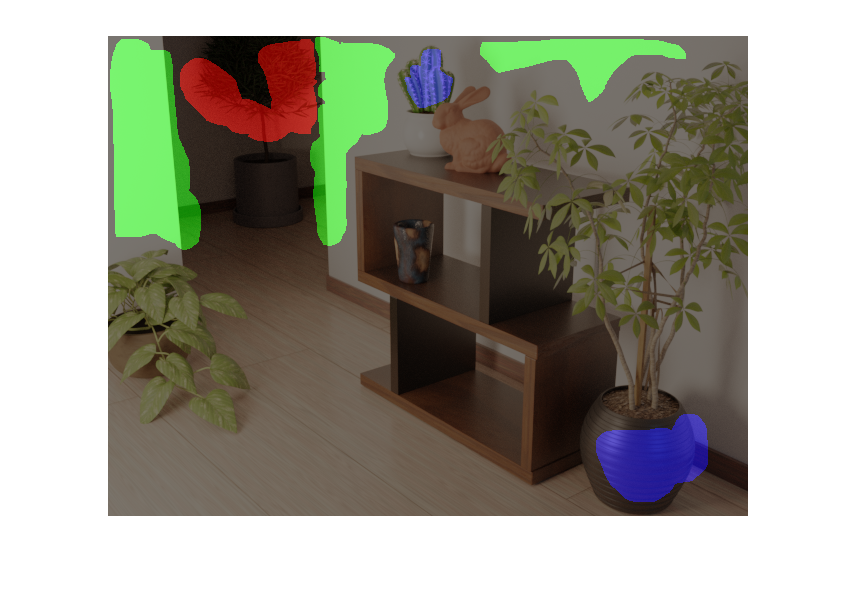} & 
	\includegraphics[trim=3.7cm 3.2cm 3.7cm 1cm, clip=true, width=.125\textwidth]{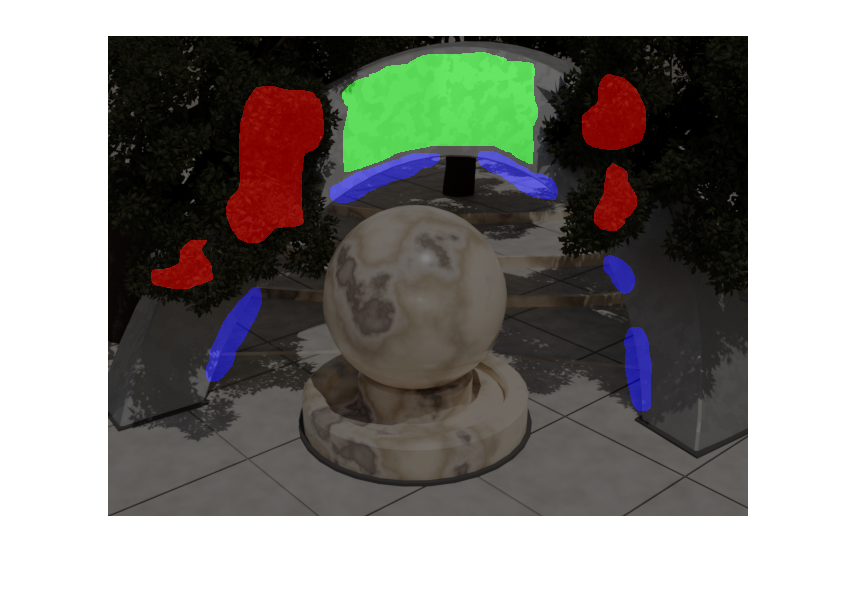}\\
	\multicolumn{4}{c}{ \footnotesize {(b) Highlights of regions of interest where ground-truth disparity is}}\\
	\multicolumn{4}{c}{ \footnotesize {estimated with high confidence. }}\\
	
	\includegraphics[trim=43cm 14cm 43cm 7.2cm, clip=true, width=.125\textwidth]{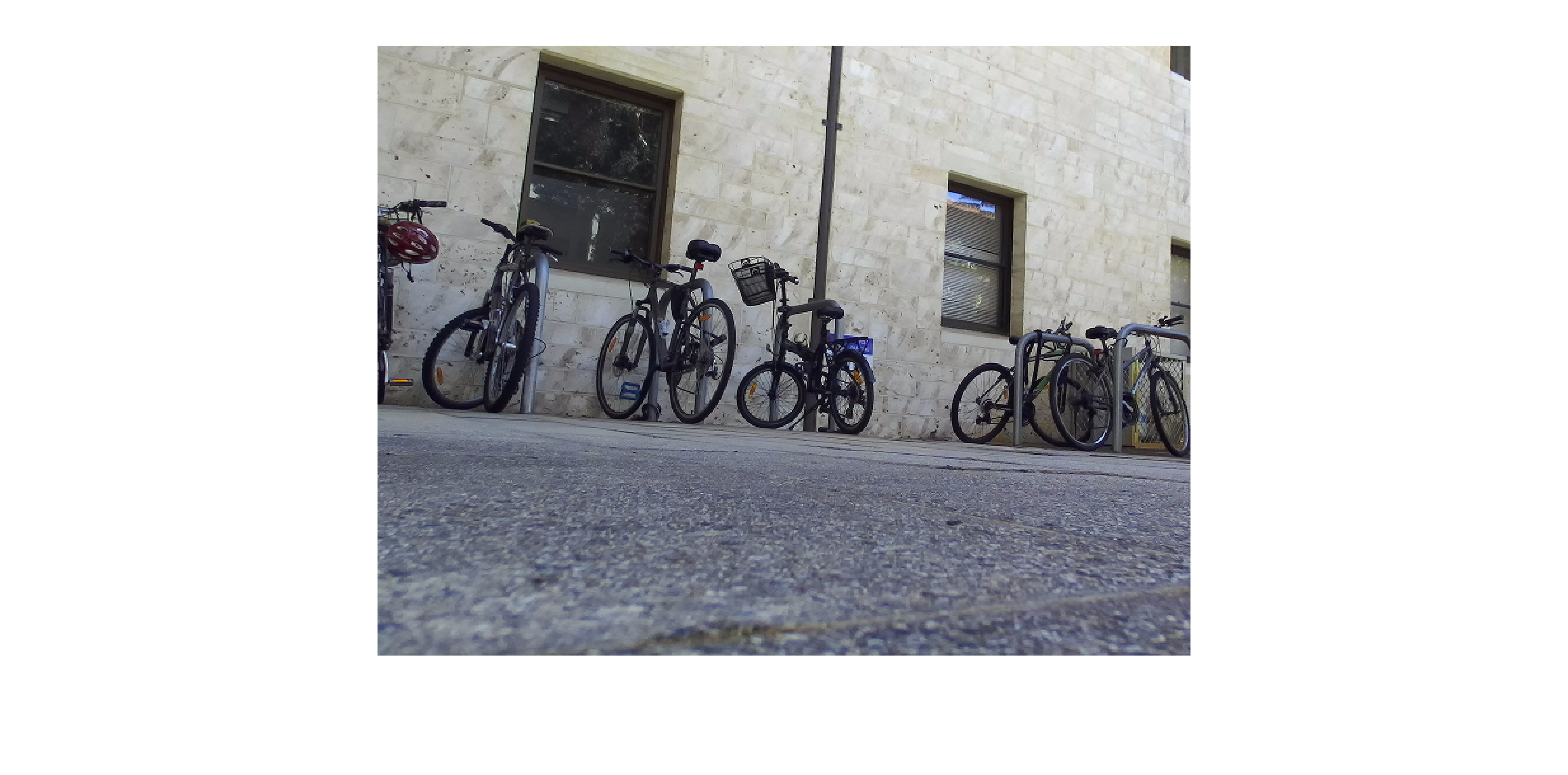} &
	 \includegraphics[trim=43cm 14cm 43cm 7.2cm, clip=true, width=.125\textwidth]{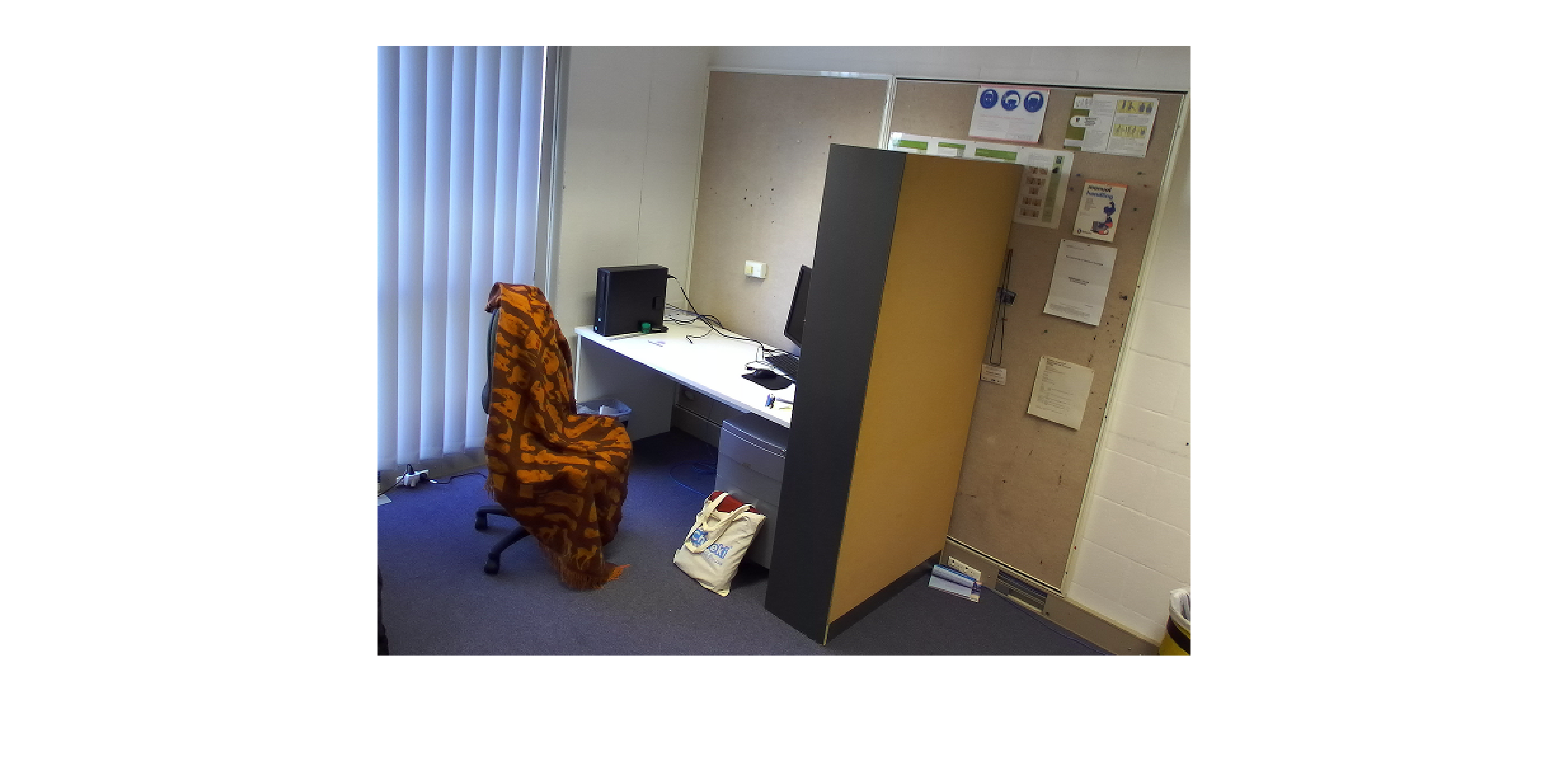} &	
	\includegraphics[trim=43cm 14cm 43cm 7.2cm, clip=true, width=.125\textwidth]{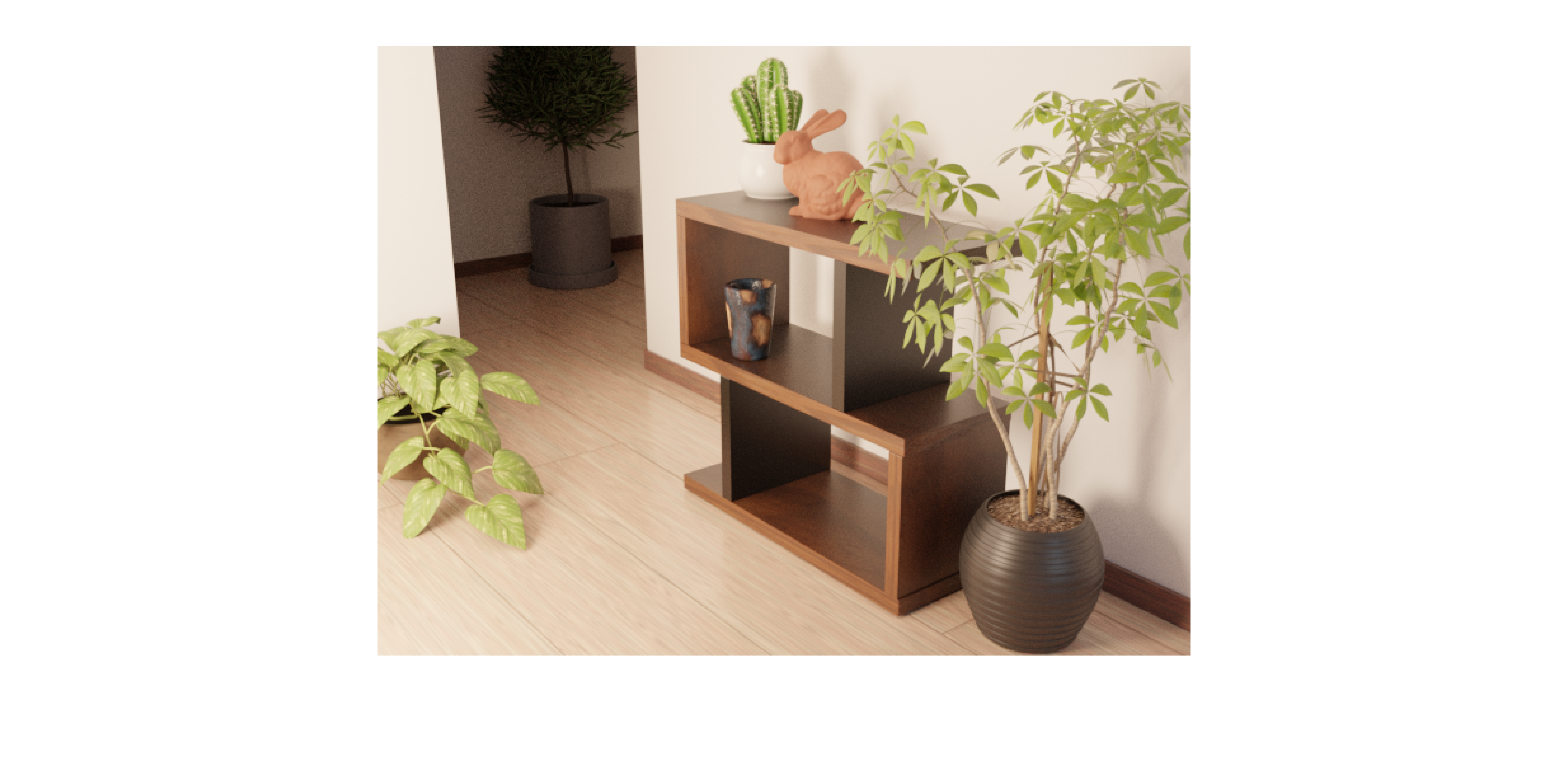} &
	 \includegraphics[trim=43cm 14cm 43cm 7.2cm, clip=true, width=.125\textwidth]{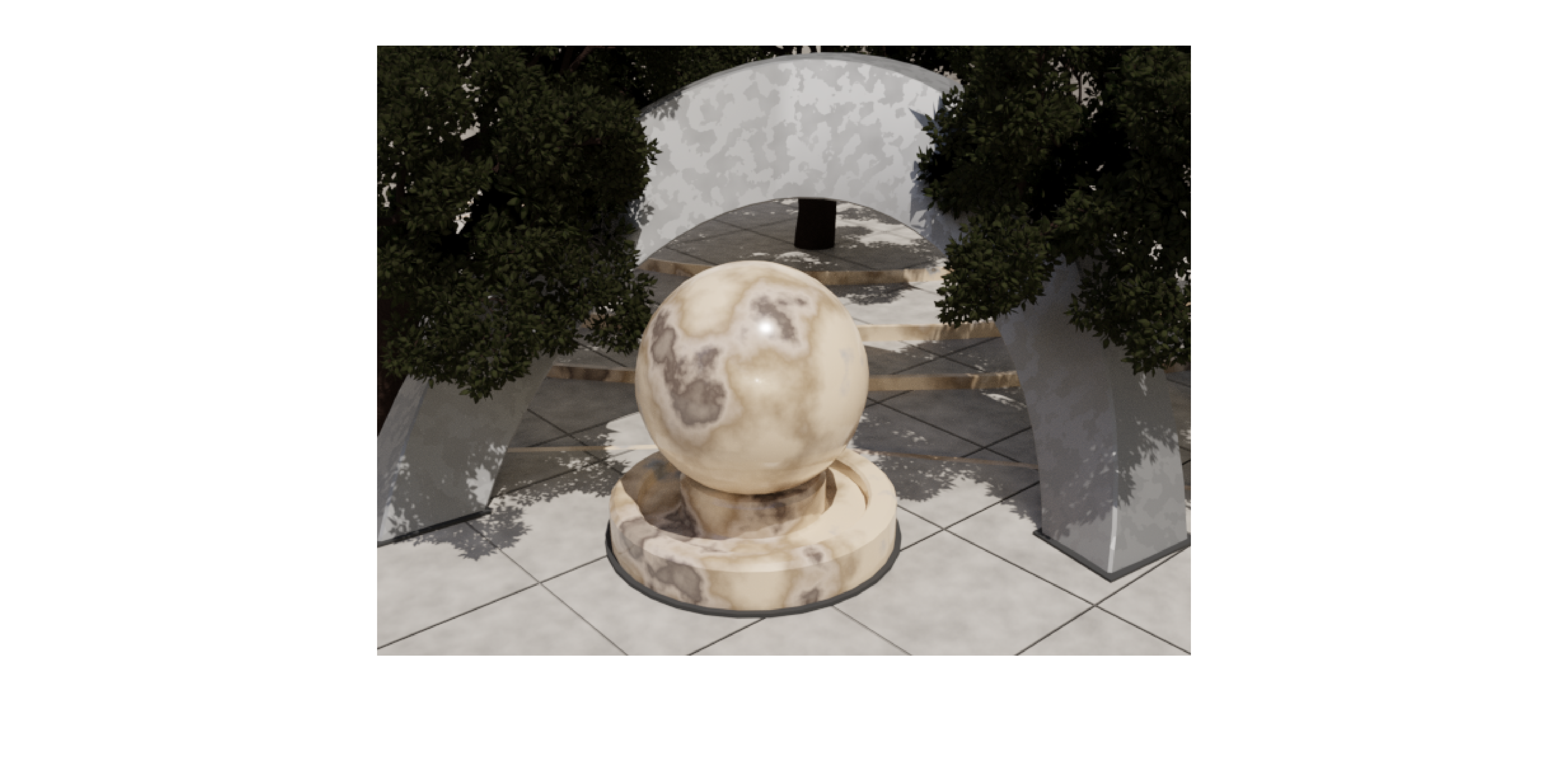}\\
	\multicolumn{4}{c}{\footnotesize{(c) Right image. }}\\
	
	\includegraphics[width=.125\textwidth]{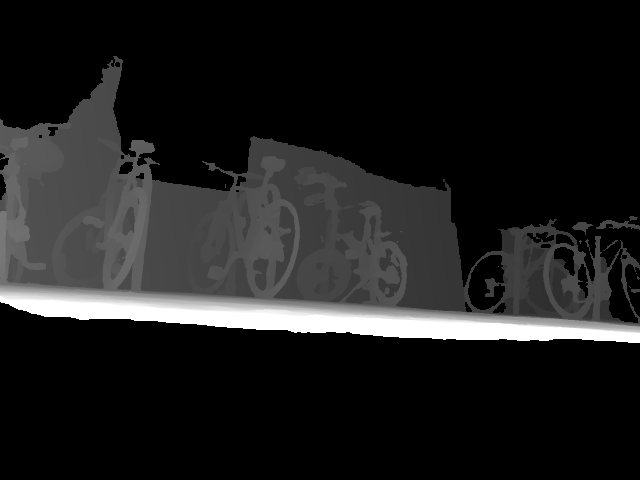} & \includegraphics[width=.125\textwidth]{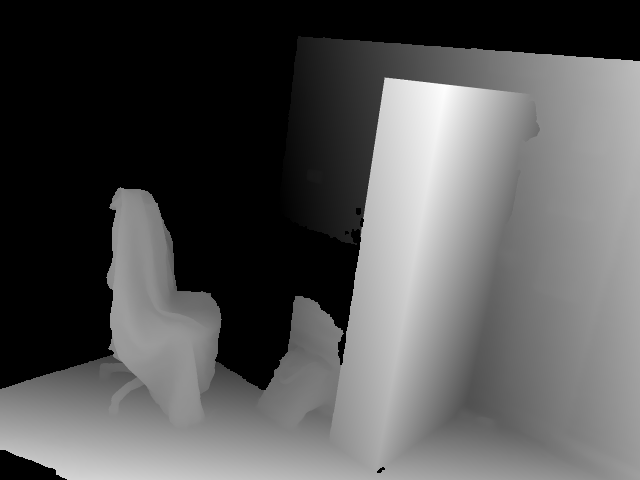} &	
	\includegraphics[width=.125\textwidth]{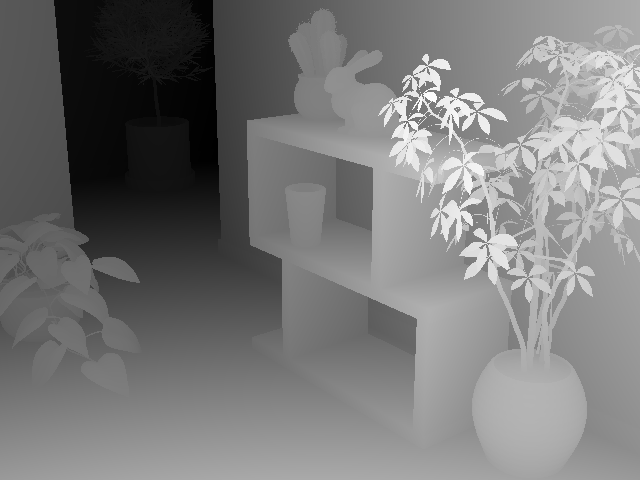} & \includegraphics[width=.125\textwidth]{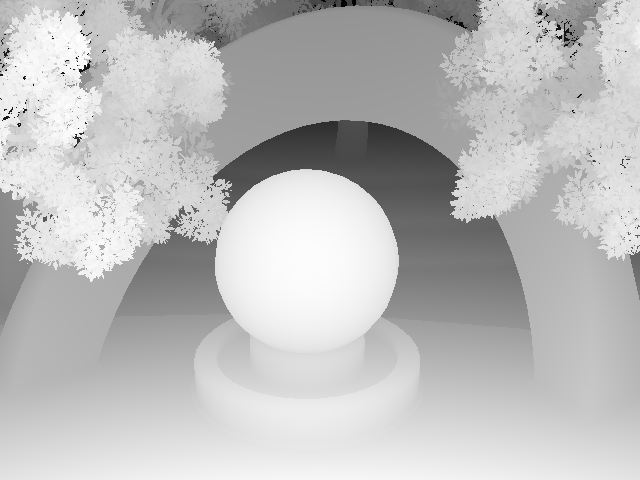}\\
	\scriptsize{Disp. $\in [9.3, 34.0]$}  & \scriptsize{Disp. $\in  [18.7, 29.9] $} &  \scriptsize{Disp. $\in [5.6, 14.5]$}&  \scriptsize{Disp. $\in [5.5, 13.2]$}\\
	\scriptsize{Depth $\in [2.1, 7.8]$}  & \scriptsize{Depth $\in [2.4, 3.3] $} &  \scriptsize{Depth $\in [7.8, 25.0]$}&  \scriptsize{Depth $\in [10.8, 25.6]$}\\
	\multicolumn{4}{c}{ \footnotesize{(d) Ground-truth disparity maps. }}
\end{tabular}
\caption{\label{fig:novel_dataset} Four images, collected in-house and  used to test $16$ state-of-the-art methods. The  green masks on some of the left images highlight the pixels where the ground-truth disparity is available.  The disparity range is shown in pixels while the depth range is in meters.}
\end{figure}

We consider  several key  methods and evaluate their performance on the stereo subset of the ApolloScape dataset~\cite{wang2019apolloscape}, and on an in-house collected set of four images.  The motivation behind this choice is two-fold.  \textbf{First}, the ApolloScape dataset is composed of stereo images taken outdoor in autonomous driving setups. Thus, it exhibits several challenges related to uncontrolled complex and varying lighting conditions, and heavy occlusions.  \textbf{Second}, the dataset is novel and existing methods have not been trained or exposed to this dataset. Thus, it can be used to assess how these methods generalize to novel scenarios. In this dataset, ground truth disparities have been acquired by accumulating 3D point clouds from Lidar and fitting 3D CAD models to individually moving cars. We also use four \textbf{in-house} images of size $\width = 640$ and $\height=480$, see  Fig.~\ref{fig:novel_dataset}, specifically designed to challenge these methods. Two of the images are of real scenes:  a Bicycles scene composed of bicycles in a parking, and an indoor Desk scene composed of office furnitures. We use a moving stereo camera to capture multiple stereo pairs, and Structure-from-Motion  (SfM) to build a 3D model of the scenes. We then render depth maps from the real cameras' viewpoints. Regions where depth is estimated with high confidence  will be used as ground-truth.  The remaining two images are synthetic, but real-looking. They include objects with complex structures, \eg thin structures such as plants, large surfaces with either uniform colors or textures and repetitive patterns, presenting several challenges to stereo-based depth estimation algorithms.


\begin{figure}[t]
\centering
\begin{tabular}{@{}c@{}c@{}c@{}}
	\includegraphics[width=.165\textwidth]{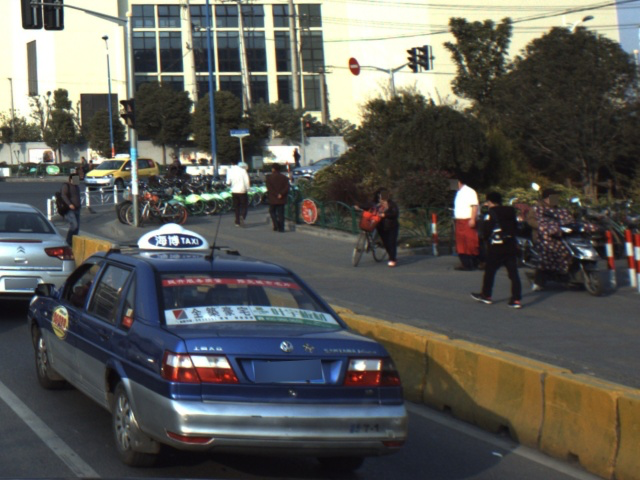} & 
	\includegraphics[width=.165\textwidth]{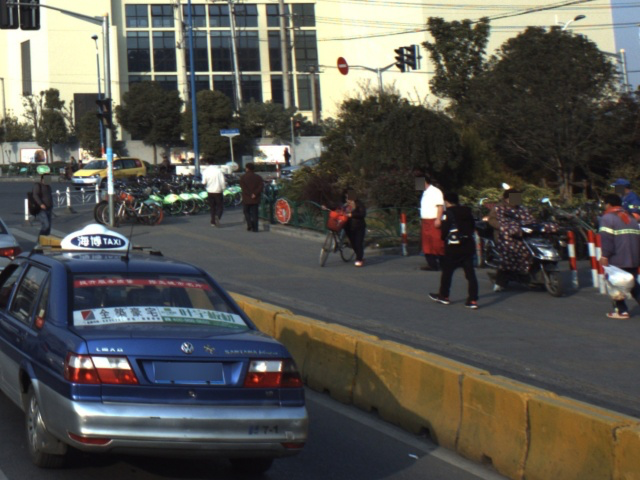} &
	\includegraphics[width=.165\textwidth]{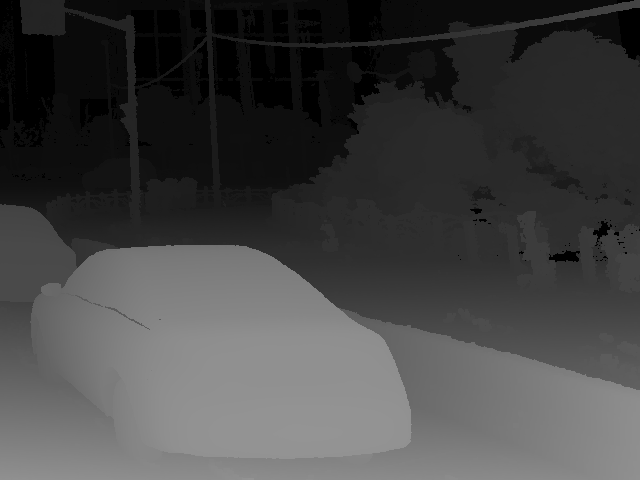} \\
	\multicolumn{3}{c}{\footnotesize{(a) Baseline: images with good lighting conditions. }}\\
	\includegraphics[width=.165\textwidth]{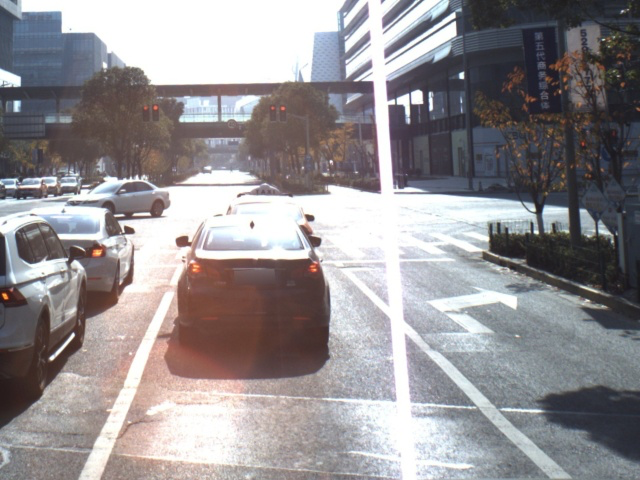} & 
	\includegraphics[width=.165\textwidth]{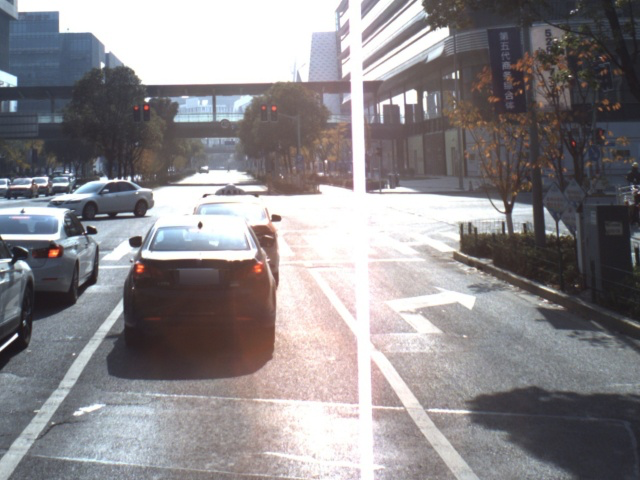} &
	\includegraphics[width=.165\textwidth]{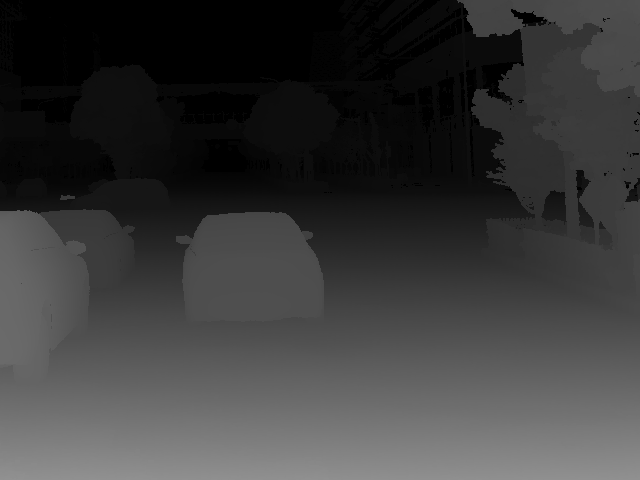} \\
	\multicolumn{3}{c}{\footnotesize{(b) Challenge: images with challenging lighting conditions.}}
\end{tabular}	
\caption{\label{fig:apolloscape_examples}  Examples of stereo pairs and their ground-truth disparity maps from the ApolloScape dataset~\cite{wang2019apolloscape}. }
\end{figure}

We have tested $16$ stereo-based methods published in $9$ papers (between $2018$ and $2019$), see below. We use the network weights  as provided by the authors. 

\vspace{6pt}
\noi\textit{(1) AnyNet~\cite{wang2019anytime}:} It is a four-stages network, which builds 3D cost volumes in a coarse-to-fine manner.  The first stage estimates a low resolution disparity map by searching on a small disparity range.  The subsequent stages estimate refined disparity maps using residual learning.

	
\vspace{6pt}
\noi\textit{(2) DeepPruner~\cite{Duggal_2019_ICCV}:}  It combines deep learning with PatchMatch~\cite{barnes2009patchmatch} to speed up inference by adaptively pruning out the potentially large search space for correspondences. Two variants have been proposed: DeepPruner (Best), which downsamples the cost volume by a factor of $4$, and   DeepPruner (Fast), which  downsamples it by a factor of $8$.
	
\vspace{6pt}
\noi\textit{(3) DispNet3~\cite{Ilg_2018_ECCV}}, an improved version of DispNet~\cite{mayer2016large} where occlusions and disparity maps are jointly estimated. 
	
\vspace{6pt}
\noi\textit{(4) GANet~\cite{zhang2019ga}: } It replaces a large number of the 3D convolutional layers in the regularization block with \textbf{(1)} two 3D convolutional layers, \textbf{(2)} a semi-global aggregation layer  (SGA), and \textbf{(3)} a  local guided aggregation layer (LGA).  SGA and LGA layers capture local and  whole-image cost dependencies. They are meant to  improve the accuracy  in challenging regions such as occlusions, large textureless/reflective regions, and thin structures. 
	
\vspace{6pt}
\noi\textit{(5) HighResNet~\cite{Yang_2019_CVPR}:} To  refine both the spatial and the depth resolutions while operating on high resolution images, this method searches for correspondences incrementally using a coarse-to-fine hierarchy.  Its hierarchical design  also allows for anytime on-demand reports of disparity. 
	
\vspace{6pt}
\noi\textit{(6)  PSMNet~\cite{chang2018pyramid}:} It  progressively regularizes a low resolution 4D cost volume, estimated from a pyramid of features. 
	
\vspace{6pt}
\noi\textit{(7) iResNet~\cite{liang2018learning}: }  The initial disparity and the  learned features are used to calculate a feature constancy map, which measures the correctness of the stereo matching. The initial disparity map and the feature constancy map are then fed into a sub-network for disparity refinement.
	
\vspace{6pt}
\noi\textit{(8) UnsupAdpt~\cite{tonioni2017unsupervised}:} It is an unsupervised adaptation approach that enables fine-tuning  without any ground-truth information. It first trains DispNet-Corr1D~\cite{mayer2016large} using  the KITTI 2012 training dataset and then adapts the network to KITTI2015 and Middlebury 2014.

\vspace{6pt}
\noi\textit{(9) SegStereo~\cite{yang2018segstereo}:} It is an unsupervised disparity estimation method, which uses segmentation masks to guide the disparity estimation. Both segmentation and disparity  maps are jointly estimated with an end-to-end network.

 The methods (1) to (7)  are supervised with ground-truth depth maps while the methods (8) and (9) are self-supervised. We compare their accuracy at runtime using  the  overall Root Mean Square Error  (RMSE) defined as:
		\begin{equation}
			\text{RMSE}^2_{\text{linear}} = { \frac{1}{\npixels} \sum_{\npixels} {| \depth_i- \estimateddepth_i |}^2 },
		\end{equation} 	
\noi and the Bad-n error defined as the percentage of pixels whose estimated disparity deviates with more than  $n$ pixels from the ground truth. We use $n \in \{0.5, 1, 2, 3, 4, 5 \}$. The Bad-n error  considers the distribution and spread of the error and thus provides a better insight on the accuracy of the methods. In addition to accuracy, we also report the computation time  and memory footprint at runtime. 



\subsection{Computation time and memory footprint}
\label{sec:performance_stereo_time}

\begin{table*}[t]
\caption{\label{tab:memory_time}  Computation time and memory consumption, at runtime, on images of size $640 \times 480$.  SegStereo~\cite{yang2018segstereo} has been tested on a PC equipped with  an Nvidia  GeForce RTX 2080. The other methods have been tested on a PC equipped with  an Nvidia Tesla K40 GPU with a 12 Go graphic memory.  See the Supplementary Material for a visual representation. }

\resizebox{\linewidth}{!}
	{

\begin{tabular}{@{}l c @{ }ccc@{}c@{ }c ccc@{}ccc@{}}  
	\toprule
	\multirow{2}{*}{\textbf{Method}} &  {\textbf{Supervision}} &\multirow{2}{*}{\textbf{Cost vol.}} & \multirow{2}{*}{\textbf{Time (s)}} &  \multirow{2}{*}{\textbf{Memory (GB)}}  & \multirow{2}{*}{\textbf{Training set}} & \multicolumn{3}{c}{\textbf{Baseline}} &   & \multicolumn{3}{c}{\textbf{Challenge}}  \\ 
		\cline{7-9} \cline{11-13}
		& \textbf{mode} & & & & & Bkg & Fg &  Bkg$+$Fg & &  Bkg & Fg &  Bkg$+$Fg\\
	\midrule
	AnyNet~\cite{wang2019anytime} &   Supervised & 3D &  $0.285$ &	$\textbf{0.232}$  & KITTI2015 &   9.46	& 10.74	&10.34	 && 9.83	& 11.60 & 11.15\\
	\cline{6-13}
							   &     & &   &	   & KITTI2012 &  9.80	&10.29	&10.20	& &9.34	&10.62	&10.61\\
	\cline{1-13}
	DeepPruner (Best)~\cite{Duggal_2019_ICCV}  & Supervised & 3D &  $8.430$ & $8.845$ &  KITTI2012$+$2015 & 9.64&	9.43	& 9.46& & 	12.38	& $\textbf{8.74}$ &10.48 \\
	\cline{1-13}
	DeepPruner (Fast)~\cite{Duggal_2019_ICCV}   &Supervised &  3D &  $3.930$ & $6.166$ &  KITTI2012$+$2015  &9.56&	9.90	&9.94	& &8.74&	9.75	&9.86 \\
	\cline{1-13}
	
	DispNet3~\cite{Ilg_2018_ECCV}  &Supervised & 3D  & $-$ & $10.953$  & CSS-ft-KITTI & 9.68	&9.62	&9.70& &	$\textbf{8.38}$	&11.00	&11.11\\
	\cline{6-13}
							  &  &    &   &	   & CSS-FlyingThings3D~\cite{mayer2016large} & 9.11	&9.64&	9.54	& & 8.97	&9.91	&10.19 \\
							  \cline{6-13}
							  &   &   &   &	   & css-FlyingThings3D~\cite{mayer2016large} &  9.29	&9.98&	9.87	& & 9.66	&10.34	&10.61 \\
	\cline{1-13}
	GANet~\cite{zhang2019ga}  &  Supervised & 4D &  $8.336$ &	3.017  & KITTI2015 & 9.55	& $\textbf{9.38}$	& $\textbf{9.39}$ &	&9.37&	9.50	&9.89  \\ 
	\cline{6-13}
							 &    & &   &	   & KITTI2012 &  9.98	&10.29&	10.25 & &10.69&	10.95&	11.55\\
	\cline{1-13}
	HighResNet~\cite{Yang_2019_CVPR} &  Supervised & 4D &  $\textbf{0.037}$ &	$0.474$  & Middleburry~\cite{scharstein2014high}, KITTI2015~\cite{menze2015object}, & 9.47&	9.91&	9.94	 & &8.58&	9.64	& $\textbf{9.78}$ \\
								  &  & & &  &  ETH3D~\cite{schops2017multi}, HR-VS~\cite{Yang_2019_CVPR}& \\
	\cline{1-13}
	PSMNet~\cite{chang2018pyramid}    & Supervised &  4D &  $1.314$	& $1.900$  & KITTI2015& 9.88	&9.81&	9.80	& &10.10&	9.42	&9.93\\ 
	\cline{6-13}
								&     & &   &	   & KITTI2012 &  10.17	&10.24	&10.29&	& 10.66&	10.33	&11.00\\
	\cline{1-13}
	iResNet~\cite{liang2018learning}  & Supervised &  3D &  $0.939$ & $7.656$ & KITTI2015 & 60.04	&61.72	&60.54& &45.87	&46.85 & 47.86\\
	\cline{6-13}
							&   &   &   &	   & ROB~\cite{rob} &  22.08	&17.16	&18.08& &	23.01	&16.51	&18.83\\
	\cline{1-13}
	UnsupAdpt~\cite{tonioni2017unsupervised}   &   Self-supervised & 3D &  $-$ & $-$ & KITTI2012 adapted to KITTI2015 & 9.44	&10.39&	10.19 & &10.10&	10.42&	10.78\\
	\cline{6-13}
									     &    &  &   &	   & Shadow-on-Truck & $\textbf{8.52}$&	10.08	&9.58 & &10.66&	10.88&	10.27\\
	\cline{1-13}
	SegStereo~\cite{yang2018segstereo}  & Self-supervised &  3D   & $0.195$ & $\sim12.00$  & CityScapes~\cite{cordts2016cityscapes} & 9.26&	10.30	&10.17& &9.03	&10.49&	10.54\\
	\bottomrule
	

\end{tabular}
}		
\end{table*}

From Table~\ref{tab:memory_time}, we can distinguish three types of methods; slow methods, \eg  PSMNet~\cite{chang2018pyramid}, DeepPruner (Best) and (Fast)~\cite{Duggal_2019_ICCV},   and GANet~\cite{zhang2019ga}, require  more than $1$ second to estimate one  disparity map. They  also require between $3$GB and $10$GB (for DispNet3~\cite{Ilg_2018_ECCV}) of memory at runtime. As such, these methods are very hard to deploy on mobile platforms. Average-speed methods, \eg AnyNet~\cite{wang2019anytime} and iResNet~\cite{liang2018learning}, produce a disparity map  in around one second. Finally, fast methods, \eg HighResNet~\cite{Yang_2019_CVPR},  require less than $0.1$ seconds. In general, methods that use 3D cost volumes are faster and less memory demanding than those that use 4D cost volumes. There are, however, two exceptions: iResNet~\cite{liang2018learning} and DeepPruner~\cite{Duggal_2019_ICCV}, which use 3D cost volumes but  require a large amount of memory at runtime.  While iResNet requires less than a second to process  images of size $\width=640, \height = 480$, since it uses 2D convolutions to regularize the cost volume, DeepPruner~\cite{Duggal_2019_ICCV} requires more than $3$ seconds.  We also observe that HighResNet~\cite{Yang_2019_CVPR}, which uses 4D cost volumes but adopts a hierarchical approach to produce disparity on demand, is very efficient in terms of computation time as it only requires $37$ms, which is almost $8$ times faster than AnyNet~\cite{wang2019anytime}, which uses 3D cost volumes.  Note also that AnyNet~\cite{wang2019anytime} can run on mobile devices due to its memory efficiency.


\subsection{Reconstruction accuracy}
\label{sec:performance_stereo_error}

\begin{figure}[t]
\centering
	\includegraphics[trim=1.9cm 9.4cm  11cm 2cm, clip=true, width=.45\textwidth]{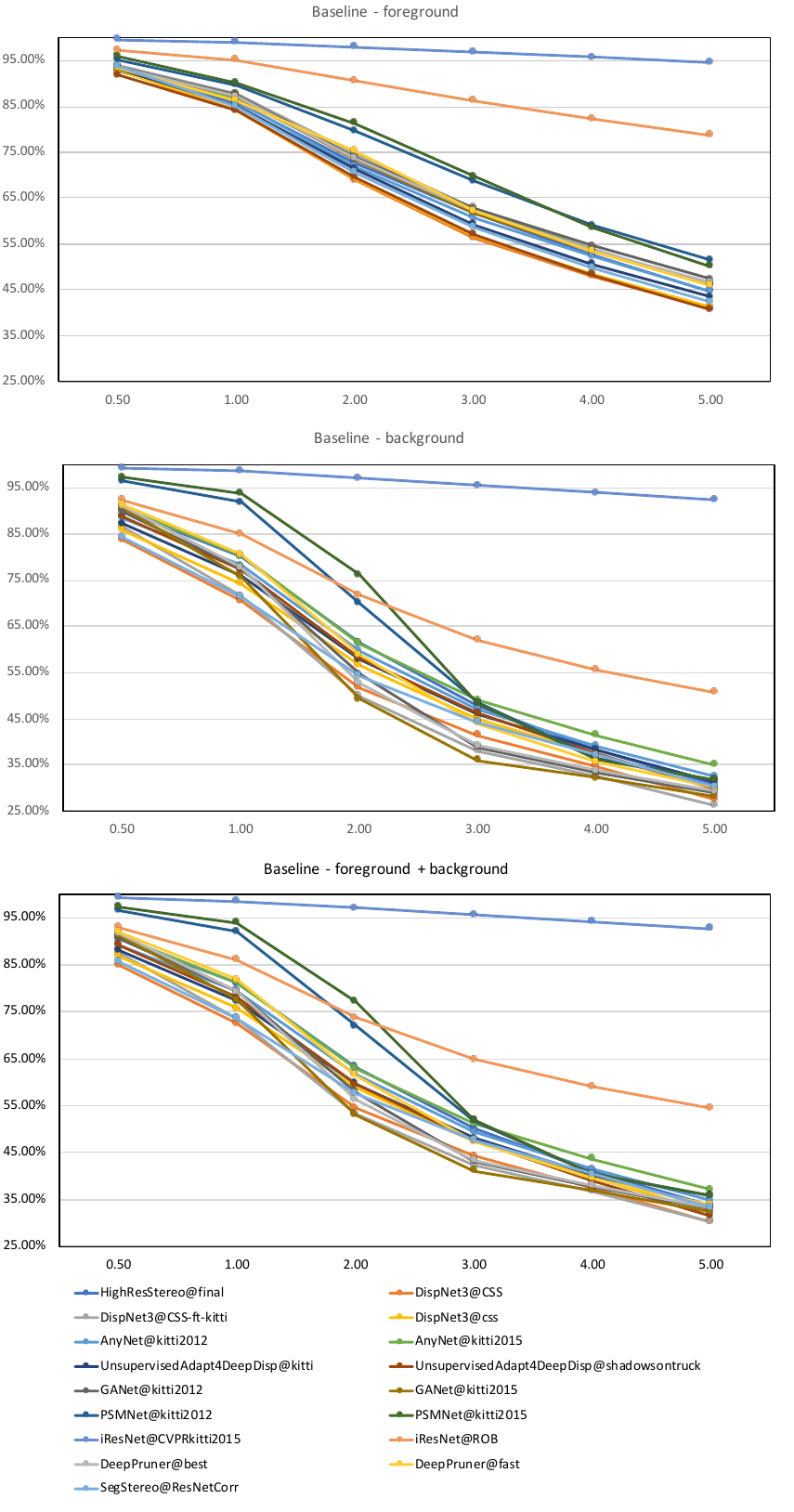} 
\caption{\label{fig:badn_apolloscape} Overall Bad-n error, $n \in [0.5,   5.0 ]$ on a selection of $141$ (baseline)   images from the stereo vision challenge of ApolloScape dataset~\cite{wang2019apolloscape}.   A similar behaviour is observed on the challenge subset, see the supplementary material. The horizontal axis is the error $n$ while the vertical axis is the percentage of pixels whose estimated disparity deviates with more than  $n$ pixels from the ground truth.  }
\end{figure}

Table~\ref{tab:memory_time}  shows the   average RMSE of each of the methods described in Section~\ref{sec:evaluation_protocol}.  We report the results on a baseline subset composed of $141$ images that look more or less like KITTI2012 images, hereinafter referred to as \emph{baseline}, and on another subset composed of $33$ images with challenging lighting conditions, hereinafter referred to as \emph{challenge}.  Here, we focus on the relative comparison across methods since some of the high errors observed  might be attributed to the way the ground-truth has been acquired in ApolloScape~\cite{wang2019apolloscape} dataset, rather than to the methods themselves.

We observe that  these methods behave almost equally on the two subsets. However,  the reconstruction error,  is significantly important, $>8$ pixels, compared to the errors reported on standard datasets such as KITTI2012 and KITTI2015.  This suggests that, when there is a significant domain gap between training and testing then the reconstruction accuracy can be significantly affected. 

We also observe the same trend on the Bad-n curves of Fig.~\ref{fig:badn_apolloscape} where, in all methods, more than $25\%$ of the pixels had a reconstruction error that is larger than $5$ pixels.  
The  Bad-n   curves  show that the errors are  large on the foreground pixels, \ie pixels that correspond to cars, with more than $55\%$ of the pixels having an error that is larger than $3$ pixels (against $35\%$ on the background pixels).  Interestingly, Table~\ref{tab:memory_time} and Fig.~\ref{fig:badn_apolloscape} show that most of the methods achieve similar reconstruction accuracies. The only exception is iResNet~\cite{liang2018learning} trained on Kitti2015 and on ROB~\cite{rob}, which had more than $90\%$, respectively $55\%$, of pixels with an error that is larger than $5$ pixels.  In all methods, less than $5\%$ of the pixels had an error that is less than $2$ pixels. This suggests that achieving sub-pixel accuracy remains an important challenge for future research.

Note that SegStereo~\cite{yang2018segstereo}, which is self-supervised, achieves a similar or better performance than many of the supervised methods. Also, the unsupervised self-adaptation method of Tonioni \etal~\cite{tonioni2017unsupervised}, which takes the baseline DispNet-Corr1D network~\cite{mayer2016large} trained on KITTI 2012 and adapts it to KITTI2015 and Middlebury 2014, achieves one of the best performances on the foreground regions.

\begin{figure*}[t]
\centering
\resizebox{.9\linewidth}{!}{%
	\begin{tabular}{@{}c@{}c@{}c@{}@{}c@{}c@{}c@{}}
	
		\includegraphics[trim=2.2cm 2.2cm 1.5cm 1cm, clip=true, width=.16\textwidth]{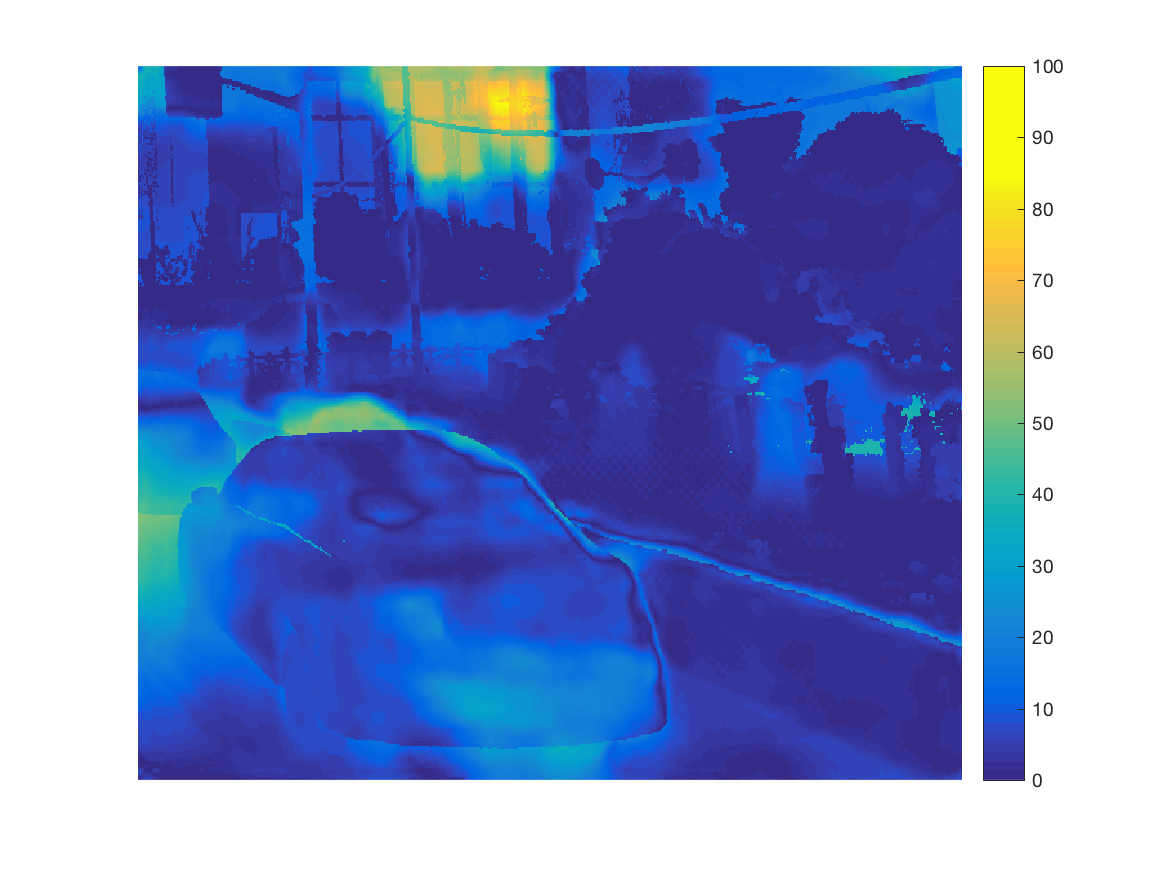} & 
		\includegraphics[trim=2.2cm 2.2cm 1.5cm 1cm, clip=true, width=.16\textwidth]{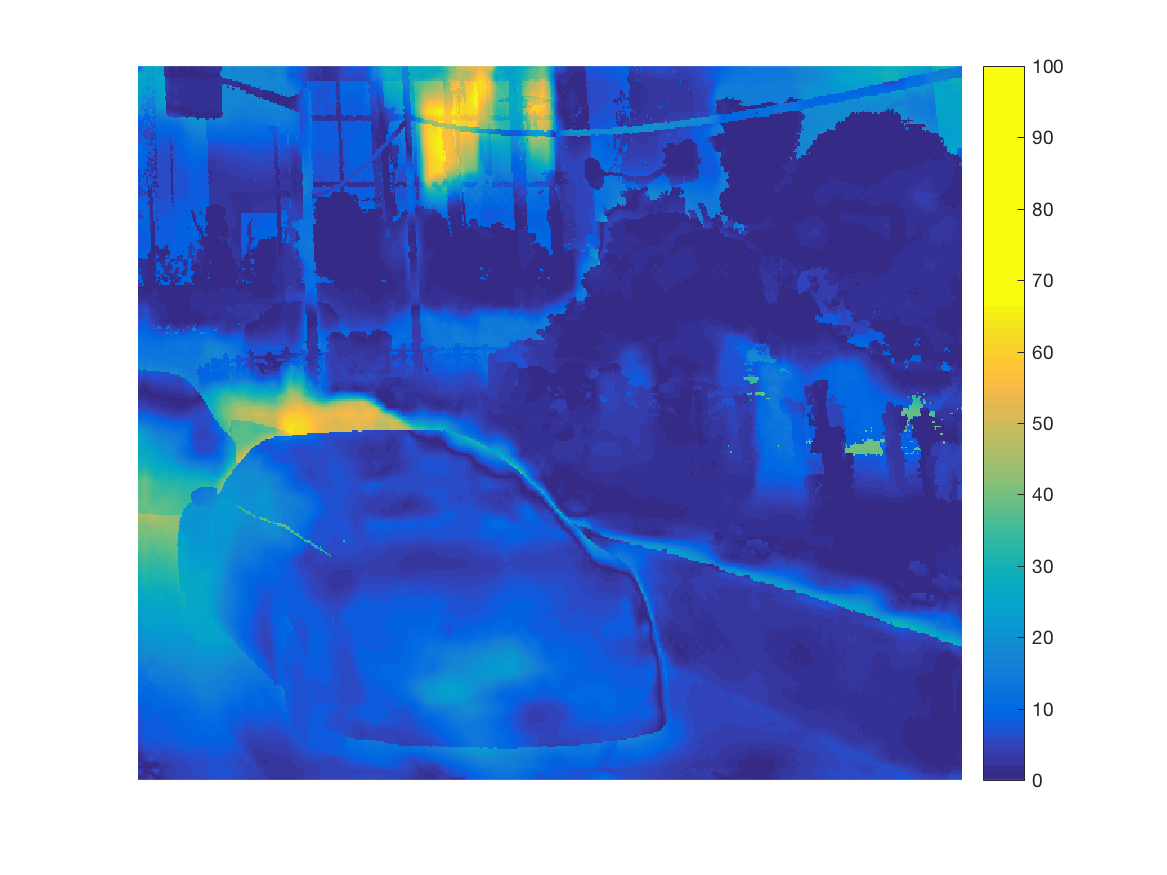} & 
		\includegraphics[trim=2.2cm 2.2cm 1.5cm 1cm, clip=true, width=.16\textwidth]{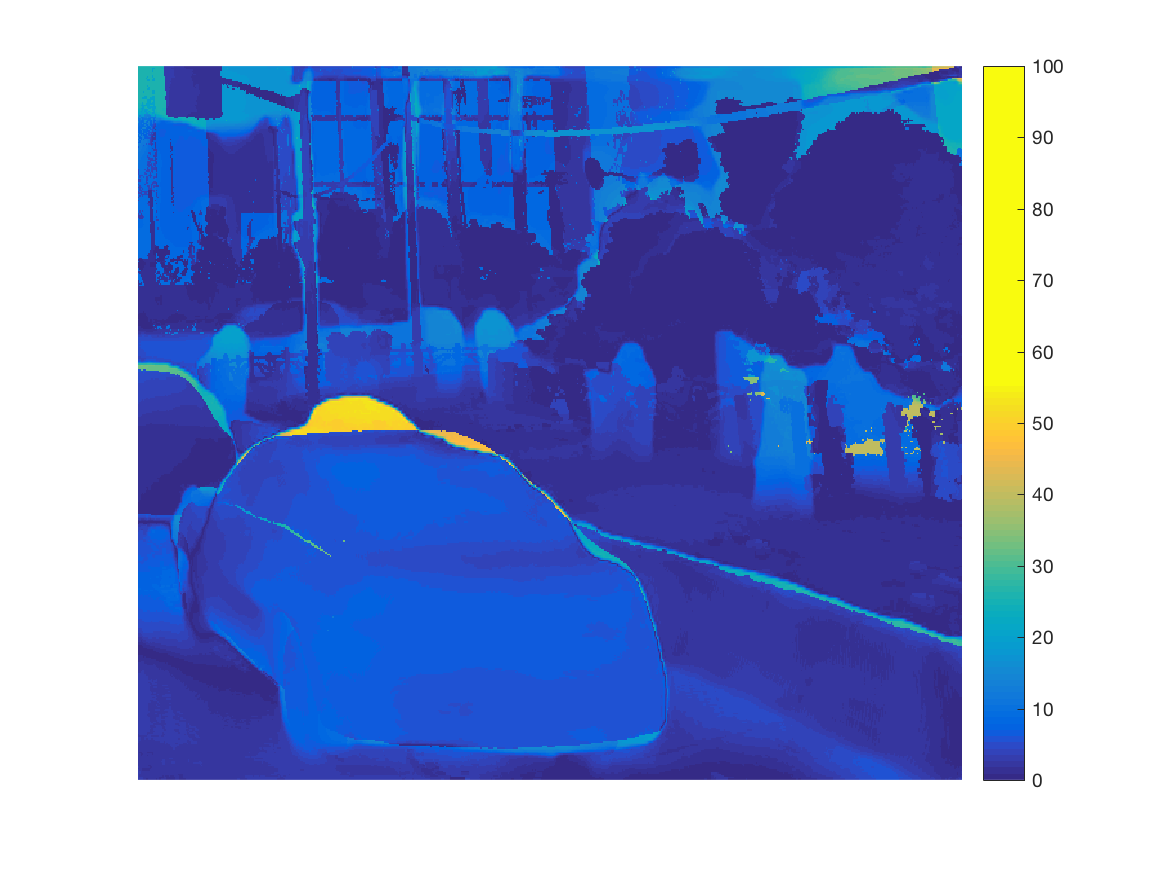} & 
		\includegraphics[trim=2.2cm 2.2cm 1.5cm 1cm, clip=true, width=.16\textwidth]{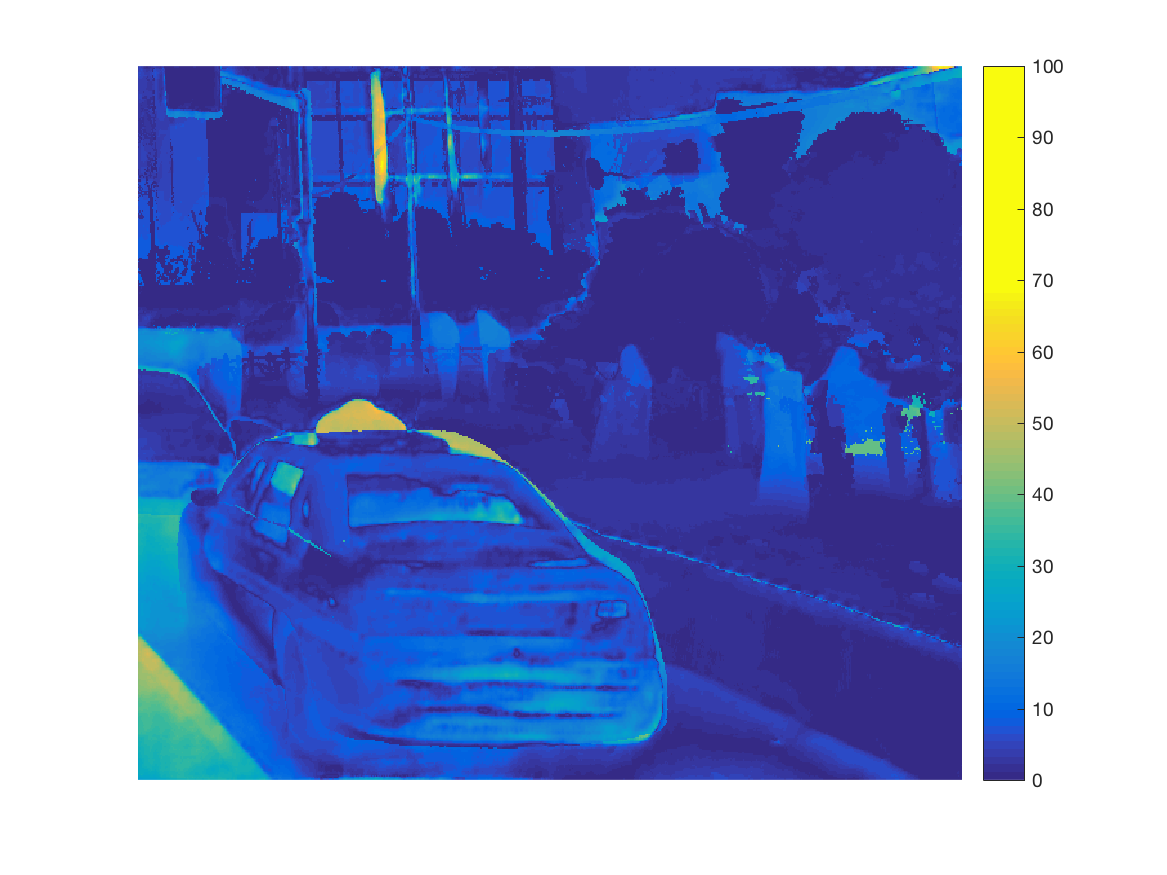} & 		
		\includegraphics[trim=2.2cm 2.2cm 1.5cm 1cm, clip=true, width=.16\textwidth]{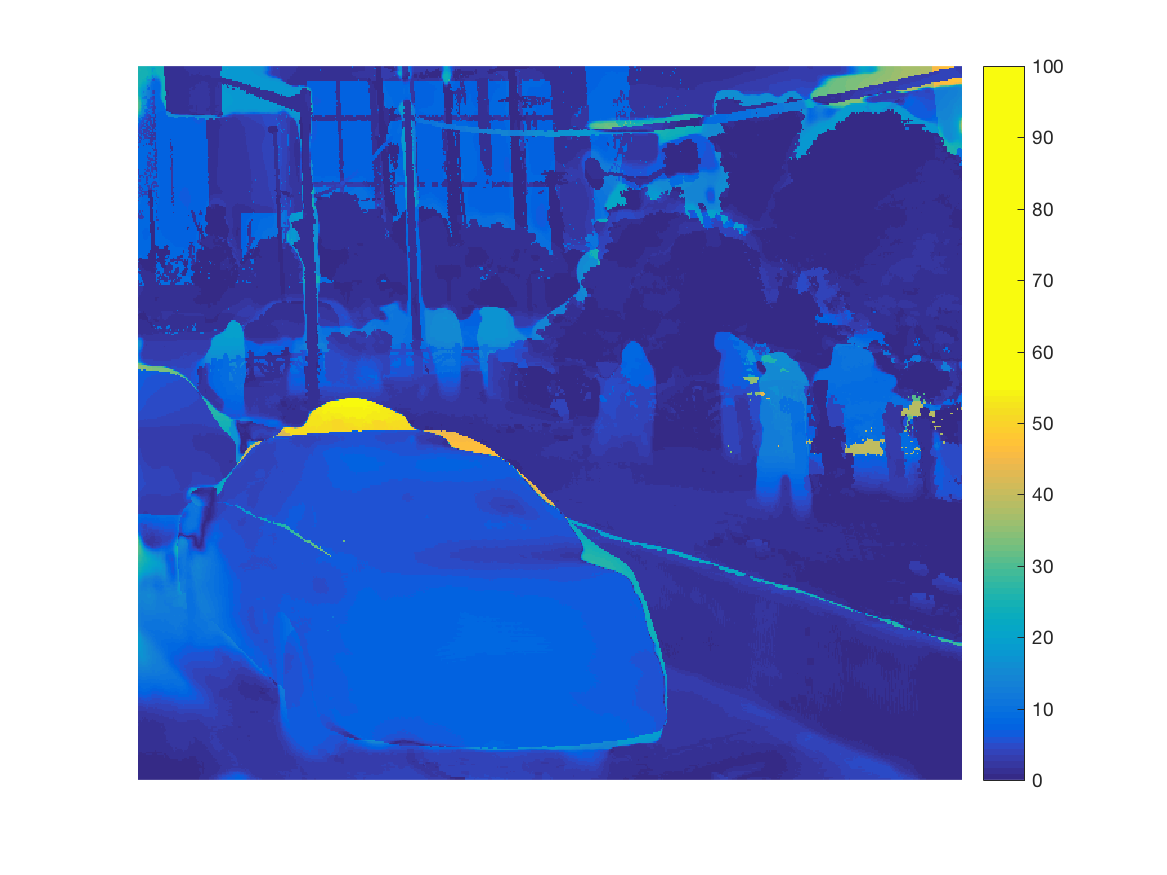} &
		\includegraphics[trim=2.2cm 2.2cm 1.5cm 1cm, clip=true, width=.16\textwidth]{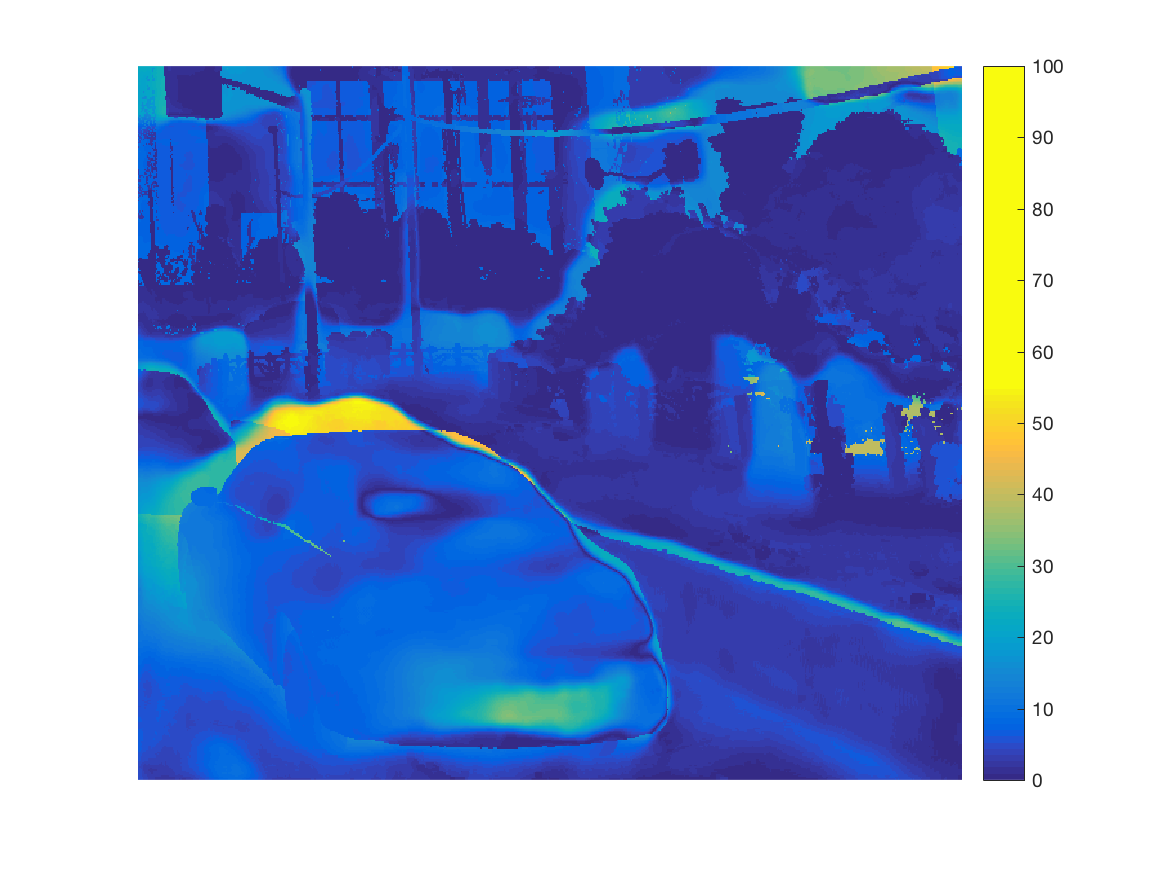} \\

		\footnotesize{AnyNet (Kitti2012)} & \footnotesize{AnyNet (Kitti2015) }& \footnotesize{DeepPruner (fast) }& \footnotesize{DispNet3 (css) }& \footnotesize{GANet (Kittit2015) }& \footnotesize{Unsup.Adapt (Kitti) }\\ 
		
		\includegraphics[trim=2.2cm 2.2cm 1.5cm 1cm, clip=true, width=.16\textwidth]{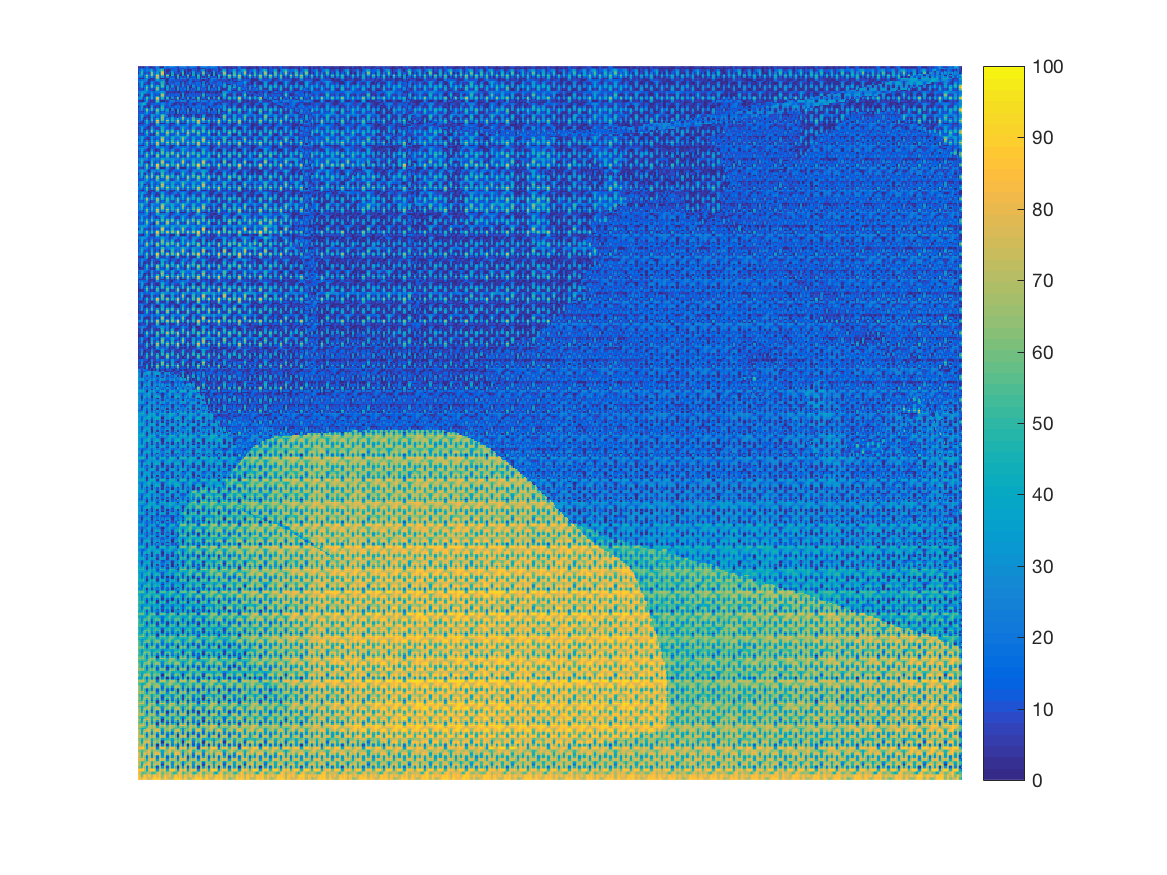} & 
		\includegraphics[trim=2.2cm 2.2cm 1.5cm 1cm, clip=true, width=.16\textwidth]{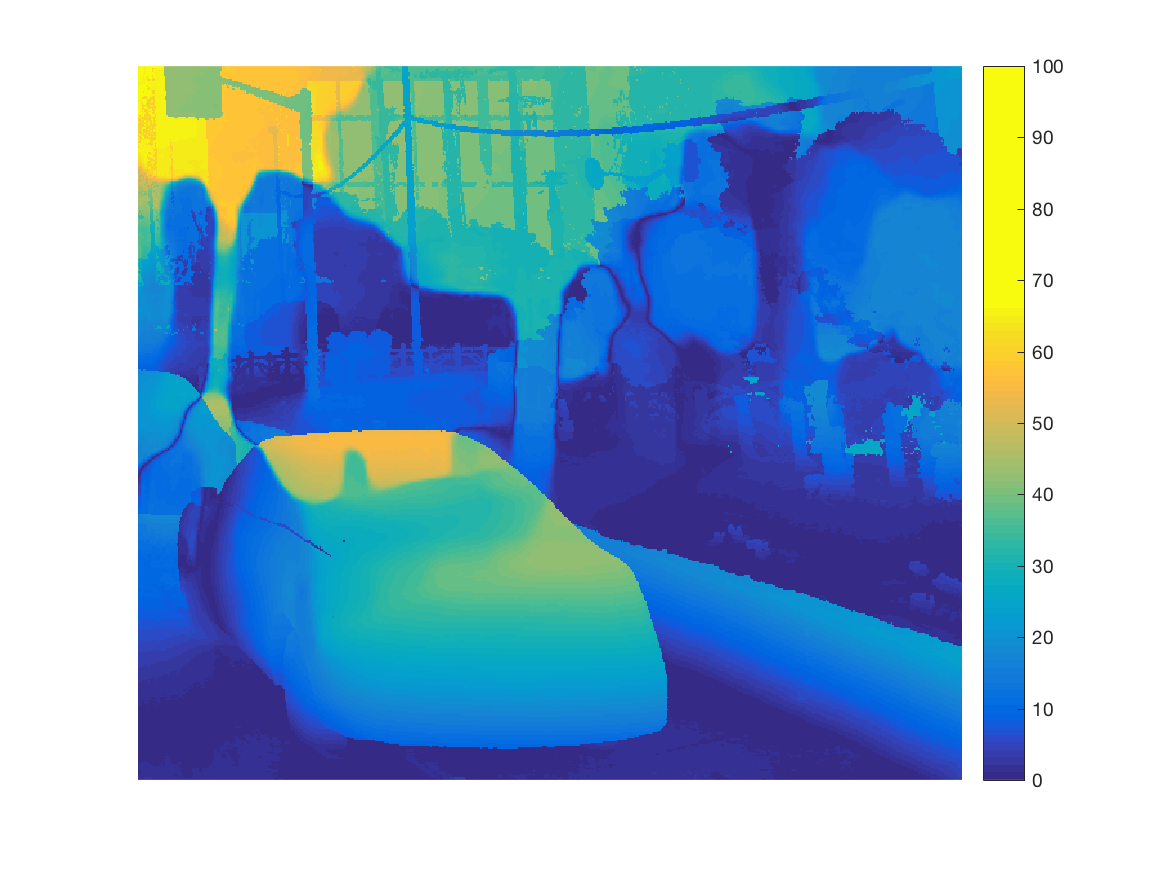} & 
		\includegraphics[trim=2.2cm 2.2cm 1.5cm 1cm, clip=true, width=.16\textwidth]{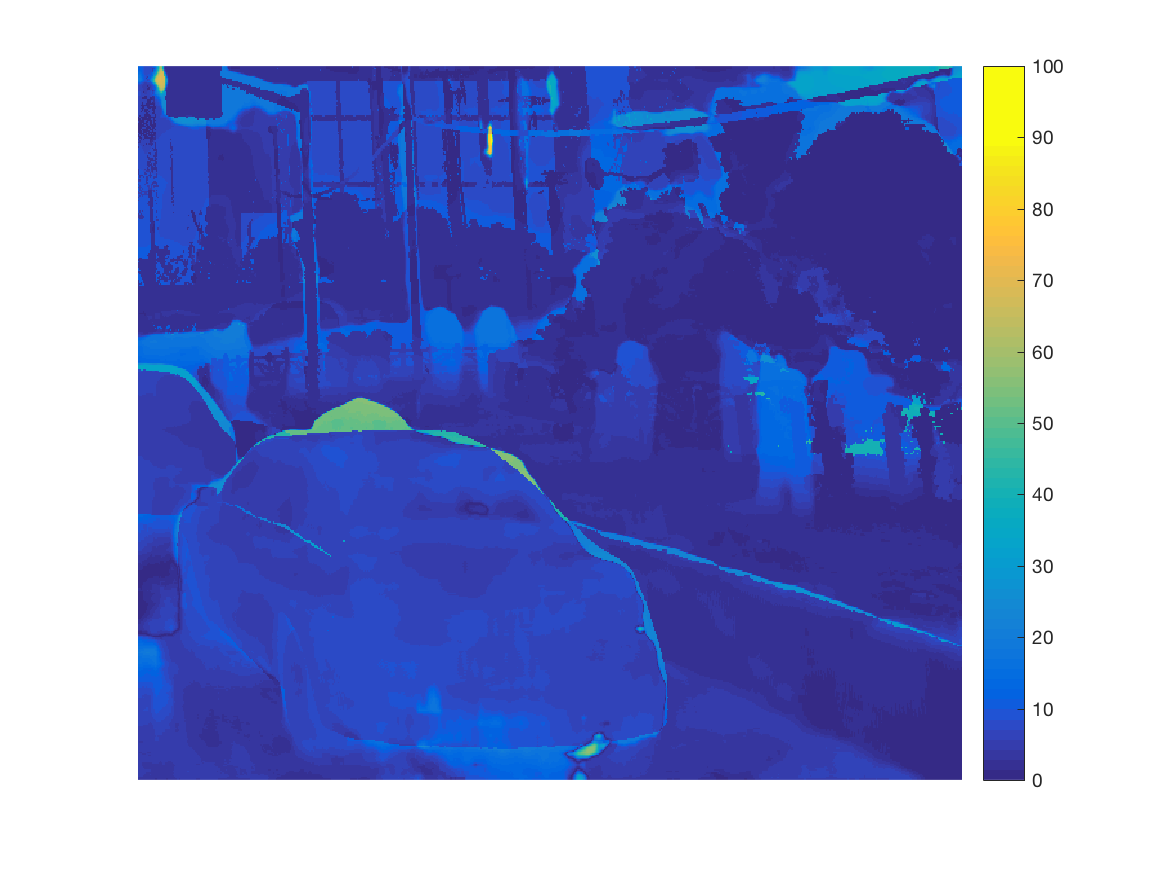} & 
		\includegraphics[trim=2.2cm 2.2cm 1.5cm 1cm, clip=true, width=.16\textwidth]{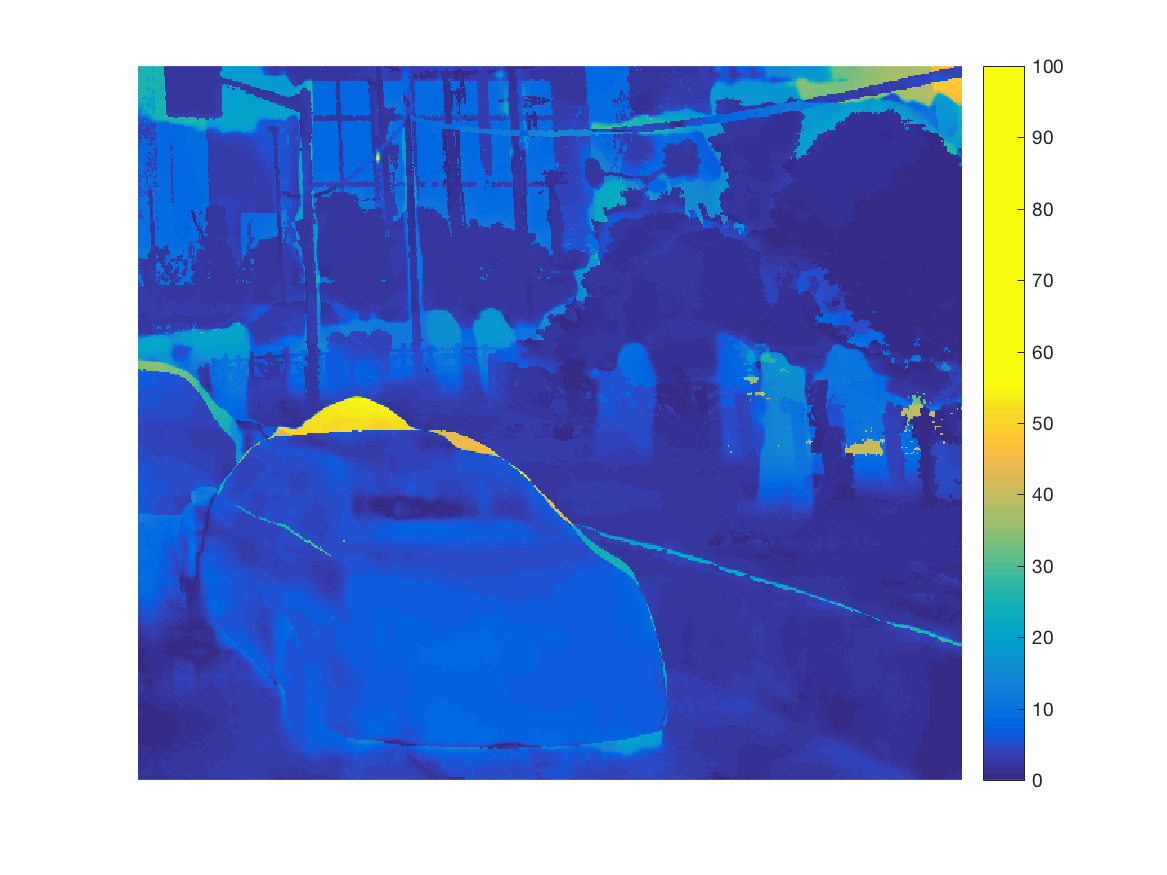} &		
		\includegraphics[trim=2.2cm 2.2cm 1.5cm 1cm, clip=true, width=.16\textwidth]{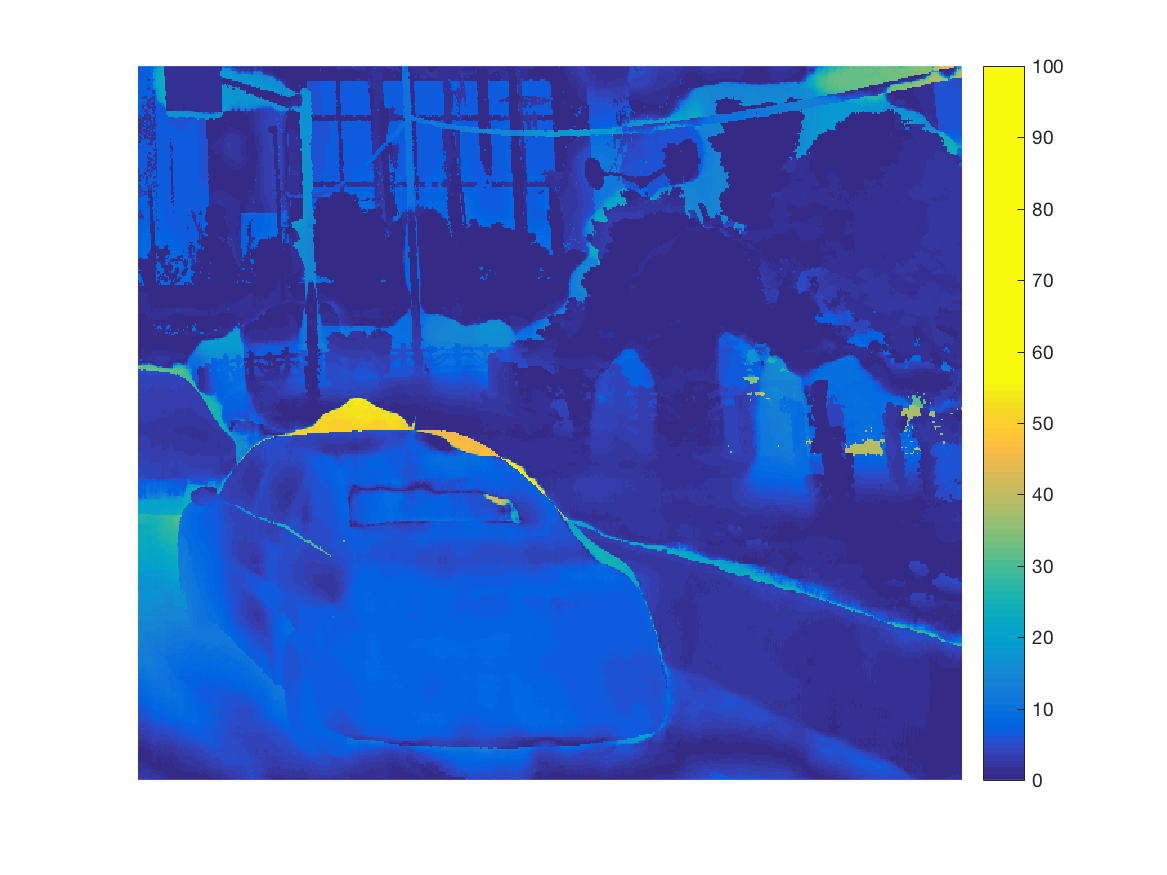} &	
		\includegraphics[trim=2.2cm 2.2cm 1.5cm 1cm, clip=true, width=.16\textwidth]{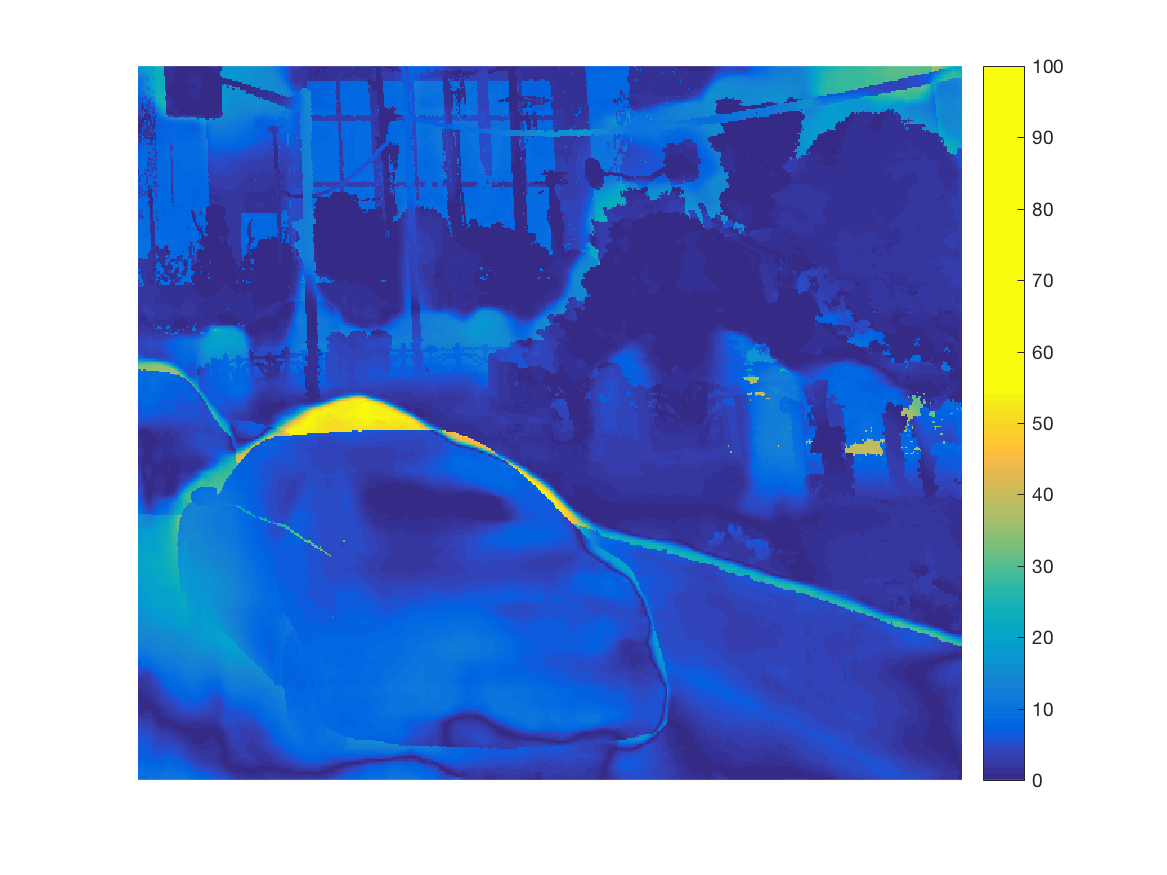} \\		
		\footnotesize{iResNet (Kitti2015)} & \footnotesize{iResNet (ROB) }& \footnotesize{PSMNet (Kitti2012) }& \footnotesize{PSMNet (Kitti2015) }& \footnotesize{SegStereo}& \footnotesize{ Unsup.Adapt }\\ 
		& & & & & \footnotesize{(shadowsontruck) } \\
		\multicolumn{6}{c}{\footnotesize{(a) Results on the images of Fig.~\ref{fig:apolloscape_examples}-(a).} }\\
		\includegraphics[trim=2.2cm 2.2cm 1.5cm 1cm, clip=true, width=.16\textwidth]{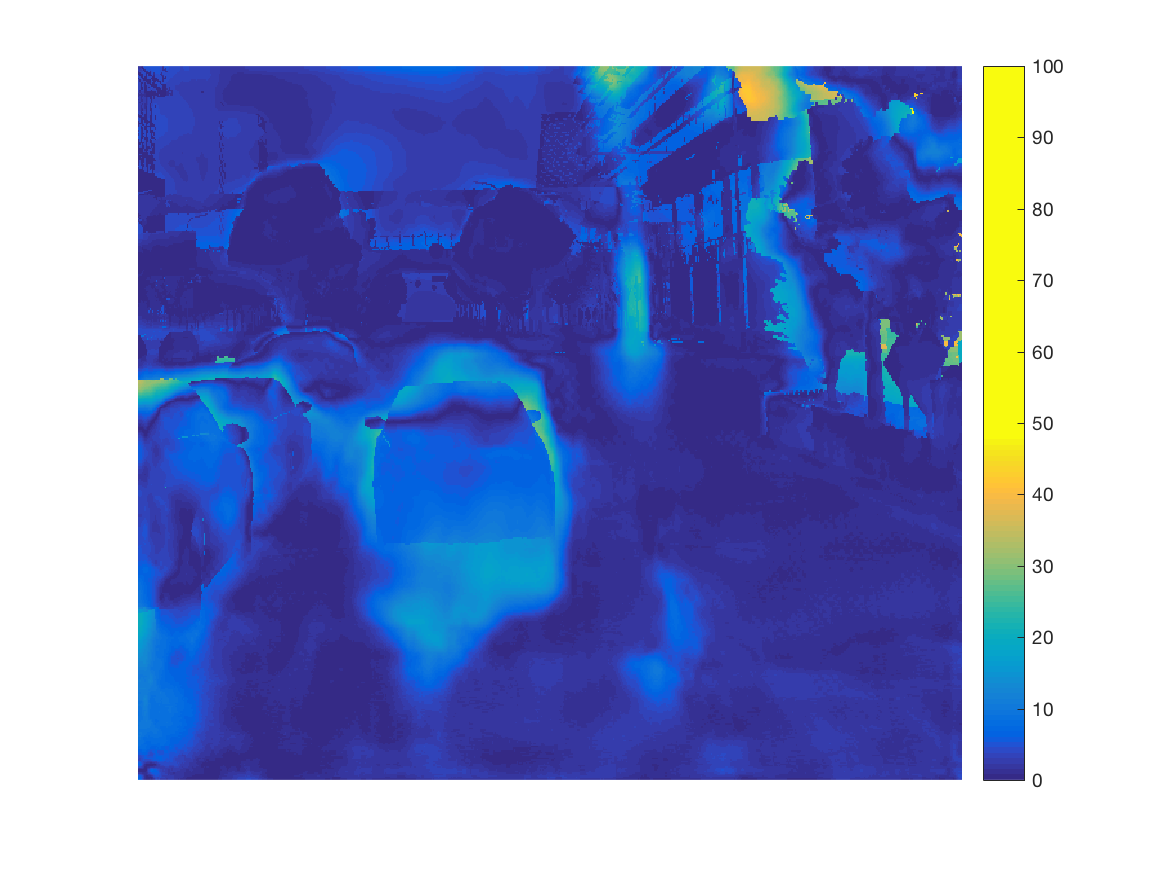} & 		
		\includegraphics[trim=2.2cm 2.2cm 1.5cm 1cm, clip=true, width=.16\textwidth]{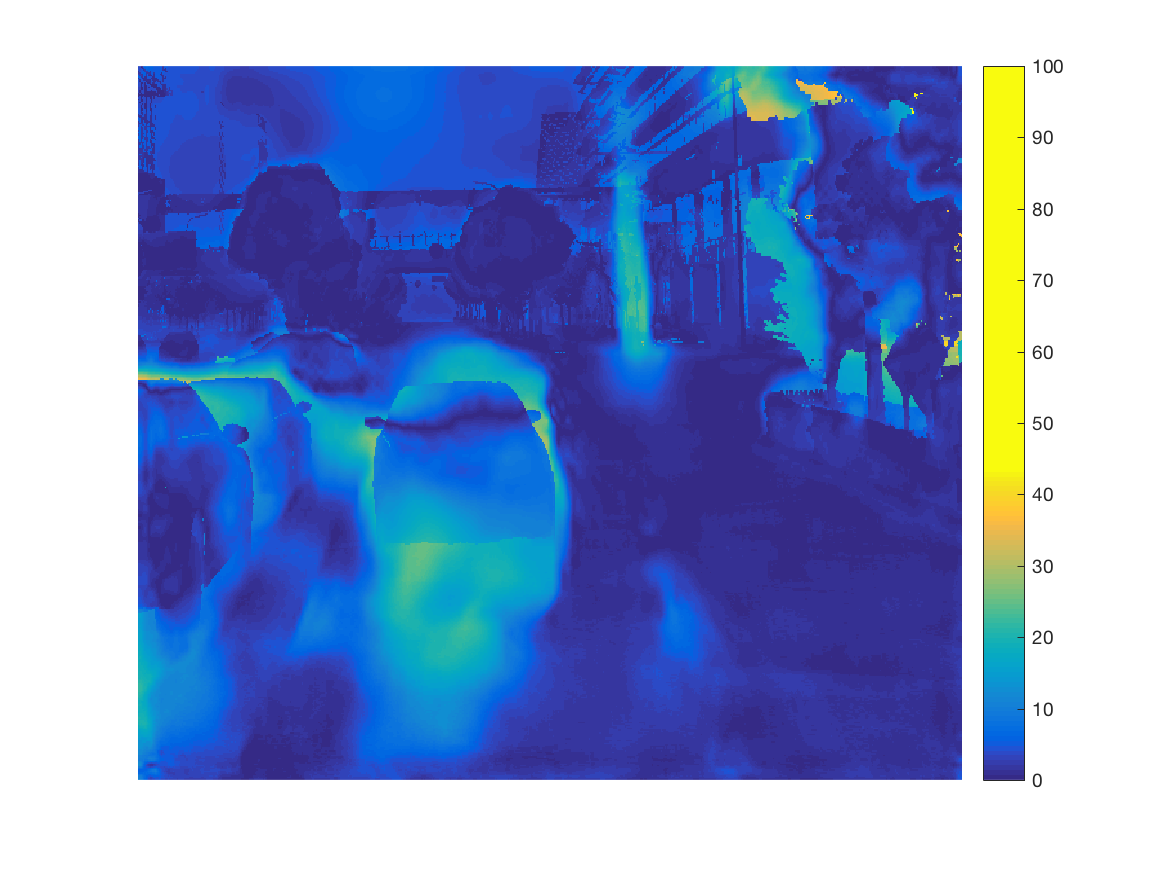} & 
		\includegraphics[trim=2.2cm 2.2cm 1.5cm 1cm, clip=true, width=.16\textwidth]{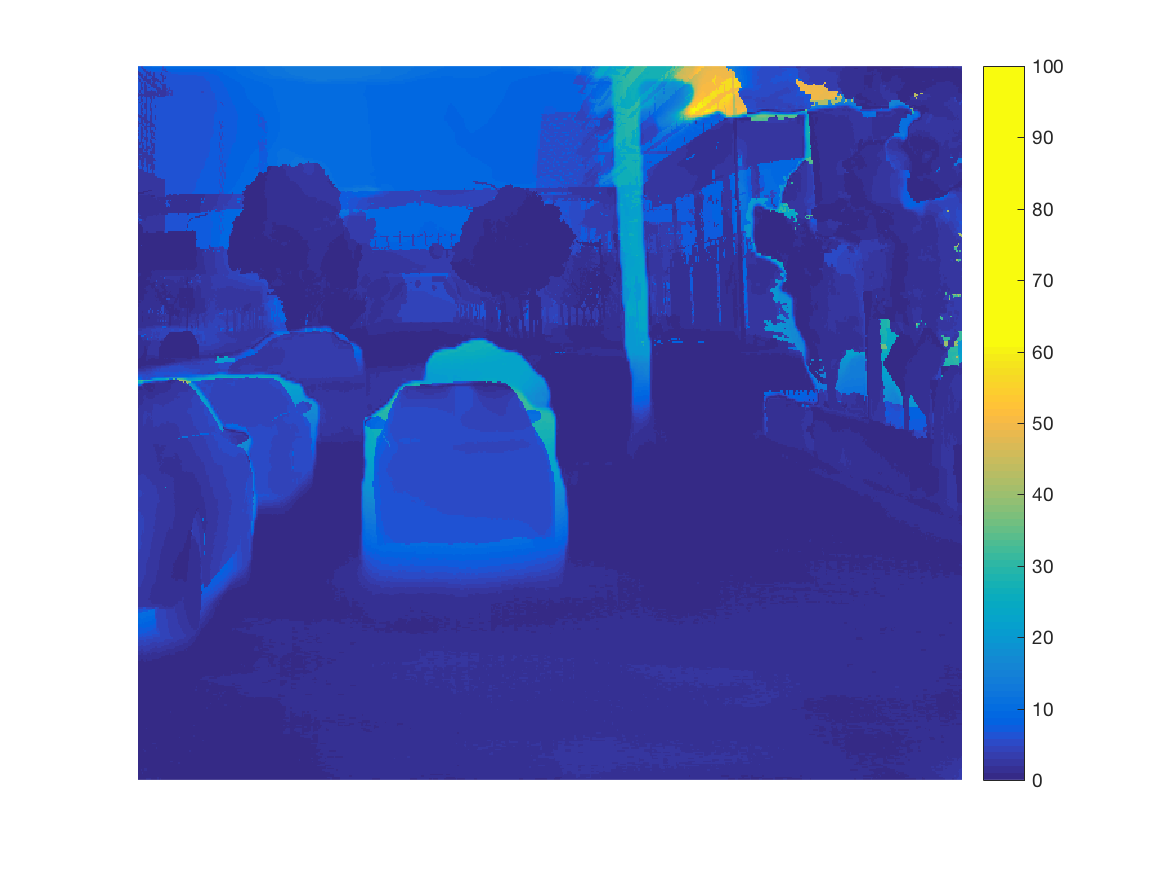} & 
		\includegraphics[trim=2.2cm 2.2cm 1.5cm 1cm, clip=true, width=.16\textwidth]{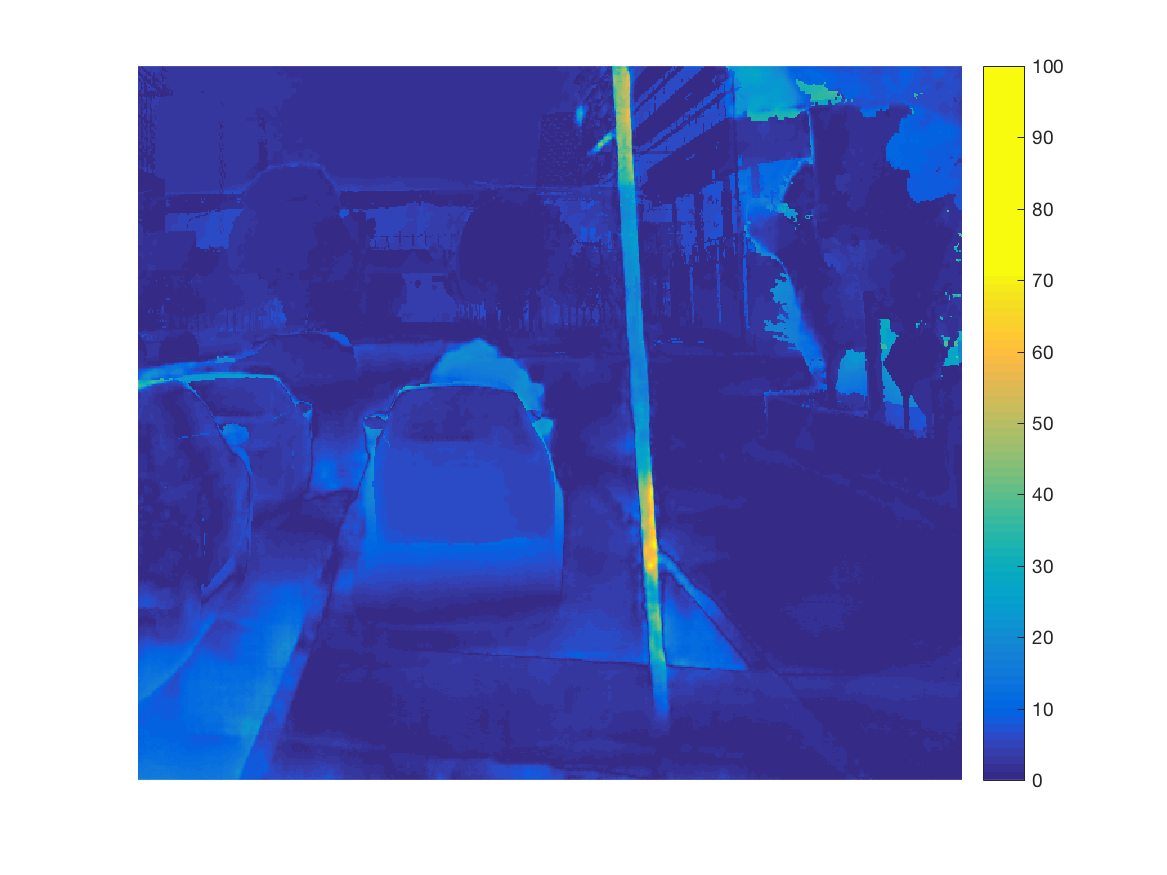} & 
		\includegraphics[trim=2.2cm 2.2cm 1.5cm 1cm, clip=true, width=.16\textwidth]{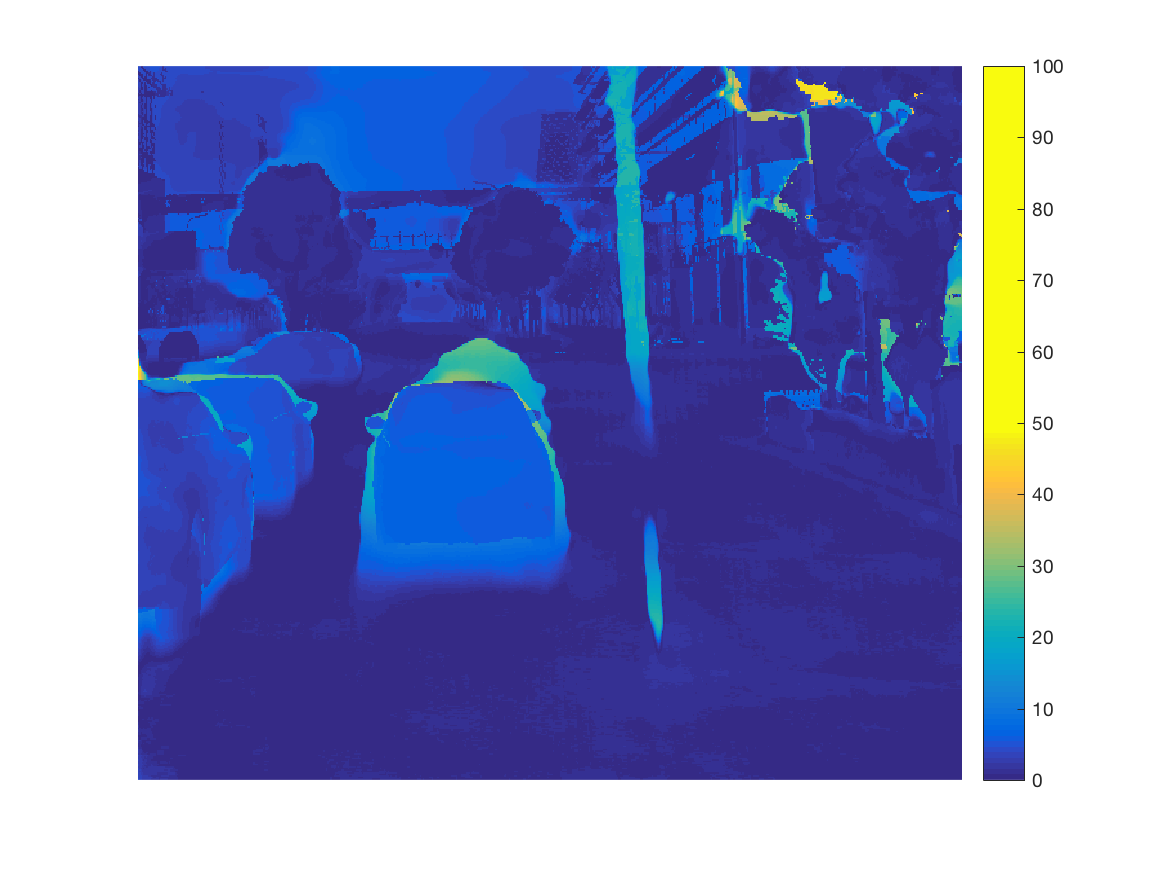} &
		\includegraphics[trim=2.2cm 2.2cm 1.5cm 1cm, clip=true, width=.16\textwidth]{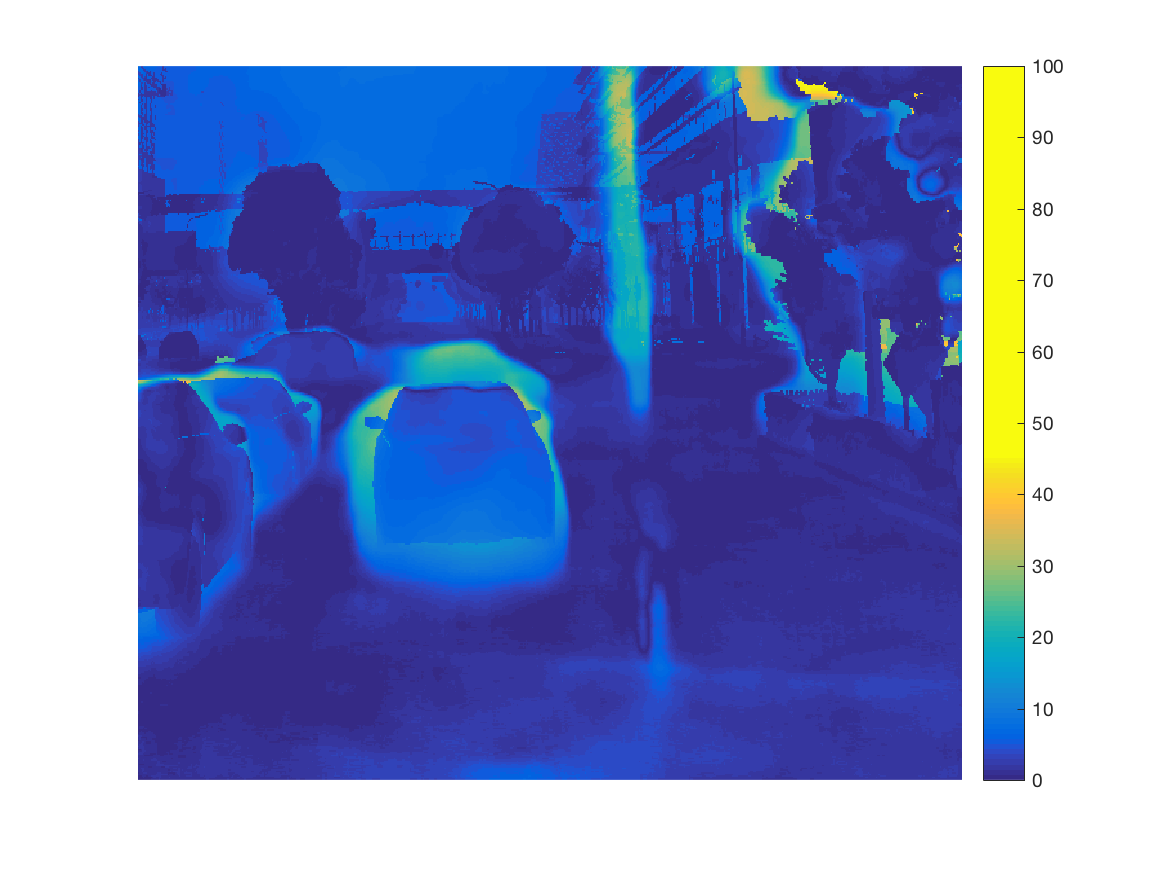} \\
		\footnotesize{AnyNet (Kitti2012)} & \footnotesize{AnyNet (Kitti2015) }& \footnotesize{DeepPruner (fast) }& \footnotesize{DispNet3 (css) }& \footnotesize{GANet (Kittit2015) }& \footnotesize{Unsup.Adapt (Kitti) }\\

		\includegraphics[trim=2.2cm 2.2cm 1.5cm 1cm, clip=true, width=.16\textwidth]{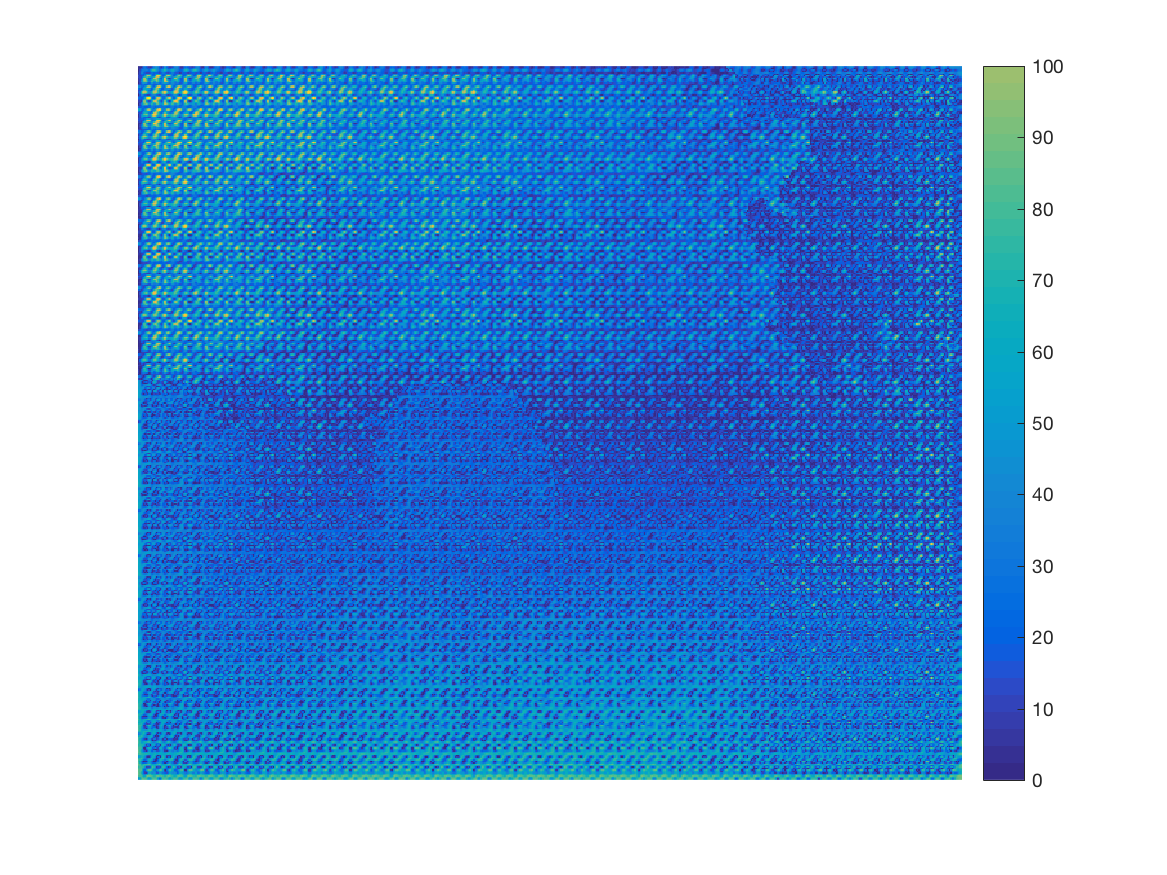} & 
		\includegraphics[trim=2.2cm 2.2cm 1.5cm 1cm, clip=true, width=.16\textwidth]{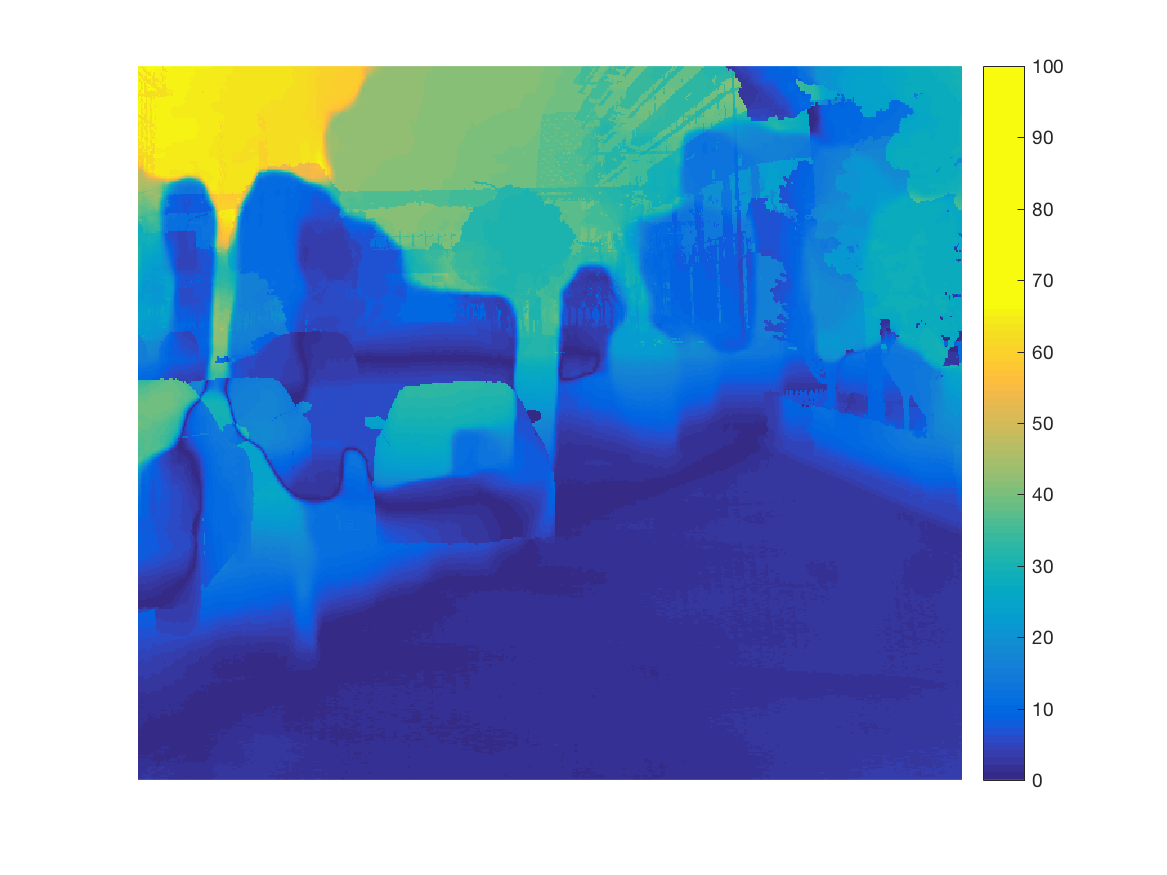} & 
		\includegraphics[trim=2.2cm 2.2cm 1.5cm 1cm, clip=true, width=.16\textwidth]{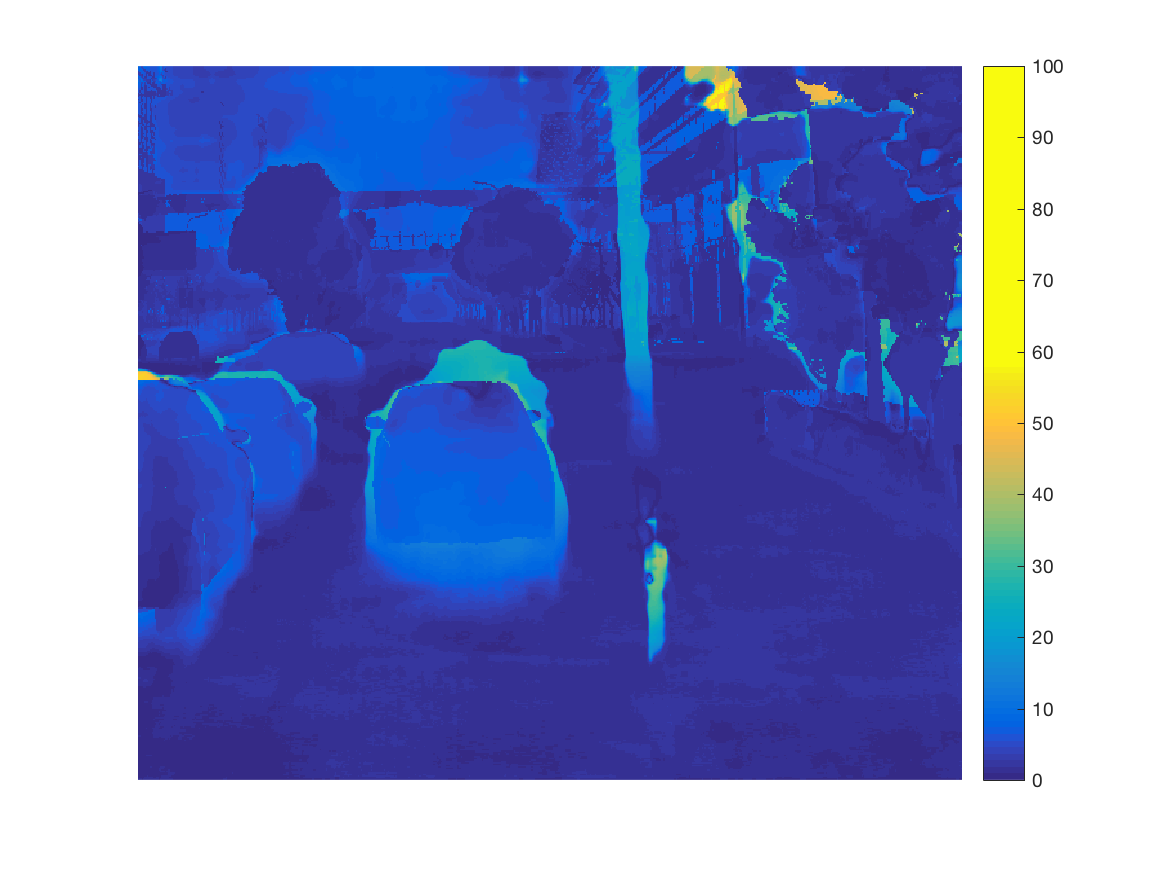} & 
		\includegraphics[trim=2.2cm 2.2cm 1.5cm 1cm, clip=true, width=.16\textwidth]{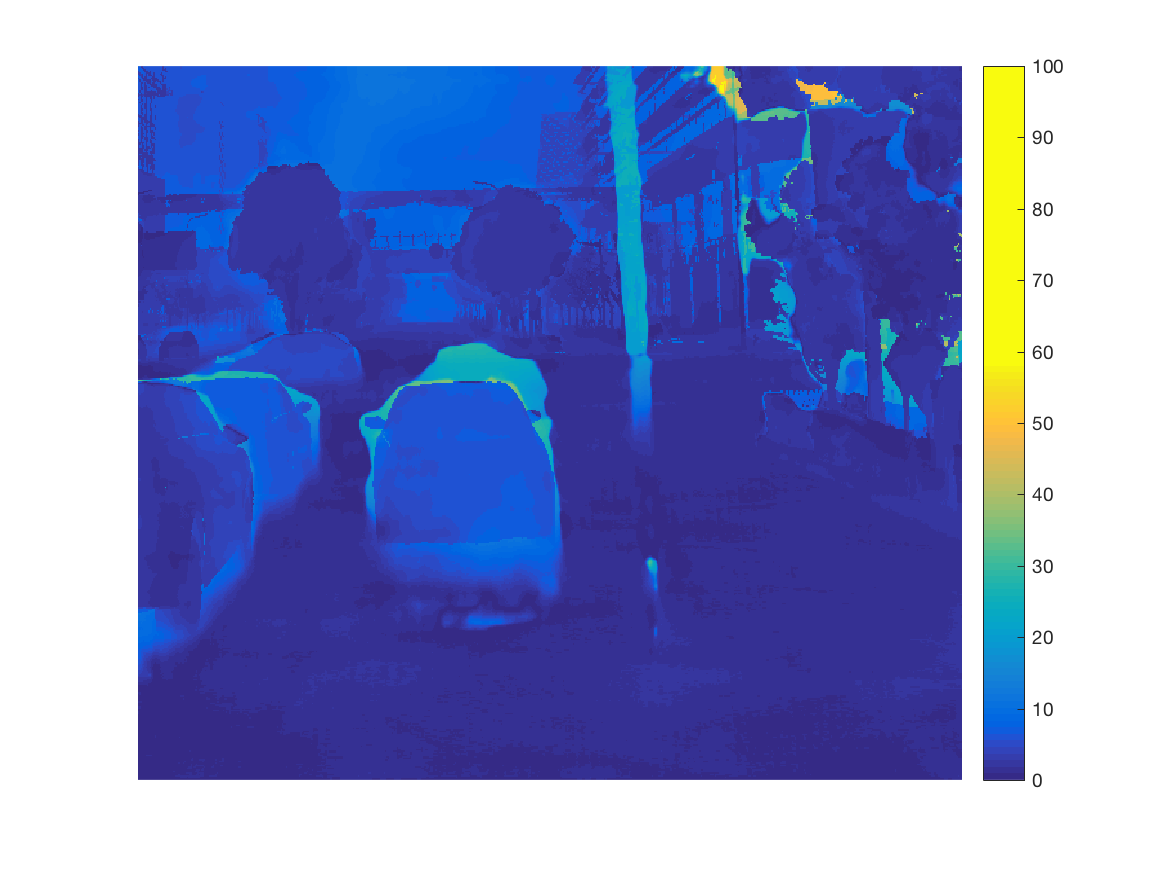} & 
		\includegraphics[trim=2.2cm 2.2cm 1.5cm 1cm, clip=true, width=.16\textwidth]{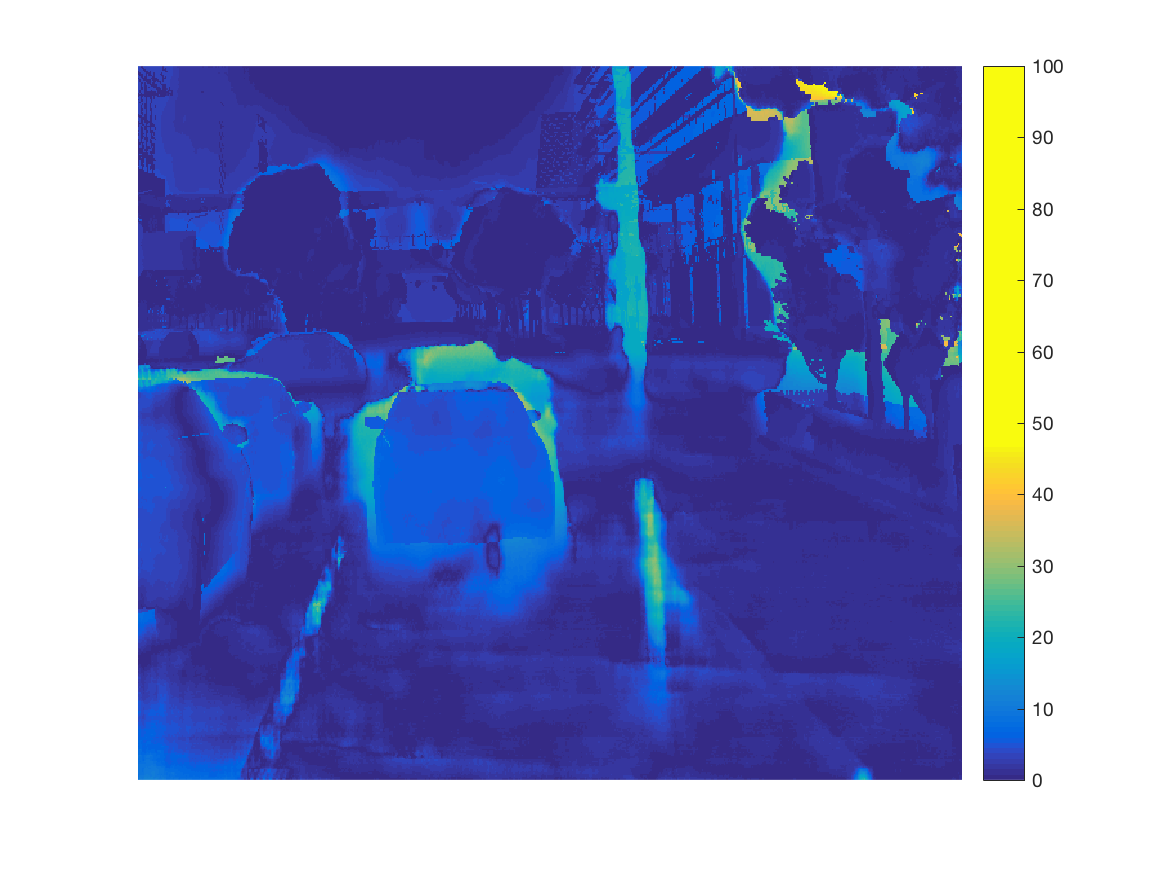} &
		\includegraphics[trim=2.2cm 2.2cm 1.5cm 1cm, clip=true, width=.16\textwidth]{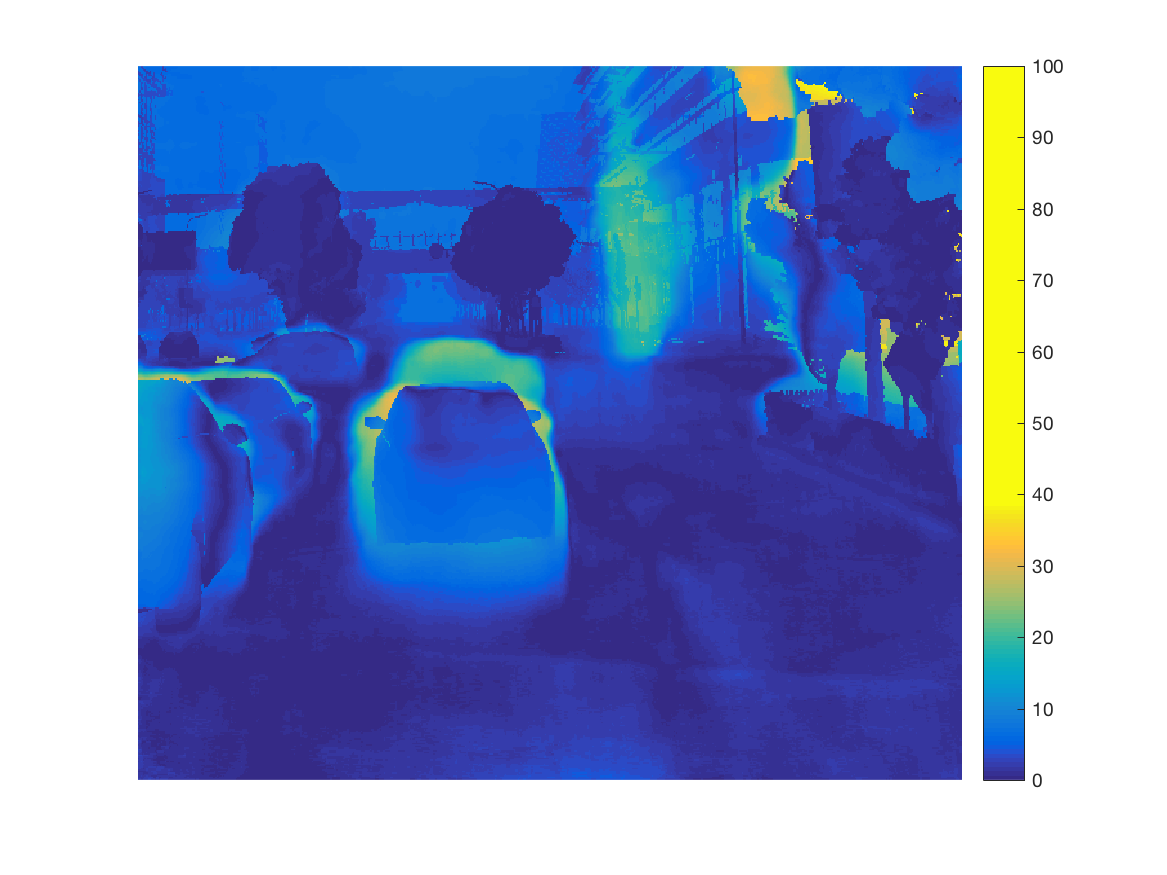} \\
		
		\footnotesize{iResNet (Kitti2015)} & \footnotesize{iResNet (ROB) }& \footnotesize{PSMNet (Kitti2012) }& \footnotesize{PSMNet (Kitti2015) }& \footnotesize{SegStereo}& \footnotesize{ Unsup.Adapt   }\\ 
		& & & & & \footnotesize{(shadowsontruck) } \\
		\multicolumn{6}{c}{\footnotesize{(b) Results on the images of Fig.~\ref{fig:apolloscape_examples}-(b).} }		
	\end{tabular}
}
\caption{\label{fig:apolloscape_errors} Pixel-wise errors between the ground-truth disparities and the disparities estimated  from the images  of Fig.~\ref{fig:apolloscape_examples}.   }
\end{figure*}

In terms of the visual quality of the estimated disparities, see Fig.~\ref{fig:apolloscape_errors},  we observe that  most of the methods were able to recover the overall shape of trees but fail to reconstruct the details especially the leaves. The reconstruction errors are high in flat areas and around object boundaries.  Also, highly reflective materials and poor lighting conditions remain a big challenge to these methods as shown in Fig.~\ref{fig:apolloscape_errors}-(b). The supplementary material provides more results on the four stereo pairs of Fig.~\ref{fig:novel_dataset}.


\section{Future research directions}
\label{sec:future}

Deep learning methods for stereo-based depth estimation  have achieved promising results. The topic, however, is still in its infancy and further developments are yet to be expected. In this section, we present some of the current issues and highlight  directions for future research.    

\vspace{6pt}
\noi\textit{(1) Camera parameters. } Most of the stereo-based  techniques surveyed in this article require rectified images.  Multi-view stereo techniques use  Plane-Sweep Volumes or back-projected  images/features. Both  image rectification and PSVs  require known camera parameters, which are challenging to estimate in the wild.   Many  papers attempted to address this problem  for monocular depth estimation and for 3D shape reconstruction by jointly optimising for the camera parameters and the geometry of the 3D scene~\cite{han2020image}.

\vspace{6pt}
\noi\textit{(2) Lighting conditions and complex material properties.  }  Poor lighting conditions and complex materials properties  remain a challenge to most of the current methods, see for example Fig.~\ref{fig:apolloscape_errors}-(b). Combining object recognition,  high-level  scene understanding, and  low-level feature learning can be one promising avenue to address these issues.

\vspace{6pt}
\noi\textit{(3) Spatial and depth resolution. } Most of the current techniques do not handle high resolution input images and generally produce depth maps of low  spatial and depth resolution. Depth resolution is particularly limited, making the methods unable to reconstruct thin structures,  \eg vegetation and hair,  and structures located at a far distance from the camera.  Although refinement modules can improve the resolution of the estimated depth maps, the gain  is still small compared to the resolution of the input images.   This has recently been addressed using hierarchical techniques, which allow on-demand reports of disparity by capping the resolution of the intermediate results~\cite{Yang_2019_CVPR}. In these methods, low resolution depth maps can be produced in realtime, and thus can be used on mobile platforms, while high resolution  maps would require more computation time.  Producing, in realtime,  accurate maps of high spatial and depth resolutions remains a challenge  for future research.  

\vspace{6pt}
\noi\textit{(4) Realtime processing. } Most deep learning methods for disparity estimation use 3D and 4D cost volumes, which are processed and regularized using 2D and 3D convolutions. They are expensive in terms of memory requirements and processing time. Developing lightweight, and subsequently fast, end-to-end deep networks remains a challenging avenue for future research.

\vspace{6pt}
\noi\textit{(5) Disparity range. } Existing  techniques uniformly discretize the disparity range. This results in multiple issues.  In particular, although the reconstruction error can be small in the disparity space, it can result in an error of meters in the depth space, especially at far ranges. One way to mitigate this is by discritizing disparity and depth uniformly in the log space.   Also, changing the disparity range requires retraining the networks.  Treating depth as a continuum could be one promising avenue for future research.

\vspace{6pt}
\noi\textit{(6) Training. } Deep networks heavily rely  on the availability of training images annotated with ground-truth labels. This is very expensive and labor intensive for  depth/disparity reconstruction. As such, the performance of the methods and their generalization ability can significantly be affected including the risk of overfitting the models to specific domains.  Existing techniques mitigate this problem by either designing loss functions that do not require 3D annotations, or by using domain adaptation and transfer learning strategies. The former, however, requires calibrated cameras.  Domain adaptation techniques, especially unsupervised ones~\cite{tonioni2019unsupervised}, are recently attracting more attention since, with these techniques, one can train with both synthetic data, which are easy to obtain, and real-world data. They also adapt, in an unsupervised manner and at run-time to ever-changing environments as soon as new images are gathered. Early results are very encouraging and thus expect in the future to see the emergence of large datasets, similar to ImageNet but for 3D reconstruction.

	 
	

\vspace{6pt}
\noi\textit{(7) Automatically learning the network architecture, its activation functions,  and its parameters from data. }  Most existing research has focused on designing novel network architectures and novel training methods for optimizing their parameters. It is only recently that some papers started to focus on automatically learning optimal architectures. Early attempts, \eg~\cite{Saikia_2019_ICCV}, focus on simple architectures. We  expect in the future to see more research on automatically learning complex disparity estimation architectures and their activation functions, using, for example, the neuro-evolution theory~\cite{stanley2019designing,bingham2020evolutionary}, which will free the need for manual network design.  
	
\section{Conclusion}
\label{sec:conclusion}

This paper provides a comprehensive survey of the recent developments in stereo-based depth estimation using deep learning techniques.  Despite their infancy, these techniques are achieving state-of-the-art  results.  Since 2014, we have entered a new era where data-driven and machine learning techniques play a central role in image-based depth reconstruction. We have seen that, from $2014$ to $2019$, more than $150$ papers on the topic have been published in the major computer vision, computer graphics,  and machine learning conferences and journals.  Even during the   final stages of this submission, more new papers are being published making it difficult to keep track of the recent developments, and more importantly, understand their differences and similarities, especially for new comers to the field. This timely survey  can thus serve as a guide to the reader to navigate this fast-growing field of research. 

Finally, there are several related topics that have not been covered in this survey. Examples include image-based 3D object reconstruction using deep learning, which has been recently surveyed by Han \etal~\cite{han2020image}, and monocular and video-based depth estimation, which requires a separate survey paper given the large amount of papers that have been published on the topic in the past $5$ to $6$ years. Other topics  include photometric stereo and active stereo~\cite{Haefner_2019_ICCV,Zheng_2019_ICCV}, which are outside the scope of this paper.

\small{
\vspace{6pt}
\noi\textbf{Acknowledgements. }  We would like to thank all the authors of the reference papers who have made their codes and datasets publicly available. This work is supported in part by Murdoch University's Vice Chancellor's Small Steps of Innovation Funding Program, and by ARC DP150100294 and DP150104251. 
}


\bibliographystyle{IEEEtran}
\bibliography{reconstruction}

\begin{IEEEbiography}[{\includegraphics[width=1in,height=1.25in,clip,keepaspectratio]{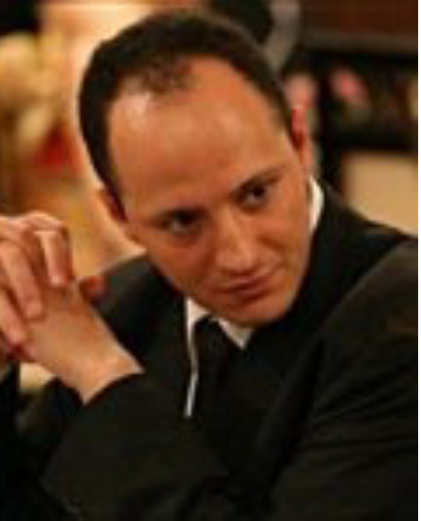}}]{Hamid Laga}  received the PhD degrees in Computer Science from Tokyo Institute of Technology in 2006. He is currently  an Associate Professor at Murdoch University (Australia) and an Adjunct Associate Professor with the Phenomics and Bioinformatics Research Centre (PBRC) of the University of South Australia. His research interests span various fields of machine learning, computer vision, computer graphics, and pattern recognition, with a special focus on the 3D reconstruction, modeling and analysis of static and deformable 3D objects, and on image analysis and big data in agriculture and health. He is the recipient of  the Best Paper Awards at SGP2017, DICTA2012, and SMI2006.
\end{IEEEbiography}

\begin{IEEEbiography}[{\includegraphics[width=1in,height=1.25in,clip,keepaspectratio]{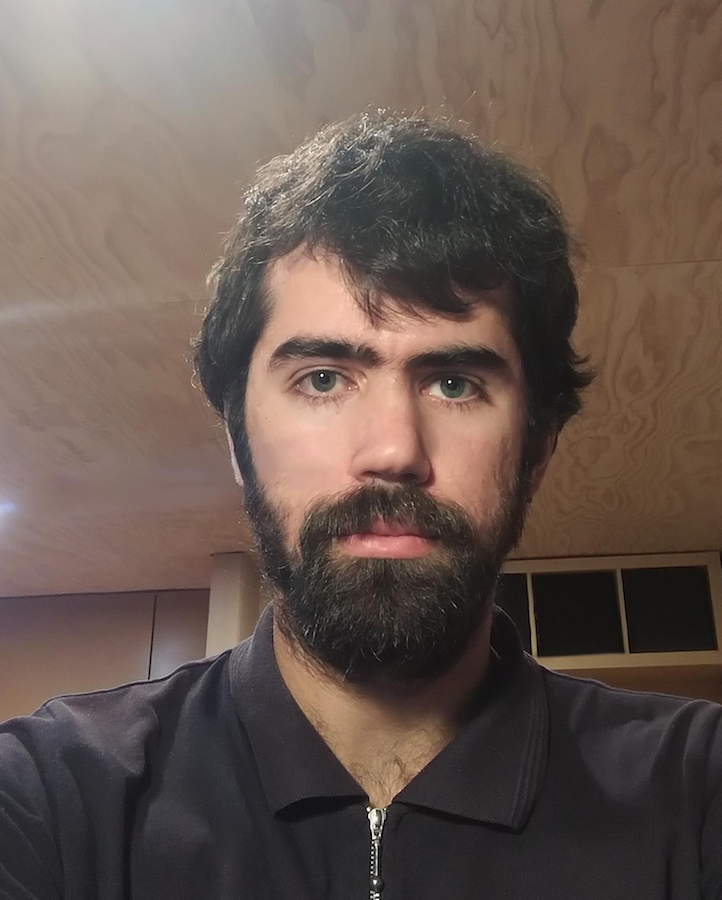}}]{Laurent Valentin Jospin} Laurent Valentin Jospin is a young research student in the field of computer vision. His main research interests include 3d reconstruction, sampling and image acquisition strategies, computer vision applied to robotic navigation and computer vision applied to environmental sciences.
Holder of a master of science in environmental engineering from EPFL since 2017 with a thesis on accuracy prediction in aerial mapping, his research career started as intern in two EPFL labs, publishing his first three papers in the process, before starting a PhD in computer science at the University of Western Australia in 2019. His thesis project focus on real time 3D reconstruction with different computer vision techniques.
\end{IEEEbiography}

\begin{IEEEbiography}[{\includegraphics[width=1in,height=1.25in,clip,keepaspectratio]{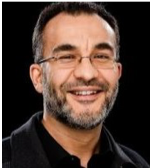}}]{Faird Boussaid}  received the M.S. and Ph.D. degrees in microelectronics from the
National Institute of Applied Science (INSA), Toulouse, France, in 1996 and 1999 respectively. He joined Edith Cowan University, Perth, Australia, as a Postdoctoral Research Fellow, and a Member of the Visual Information Processing Research Group in 2000. He joined the University of Western Australia, Crawley, Australia, in 2005, where he is currently a Professor. His current research interests include neuromorphic engineering, smart sensors, and machine learning.
\end{IEEEbiography}

\begin{IEEEbiography}[{\includegraphics[width=1in,height=1.25in,clip,keepaspectratio]{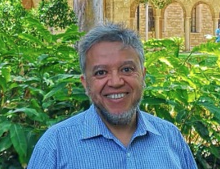}}]{Mohammed Bennamoun}  is Winthrop Professor in the Department of Computer Science and Software Engineering at UWA and is a researcher in computer vision, machine/deep learning, robotics, and signal/speech processing. He has published 4 books (available on Amazon, 1 edited book, 1 Encyclopedia article, 14 book chapters, 120+ journal papers, 250+ conference publications, 16 invited and keynote publications. His h-index is 47 and his number of citations is 10,000+ (Google Scholar). He was awarded 65+ competitive research grants, from the Australian Research Council, and numerous other Government, UWA and industry Research Grants. He successfully supervised +26 PhD students to completion. He won the Best Supervisor of the Year Award at QUT (1998), and received award for research supervision at UWA (2008 and 2016) and Vice-Chancellor Award for mentorship (2016).  He delivered conference tutorials at major conferences, including: IEEE  CVPR 2016, Interspeech 2014, IEEE ICASSP, and ECCV. He was also invited to give a Tutorial at an International Summer School on Deep Learning (DeepLearn 2017).
\end{IEEEbiography}

\end{document}